\PassOptionsToPackage{capitalize}{cleveref}
\documentclass{fairmeta}

\usepackage[utf8]{inputenc}
\usepackage{url}
\usepackage{amsfonts}
\usepackage{nicefrac}
\usepackage{amsmath}
\usepackage{amssymb}
\usepackage{mathtools}
\usepackage{amsthm}
\usepackage{enumitem}
\usepackage{makecell}

\theoremstyle{plain}

\theoremstyle{definition}

\theoremstyle{remark}

\AtBeginDocument{\hbadness=10000\hfuzz=200pt}

\title{{\setstretch{1.08}\fontsize{17}{20}\selectfont Phys4D: Injecting Physical Structure into Video Diffusion via Lifted RGB-D-Motion 4D Interfaces\par}}

\author[1,*]{Haoran Lu}
\author[1,*]{Shang Wu}
\author[3,4,*]{Songling Liu}
\author[1]{Jianshu Zhang}
\author[1]{Maojiang Su}
\author[1]{Guo Ye}
\author[1]{Chenwei Xu}
\author[2]{Lie Lu}
\author[2]{Pranav Maneriker}
\author[2]{Fan Du}
\author[1]{Zhaoran Wang}
\author[1]{Han Liu}

\affiliation[1]{Northwestern~University}
\affiliation[2]{Dolby~Laboratories}
\affiliation[3]{Institute~of~Automation,~Chinese~Academy~of~Sciences,~Beijing~100190}
\affiliation[4]{University~of~Chinese~Academy~of~Sciences}

\contribution[*]{Equal contribution}

\abstract{
Recent video diffusion models generate increasingly realistic videos, yet visual plausibility alone does not imply physical understanding. Physical laws govern evolving 3D scene states, while videos observe only their 2D projections. We introduce \textbf{Phys4D}, a structure-focused, appearance-constrained physics injection framework that transfers simulation-derived physical structure into pretrained video diffusion models while preserving their real-video generative prior. Phys4D operationalizes physics-aware video generation through a prior-preserving RGB-D-motion interface. Our key principle is \textbf{simulation teaches physics, not appearance}: simulation provides dense structural supervision, including depth, motion, and masks, while adaptation remains anchored to the pretrained real-video prior rather than drifting toward simulator-specific visual statistics. To this end, Phys4D exposes geometry and motion with lightweight prediction heads, injects local physical structure through gated geometry-motion supervision, and aligns sequence-level evolution using a simulation-grounded trajectory objective. Supported by a scalable coupled-physics simulation system, Phys4D improves observable physical behavior, geometry-motion consistency, and trajectory-level evolution across multiple video generation backbones, while maintaining visual realism and fidelity. We release the code, datasets, introduction slides, and video demos on the project page: https://sensational-brioche-7657e7.netlify.app/
}

\begin{document}

\maketitle

\section{Introduction}
\label{sec:intro}

\begin{figure*}[htb]
\centering
\includegraphics[width=\textwidth]{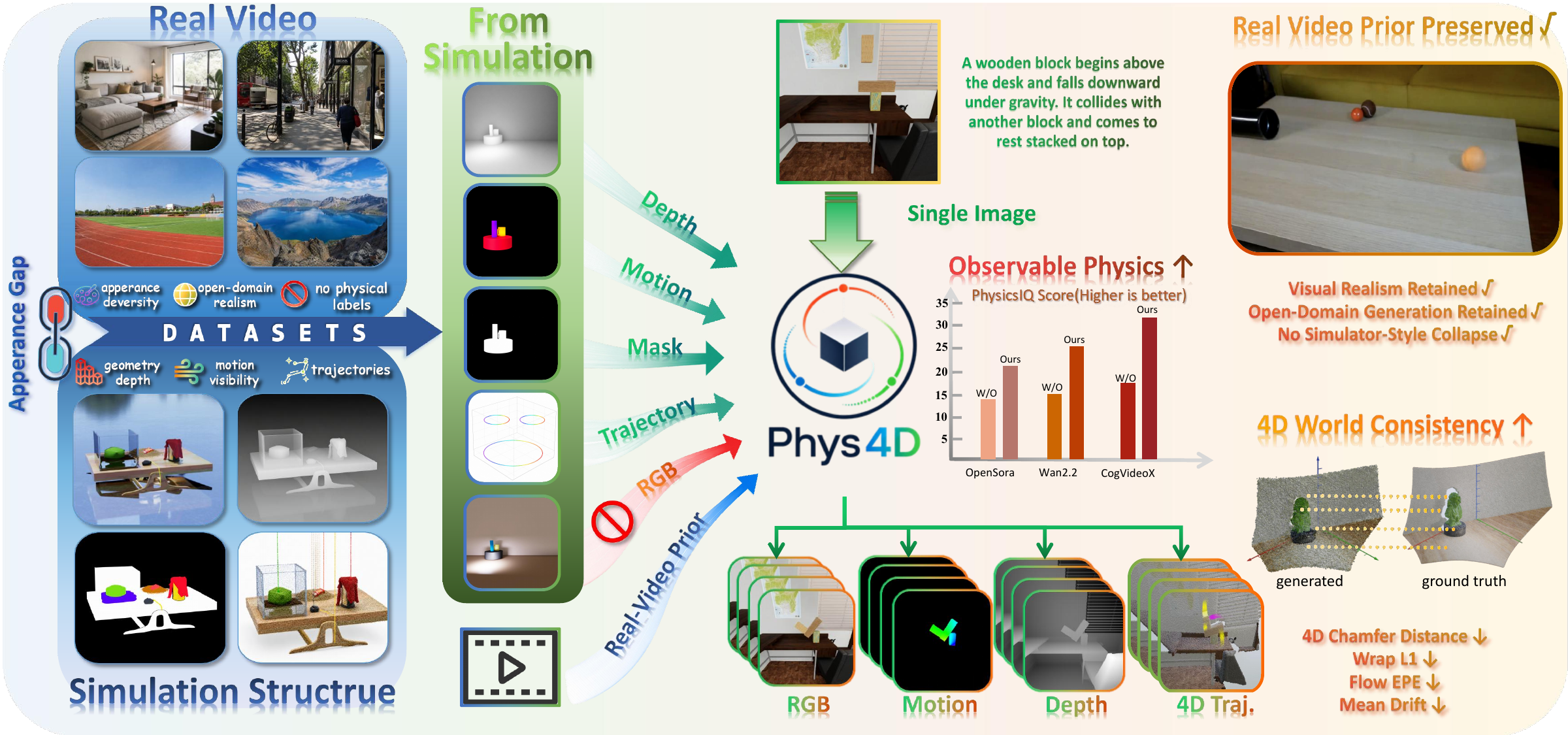}
\caption{
\textbf{Phys4D: simulation teaches physics, not appearance.}
Phys4D uses simulation-derived depth, motion, masks, and trajectories to inject physical structure into pretrained video diffusion models through a 4D world interface, while blocking simulator RGB as the dominant appearance target.
This improves observable physics and 4D geometry--motion consistency while preserving the real-video generative prior.
}
\label{fig:head_picturel}
\end{figure*}

Recent video generation models~\cite{openai2024sora, google2024veo2, polyak2024movie, chen2025teleworld,yu2024wonderworld,Yu2023WonderJourneyGF,Huang2025VoyagerLA} have made remarkable progress in visual realism. Yet visual plausibility should not be mistaken for physical understanding~\cite{vbench2025vbench2,motamed2025physicsiq}. Physical laws govern world states in 3D, including geometry, depth ordering, object motion, contact, and deformation, whereas a video observes only their 2D projection. We therefore view physics-aware video generation as requiring \textbf{temporally consistent 3D scene evolution}. We operationalize this view through a scalable \textbf{RGB-D-motion interface}: throughout this paper, \textbf{``4D'' refers to geometry and motion evolving over time, rather than full volumetric dynamic scene reconstruction.} This interface remains compatible with video diffusion models while exposing physical variables for geometry-level and motion-level supervision.

A central challenge is how to inject physical knowledge into video generation models. Fine-grained dynamics involve diverse physical regimes, including motion, contact, and multi-object coupling. Manually translating each regime into generator-specific losses, rewards, or constraints does not scale to open-ended interactions. We instead use physics-based simulation as an \textbf{executable form of human physical knowledge}: physical principles are encoded through solvers, constraints, contacts, materials, and coupling mechanisms, and rendered into video-compatible supervision. This provides dense, temporally aligned signals---including depth, motion, visibility, object masks, and trajectories.

However, simulation also introduces a key problem. Simulated videos are generated from procedural assets and renderer-specific statistics, making them less visually diverse than internet-scale real videos and mismatched in appearance. Directly fitting a video generator to such data risks transferring simulator-specific visual biases and degrading the pretrained real-video prior. Thus, simulation must be used selectively: \textbf{to prioritize physical structure while constraining appearance drift}.

This leads to \textbf{Phys4D}, a structure-focused, appearance-constrained physics injection framework that transfers simulation knowledge into pretrained video diffusion models while preserving their real-video generative prior. Phys4D proceeds in three stages. First, it constructs a \textbf{prior-preserving RGB-D-motion interface} by freezing the pretrained video backbone and training only lightweight depth and motion heads. These heads are supervised by pseudo-labels from strong monocular estimators on both self-generated videos and curated in-the-wild internet videos, allowing the model to expose geometry and motion without altering its RGB generator. Second, Phys4D injects local physical structure from simulation through \textbf{gated geometry-motion supervision}. Rather than directly fine-tuning the full generator on simulator videos, simulation-induced updates are confined to high-noise adapters and mask-weighted toward physics-relevant dynamic regions, where depth, motion, and warp consistency couple geometry with motion. Third, Phys4D aligns the \textbf{global 4D evolution} of generated scenes with simulation. Since trajectory-level alignment is a sample-level, non-local property that depends on completed scene evolution and is not captured by local pixel- or correspondence-based losses, we lift generated RGB-D-motion predictions into 4D trajectories and optimize a simulation-grounded unified policy objective.

To support these stages, we build a \textbf{scalable coupled-physics simulation system} that provides structural supervision for Phys4D. Unlike narrow single-regime simulation pipelines, our system asynchronously generates diverse interactions across rigid, deformable, fluid, and coupled dynamics, and exports unified geometry, motion, mask, visibility, and trajectory annotations. This design enables local geometry-motion injection and global trajectory-level alignment without relying on manually designed, category-specific physics rules.

We conduct experiments to verify whether Phys4D improves physical consistency while preserving the pretrained generative prior. We evaluate generated videos from three complementary perspectives: observable video-level physics, 4D geometry-motion consistency, and preservation of visual quality and diversity. Across multiple video generation backbones, Phys4D improves fine-grained physical behavior and produces more coherent depth, motion, and trajectory evolution, while maintaining appearance realism and open-domain generation ability.

\noindent\textbf{In summary, our contributions are as follows:}
\begin{itemize}[leftmargin=*, itemsep=0.1em, topsep=0.3em]
    \item We formulate physical knowledge injection for video generation as a \textbf{structure-transfer problem}, where simulation provides geometry, motion, and trajectory supervision while the pretrained real-video generative prior is preserved.

    \item We introduce \textbf{Phys4D}, a structure-focused physics injection framework that constructs a prior-preserving RGB-D-motion interface, injects local physical structure through gated geometry-motion supervision, and aligns global scene evolution through a unified trajectory-level policy objective.

    \item We build a \textbf{scalable coupled-physics simulator} and show that Phys4D improves fine-grained physical behavior, geometry-motion consistency, and trajectory-level evolution across multiple video backbones while maintaining visual realism, diversity, and open-domain generative capacity.
    \item To facilitate reproducibility and broader use, we release our \textbf{ code, simulation datasets, introduction slides, and video demos} on the project page.
\end{itemize}
\section{Related Work}
\label{sec:related}
\textbf{Physics-aware video generation and evaluation.}
Large-scale video generators achieve strong visual realism, but physical state remains largely implicit in RGB videos~\cite{openai2024sora,polyak2024movie,yang2025cogvideoxtexttovideodiffusionmodels,tencent2024hunyuanvideo,wan2025wanopenadvancedlargescale}. Recent benchmarks expose this gap through tests of physical commonsense, dynamics, and intrinsic faithfulness~\cite{phygenbench2025,motamed2025physicsiq,videophy2024,vbench2025vbench2}. Physics-aware methods improve plausibility using motion priors, reinforcement learning, reasoning, or action-conditioned simulation interfaces~\cite{xu2026motionforcing,zhang2026physrvg,Yuan_2025_NewtonGen,liu2026realwonder}. However, they typically operate on video-level outcomes, RGB-observable cues, or task-specific controls. Phys4D instead injects physical knowledge as structure: geometry, motion, and trajectory evolution.

\textbf{4D scene evolution and simulation-based structural supervision.}
Dynamic scene modeling has developed rich 3D/4D representations, including neural fields, dynamic Gaussians, diffusion-based 4D reconstruction, and dynamic point maps~\cite{pumarola2021dnerf,park2021hypernerf,luiten2023dynamic3dgs,wu20244d,wu2024cat4d,sucar2025dynamicpointmaps}. Physically grounded 4D generation further couples explicit geometry with simulation or physical priors~\cite{xie2024physgaussian,dreamphysics2025,chen2025physgen3d,li2025wonderplay}. These methods are effective for reconstruction, animation, and controlled 3D generation, but often rely on specialized representations or explicit 4D optimization. Phys4D instead uses simulation-derived structural supervision to train a prior-preserving RGB-D-motion interface for video diffusion.

\section{Problem Definition}
\label{sec:problem_definition}
Let $c$ denote the generation condition, which may be a text prompt $p$, an initial
image $I_{\mathrm{ref}}$, or an image-text pair $(I_{\mathrm{ref}},p)$. Given $c$, our
goal is to generate a video $\{I_t\}_{t=0}^{T}$ that is visually realistic while
following physically plausible scene evolution. Since physical states such as geometry, depth ordering,
contact, deformation, and motion are only partially observed through 2D videos, we
formulate physics-aware video generation as a \textbf{structure-transfer problem}:
simulation should supervise physical structure, rather than RGB appearance.

We expose physical structure through a compact \textbf{RGB-D-motion interface}. Let
$G_{\theta_0}$ be a pretrained video diffusion model with a strong real-video prior. We
seek an augmented model $G_{\theta}$ that preserves this prior while predicting depth,
motion, and lifted 4D trajectories:
\[
G_{\theta}(c,\epsilon)
\longmapsto
\Big(
\{I_t\}_{t=0}^{T},
\{D_t\}_{t=0}^{T},
\{F_{t\rightarrow t+1}\}_{t=0}^{T-1},
\mathcal{T}
\Big),
\]
where $D_t \in \mathbb{R}^{H \times W}$ is depth,
$F_{t\rightarrow t+1} \in \mathbb{R}^{H \times W \times 2}$ is inter-frame motion,
and $\mathcal{T}$ denotes trajectories lifted from the predicted geometry and motion.
Throughout this paper, ``4D'' refers to geometry and motion evolving over time, rather
than full volumetric dynamic reconstruction.

Simulation provides structural supervision
$\mathcal{Y}^{\mathrm{sim}}_{\mathrm{structure}}=\{D,F,M,\mathcal{T}\}^{\mathrm{sim}}$,
covering depth, optical flow, masks, and trajectories. These signals encode
executable physical knowledge, while simulated RGB frames are used only through
constrained adaptation rather than as the generator's dominant fitting target.
Thus, Phys4D performs \textbf{structure-focused, appearance-constrained physical
knowledge injection}: it preserves the pretrained model's open-domain visual
prior while improving depth, motion, and trajectory consistency. Detailed
definitions of the lifted geometry-motion quantities are provided in
Appendix~\ref{app:problem_fomulation}.
\section{Scalable Physics Simulation for 4D Supervision}
\label{sec:simulation}

\begin{figure*}[htbp]
\centering
\includegraphics[width=1.0\textwidth]{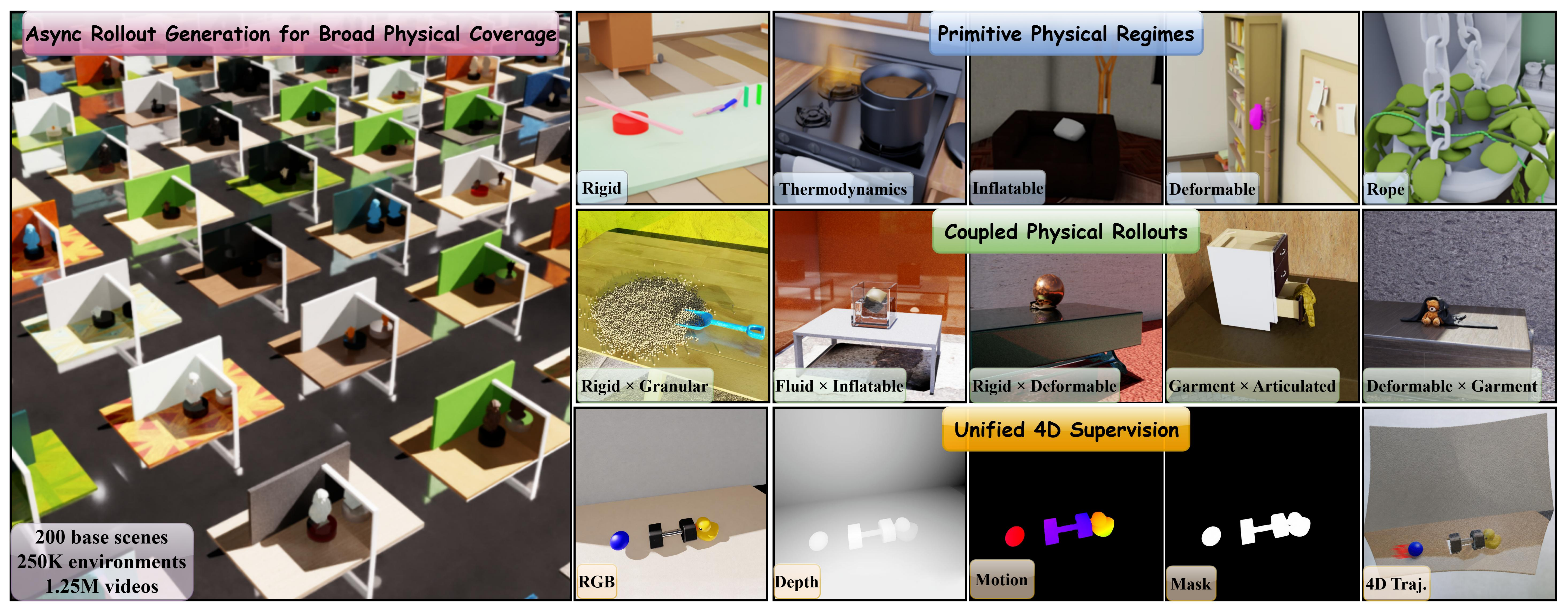}
\caption{
\textbf{Simulation as a scalable physical teacher.}
\textbf{Left:} Asynchronous rollouts expand 200 scenes into 250K environments and 1.25M videos.
\textbf{Top right:} The simulator covers primitive physical regimes.
\textbf{Middle right:} These regimes are composed into coupled physical rollouts.
\textbf{Bottom right:} Each rollout exports RGB, depth, motion, masks, and 4D point trajectories.
}
\label{fig:simulation_framework}
\end{figure*}
Real videos provide rich appearance priors but rarely expose dense 4D physical states such as depth, motion, and object masks. We therefore use simulation not as a target visual domain, but as an executable physical teacher that converts solvers and constraints into video-compatible structural supervision. To scale this supervision, we employ \textbf{asynchronous rollout generation}, where independent scenes run in parallel and resources are recycled upon completion. This covers randomized physical parameters, object configurations, and coupled interactions, producing RGB-D-motion-trajectory annotations for Stage-II local geometry-motion injection and Stage-III global 4D alignment while preserving the pretrained real-video prior.

\textbf{Unified coupled-physics framework.}
Based on Isaac Sim~\cite{xiang2024isaac} and GarmentLab~\cite{lu2024garmentlab,lu2024unigarment}, we integrate rigid-body dynamics, PBD, FEM, and task-specific solvers into a unified scene-level framework. It supports isolated regimes such as rigid objects, garments, fluids, deformables, ropes, granular materials, and thermodynamic state changes, as well as cross-regime coupling including rigid-fluid, rigid-garment, rigid-deformable, articulated-soft, and rigid-granular interactions. These coupled rollouts provide geometry, motion, visibility, and trajectory supervision for Phys4D. Detailed simulation settings are provided in Appendix~\ref{app:physics}.

\textbf{Domain randomization and 4D supervision.}
We randomize physical parameters, interaction layouts, geometry, and visual  factors to learn invariant geometry-motion regularities rather than renderer-specific shortcuts. Randomizing mass, friction, restitution, stiffness, gravity, and scale exposes regimes governed by shared physical laws, while varying layouts and coupled interactions prevents memorization of fixed motion templates. All rollouts are exported through a unified 4D supervision interface with RGB frames, depth, scene flow, dynamic masks, camera poses, and spatiotemporal point trajectories, abstracting solver-specific states into common geometry-motion-trajectory supervision. With asynchronous generation, 200 base scenes are expanded into approximately 250K environments; five 10-second camera rollouts per environment yield 1.25M annotated videos, about 3.5K hours, and over 15TB of multi-modal annotations.

\section{Method}
\begin{figure*}[htbp]
\centering
\includegraphics[width=\textwidth]{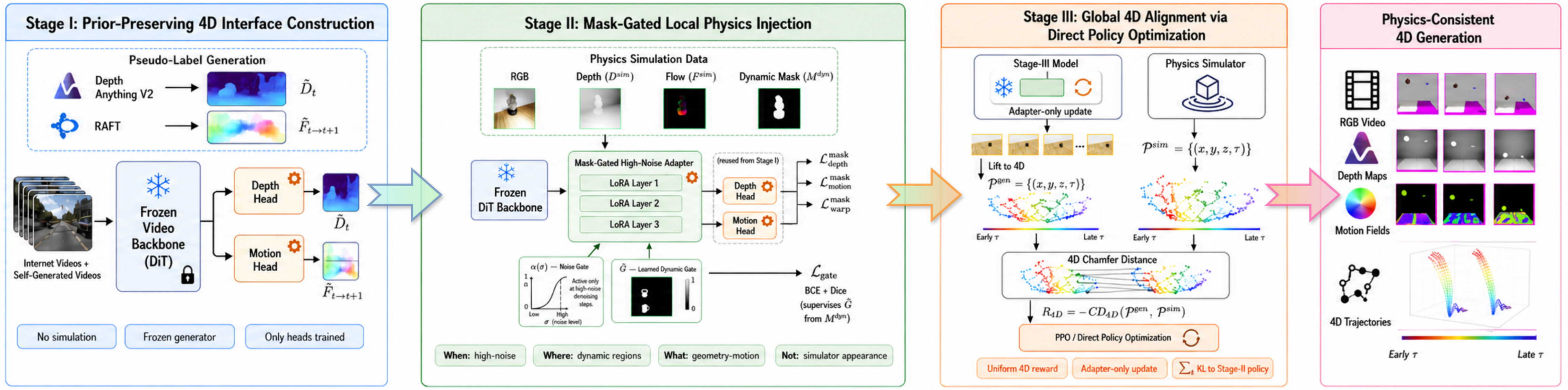}
\caption{
\textbf{Overview of the Phys4D training pipeline.}
\textbf{Stage I} freezes the video DiT and trains depth/motion heads to build an RGB-D-motion interface.
\textbf{Stage II} injects local physical structure through mask-gated high-noise adapters with simulation depth, motion, and warp supervision.
\textbf{Stage III} aligns generated 4D point trajectories with simulator trajectories using a 4D Chamfer reward.
}
\label{fig:pipeline}
\end{figure*}

Phys4D injects simulation-derived physical structure into a pretrained video diffusion model through a prior-preserving RGB-D-motion interface (Fig.~\ref{fig:pipeline}). 
It proceeds in three stages: Stage I (Sec.~\ref{method:pretrain}) freezes the video backbone and trains lightweight depth and motion heads; Stage II (Sec.~\ref{method:sft}) injects local geometry-motion structure through high-noise, mask-gated adapters; and Stage III (Sec.~\ref{method:rl}) aligns generated 4D trajectories with simulator trajectories using a trajectory-level reward.

\subsection{Stage I: Prior-Preserving 4D Interface Construction}
\label{method:pretrain}

Before physics injection, Stage I exposes geometry and motion through a prior-preserving RGB-D-motion interface. 
It uses no simulation data: the video backbone is frozen, and only lightweight depth and motion heads are trained.

\textbf{Model Architecture.}
We attach two auxiliary heads to a pretrained DiT-style video diffusion backbone.
The motion head predicts inter-frame optical flow $\hat F_{t\rightarrow t+1}$,
and the depth head predicts per-frame depth $\hat D_t$. Following
\cite{Chen2025VideoDA}, the depth head incorporates temporal attention and
multi-scale feature fusion for temporally coherent depth estimation. The DiT
backbone remains frozen throughout Stage I; only the auxiliary heads are
optimized. Details are provided in Appendix~\ref{app:impl_model}.

\textbf{Data Sources.}
We train the heads with pseudo labels on curated internet videos and self-generated videos from the pretrained model. 
Internet videos provide real-world diversity, while self-generated videos align the heads with the model's output distribution. 
Pseudo depth and flow are obtained from off-the-shelf estimators~\cite{Chen2025VideoDA,Teed2020RAFTRA}, with confidence masks and per-clip depth normalization for denoising; details are in Appendix~\ref{app:pretrain_data}.

\textbf{Training Objective.}
Stage I uses supervised regression to match predicted depth and flow to the
pseudo labels. The base video generation objective is unchanged, and gradients
are applied only to the auxiliary heads. For details please refer to Appendix ~\ref{app:stage1_loss}.
\subsection{Stage II: Mask-Gated Local Physics Injection}
\label{method:sft}
Stage I exposes an RGB-D-motion interface, but it does not teach physical
behavior. The key idea of Stage II is to inject simulation-derived physical
structure through a controlled pathway, rather than fine-tuning the generator to
imitate simulator appearance. We restrict simulation supervision along three
axes: \textbf{when to learn}, \textbf{where to learn}, and \textbf{what to
learn}. Specifically, simulation-induced updates are applied only through
lightweight physics adapters, activated at high-noise structure-forming denoising
steps, and concentrated on dynamic regions where physical changes occur. The
pretrained video backbone remains \textbf{frozen}; gradients from
simulation-derived losses update only the \textbf{physics adapters},
\textbf{learned gates}, and \textbf{auxiliary geometry-motion heads}.

\textbf{Where and when to inject: mask-gated physics adapter.}
We introduce a LoRA-style residual adapter as the \textbf{only trainable pathway}
through which simulation modifies the generator. Given frozen backbone features
$h_\ell$ at layer $\ell$, the adapted features are
\begin{equation}
h_\ell'
=
h_\ell
+
\alpha(\sigma)\,
\hat G_\ell
\odot
A_\ell(h_\ell),
\label{eq:gated_adapter}
\end{equation}
where $A_\ell$ is a lightweight physics adapter, $\sigma$ is the denoising noise
level, $\alpha(\sigma)$ is a noise gate, and $\hat G_\ell$ is a learned
dynamic-region gate. The noise gate controls \textbf{when} simulation can affect
generation: physics supervision is injected at high-noise steps that determine
coarse structure, layout, and motion, while low-noise appearance-refinement steps
remain dominated by the pretrained real-video prior. The dynamic gate controls
\textbf{where} the adapter acts: simulator object and dynamic masks supervise
$\hat G_\ell$ during training, but no simulator masks are required at inference.
Thus, simulation-induced adaptation is localized in \textbf{denoising time},
\textbf{image space}, and \textbf{parameter space}.

\textbf{What to learn: geometry, motion, and temporal consistency.}
Simulation provides depth, optical flow or projected motion, visibility, and
dynamic-region annotations. We use these signals to supervise \textbf{physical
structure while constraining appearance drift}. Dynamic masks are used as soft
weights, so the model preserves global scene geometry while emphasizing object,
deformation, and interaction regions. The depth head is supervised by
simulator depth, and the motion head is supervised by simulator flow or projected
motion, with visibility masks removing occluded or invalid correspondences. The
same dynamic annotations also train the gate to identify physics-relevant regions
from model features.

Per-frame depth and motion supervision alone do not guarantee that geometry
evolves coherently over time. We therefore include a \textbf{simulation-flow
guided geometry warp loss}:
\begin{equation}
\mathcal{L}_{geom\_warp}^{mask}
=
\sum_{t,u}
M_{t\rightarrow t+1}^{vis}(u)\,
w_t(u)\,
\left\|
\mathcal{W}
\left(
\hat D_t,
F_{t\rightarrow t+1}^{sim}
\right)(u)
-
\hat D_{t+1}(u)
\right\|_1 ,
\label{eq:mask_geom_warp}
\end{equation}
where $\mathcal{W}(\cdot)$ warps predicted depth according to simulator-provided
motion, $M_{t\rightarrow t+1}^{vis}$ masks invalid or occluded correspondences,
and $w_t(u)$ increases the weight of dynamic regions. This term teaches
\textbf{local geometry--motion coupling} under physically valid correspondences:
geometry at time $t$, transported by simulator motion, should agree with
geometry at time $t{+}1$.

We also include a weak \textbf{RGB warp self-consistency} term on generated
frames to stabilize temporal appearance. Importantly, this term does not use
simulator-rendered RGB as a photometric target; it only encourages adjacent
generated frames to be consistent under motion. Thus, Stage II learns local
depth, motion, and temporal geometry-motion consistency while avoiding direct
fitting to simulator RGB appearance.

\textbf{Overall objective.}
The Stage-II objective combines high-noise adapter training, mask-weighted
geometry-motion supervision, generated-frame RGB self-consistency, and gate
learning:
\begin{equation}
\mathcal{L}_{S2}
=
\mathcal{L}_{FM}^{high}
+
\lambda_R \mathcal{L}_{rgb\_warp}^{self}
+
\lambda_D \mathcal{L}_{depth}^{mask}
+
\lambda_F \mathcal{L}_{motion}^{mask}
+
\lambda_W \mathcal{L}_{geom\_warp}^{mask}
+
\lambda_G \mathcal{L}_{gate}.
\label{eq:stage2_objective}
\end{equation}
Here the high-noise flow-matching term $\mathcal{L}_{FM}^{high}$ uses simulator-conditioned RGB latents only through the adapter pathway to keep the adapted denoising trajectory on the video manifold. Table~\ref{tab:app_stage2_loss_summary} summarizes which Stage-II losses use simulator RGB, which use structural annotations, and which modules they update. Although simulation provides the
structural annotations used in Stage II, simulator RGB is not used as a direct
appearance target. All gradients from simulation-derived losses are blocked from
the frozen backbone and update only the lightweight adapters, learned gates, and
auxiliary heads. Stage II therefore teaches \textbf{local geometry-motion
physics} through a controlled structural pathway. Details are provided in
Appendix~\ref{app:stage2_losses}--\ref{app:stage2_objective}.
\subsection{Stage III: Simulation-Grounded 4D Alignment by Direct Policy Optimization}
\label{method:rl}
Stage II injects local geometry-motion consistency, but adjacent-frame losses do
not fully determine whether the completed generated evolution follows the
intended physical trajectory. Stage III therefore performs an offline
simulation-grounded trajectory alignment stage. Importantly, Stage III is
performed only on simulation-paired training scenes, where the simulator provides
the corresponding ground-truth 4D trajectory under the same initial state,
camera, physical parameters, and prompt. It is not an online reward applied to
arbitrary real images or open-domain prompts.

\textbf{Simulation-Paired Rollouts.}
For each Stage-III rollout, we sample a simulation scene with a known scene ID,
initial state, asset configuration, physical parameters, camera intrinsics and
extrinsics, and text prompt. The generated video is conditioned on the
simulator-rendered initial frame and the same text description. The corresponding
reference trajectory \(P_{\mathrm{sim}}\) is the simulator trajectory from this
same rollout. Thus, \(P_{\mathrm{sim}}\) is available only because the rollout is
paired with a simulator state, not because arbitrary generated videos have
ground-truth physical trajectories.

Given a generated RGB-D-motion sequence, we use the frozen Stage-I interface to
lift visible dynamic regions into a spatiotemporal point set, while the simulator
provides the paired reference point set from the same scene configuration and
rollout: $P_{\mathrm{gen}}=\{(x,y,z,\tau)\}$ and 
$P_{\mathrm{sim}}=\{(x,y,z,\tau)\}$.

Dynamic and visibility masks are used only to filter or weight active regions,
so that the alignment focuses on physical changes rather than static background.

\textbf{Simulation-Grounded 4D Alignment Reward.}
We use a unified metric form for trajectory alignment. For two 4D points
\(p=(x_p,\tau_p)\) and \(q=(x_q,\tau_q)\), we define:
\begin{equation}
d_{4D}(p,q)
=
\|x_p-x_q\|_2^2
+
\lambda_t |\tau_p-\tau_q|^2 .
\end{equation}
The generated and simulated trajectories are compared with a symmetric 4D
Chamfer distance:
\begin{equation}
CD_{4D}(P,Q)
=
\frac{1}{|P|}\sum_{p\in P}\min_{q\in Q}d_{4D}(p,q)
+
\frac{1}{|Q|}\sum_{q\in Q}\min_{p\in P}d_{4D}(p,q).
\end{equation}
The Stage-III reward is:
\begin{equation}
R_{4D}(V;s)=-CD_{4D}\!\left(P_{\mathrm{gen}}^{dyn},P_{\mathrm{sim}}^{dyn}\right)
\end{equation}
where $s$ denotes the paired simulation scene. The reward is
category-agnostic in its metric form, but it relies on simulator-paired
trajectories generated by category- or solver-specific physical simulators. It
should therefore be understood as a simulation-grounded trajectory alignment
reward, rather than a universal physical-law reward.

\textbf{Direct Policy Optimization.}
We formulate the stochastic reverse denoising process as a policy
\(\pi_{\theta+\phi}\), where \(\theta\) denotes the frozen pretrained video
backbone and \(\phi\) denotes the lightweight Stage-II LoRA adapters. Stage III
updates only these adapters:
\begin{equation}
\max_{\phi}
\;
\mathbb{E}_{s\sim \mathcal{S}_{sim},\, V\sim \pi_{\theta+\phi}(\cdot|I^{sim}_0,c_s)}
\left[
R_{4D}(V;s)
-
\beta
D_{\mathrm{KL}}
\left(
\pi_{\theta+\phi}
\|
\pi_{\theta+\phi_{S2}}
\right)
\right],
\label{eq:stage3_objective}
\end{equation}
where \(\phi_{S2}\) is the Stage-II checkpoint, \(I^{sim}_0\) is the
simulator-rendered initial frame, and \(c_s\) is the paired text description.

To avoid reward hacking, the depth head, motion head, and learned gate module are frozen during Stage III. Only the Stage-II LoRA adapters
are optimized. After Stage-III training, the learned adapters are used for
ordinary inference. No simulator trajectory, simulator depth, simulator mask, or
\(P_{\mathrm{sim}}\) is required at inference time. In this sense, Stage III
learns adapter parameters from simulation-paired trajectory alignment and
transfers the learned trajectory prior to open-domain generation.
\section{Experiments}
Our central hypothesis is that simulation can transfer physical structure into a pretrained video generator without replacing its real-video generative prior. In this section, we aim to answer the following research questions through our experiments:

\noindent\textbf{RQ1:} Does Phys4D improve observable physical plausibility in generated videos?

\noindent\textbf{RQ2:} Does Phys4D improve 4D world-level geometry-motion-trajectory consistency?

\noindent\textbf{RQ3:} Does Phys4D preserve the pretrained real-video generative prior?

\noindent\textbf{RQ4:} Which design choices are responsible for physical knowledge injection?

\subsection{Simulation-Derived Physics Transfers to Generated Videos}
\label{sec:rq1_video_physics}
\begin{figure*}[htbp]
    \centering
    \includegraphics[width=\textwidth]{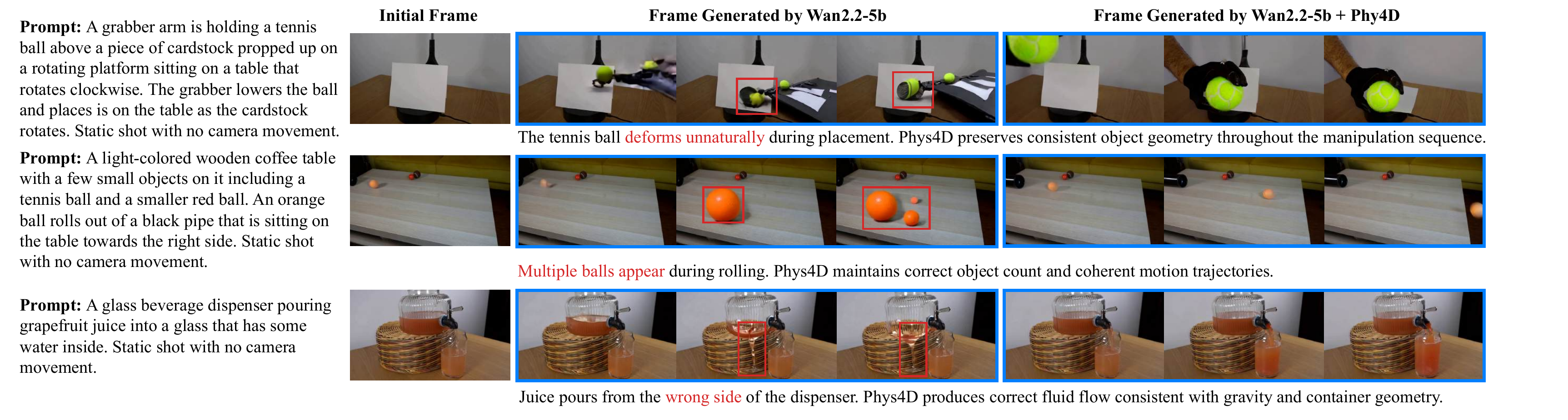}
    \caption{\textbf{ Qualitative comparison on Physics-IQ scenarios.} We compare Wan2.2-5B baseline (middle) with Wan2.2-5B + Phys4D (right) across three physical interactions; Phys4D improves geometry consistency, motion plausibility, and temporal stability, while the baseline shows shape distortion and incoherent physical behavior.}
    \label{fig:quality_demo}
\end{figure*}
\label{sec:experiments}

\textbf{Benchmarks and protocols.}  
We evaluate observable video-level physical plausibility on three external diagnostics:
Physics-IQ~\cite{motamed2025physicsiq}, VBench-2.0 Physics~\cite{vbench2025vbench2},
and PhyGenBench~\cite{phygenbench2025}. Following Sec.~\ref{sec:problem_definition},
we denote the generation condition as $c$. Physics-IQ follows a switch-frame I2V
protocol, where $c=I_{\mathrm{ref}}$ or $c=(I_{\mathrm{ref}},p)$ when a prompt is
available. VBench-2.0 Physics and PhyGenBench follow prompt-based T2V protocols, where
$c=p$.

\textbf{Implementation.}  
We apply Phys4D to CogVideoX-5B~\cite{yang2025cogvideoxtexttovideodiffusionmodels},
WAN2.2-5B~\cite{wan2025wanopenadvancedlargescale}, and
Open-Sora V1.2~\cite{peng2025opensora}. For Physics-IQ, we generate 5-second
continuations from the provided switch frame. For VBench-2.0 Physics and PhyGenBench,
we use their official prompt suites and evaluation pipelines. 

\begin{table}[htbp]
\centering
\setlength{\tabcolsep}{1.8pt}
\caption{\textbf{RQ1: External video-level physics evaluation.}
Phys4D improves observable physical plausibility across three backbones and three independent physics diagnostics. 
Physics-IQ is reported under our matched switch-frame I2V protocol.}
\label{tab:rq1_external_physics}
\resizebox{\linewidth}{!}{
\begin{tabular}{ll l cccc}
\toprule
\textbf{Backbone} & \textbf{Params} & \textbf{Method}
& \textbf{Physics-IQ MSE} $\downarrow$
& \textbf{Physics-IQ (Verified)} $\uparrow$
& \textbf{VBench2.0-Physics} $\uparrow$
& \textbf{Phygenbench} $\uparrow$ \\
\midrule
CogVideoX      
& 5B   
& Base     
& $0.013 \pm 0.002$ 
& $25.38 \pm 1.4$
& $0.4849 \pm 0.014$ 
& $0.45 \pm 0.02$ \\

CogVideoX      
& 5B   
& + Phys4D 
& $\mathbf{0.008 \pm 0.001}$ 
& $\mathbf{37.20 \pm 1.9}$
& $\mathbf{0.5954 \pm 0.018}$ 
& $\mathbf{0.58 \pm 0.03}$ \\

\midrule
WAN2.2         
& 5B   
& Base     
& $0.016 \pm 0.002$ 
& $21.33 \pm 1.3$
& $0.4928 \pm 0.015$ 
& $0.45 \pm 0.02$ \\

WAN2.2         
& 5B   
& + Phys4D 
& $\mathbf{0.013 \pm 0.001}$ 
& $\mathbf{35.80 \pm 1.8}$
& $\mathbf{0.6247 \pm 0.017}$ 
& $\mathbf{0.57 \pm 0.03}$ \\

\midrule
Open-Sora V1.2 
& 1.1B 
& Base     
& $0.021 \pm 0.003$ 
& $14.5 \pm 1.2$ 
& $0.4623 \pm 0.016$ 
& $0.44 \pm 0.02$ \\

Open-Sora V1.2 
& 1.1B 
& + Phys4D 
& $\mathbf{0.014 \pm 0.002}$ 
& $\mathbf{24.7 \pm 1.6}$ 
& $\mathbf{0.5899 \pm 0.019}$ 
& $\mathbf{0.57 \pm 0.03}$ \\
\bottomrule
\end{tabular}
}
\end{table}

\textbf{Results.}  
Table~\ref{tab:rq1_external_physics} shows consistent gains from Phys4D across three backbones and three external physics diagnostics.
This suggests that simulation-derived physical structure transfers to generated videos beyond simulator-specific metrics.
Additional analysis is provided in Appendix~\ref{app:experiment_physics}.

\subsection{RQ2: Phys4D Improves 4D Geometry--Motion--Trajectory Consistency}
\label{sec:rq2_4d_diagnostics}

\textbf{Diagnostics.}
We evaluate whether the video-level gains in RQ1 are reflected in the RGB-D-motion interface.
Since real videos lack dense geometry and trajectory ground truth, we use held-out simulated scenes with ground-truth depth, flow, and object trajectories.
Our diagnostics cover four levels: per-frame geometry, local geometry-motion consistency, global trajectory evolution, and novel-time continuity.
We report AbsRel for geometry, Depth Warp L1 and Flow EPE for local consistency, 4D Chamfer, Mean Drift, and Fail Rate for trajectory evolution, and novel-time depth/warp errors for interpolation.
Full metrics are provided in Appendix~\ref{app:rq2_4d_diagnostics}.

\textbf{Implementation.}
For RGB-only baselines, we lift generated videos using off-the-shelf depth and motion estimators, DAV2~\cite{depth_anything_v2} and SEA-RAFT~\cite{wang2024sea}.
Phys4D is evaluated with its learned RGB-D-motion interface.
All methods use the same held-out simulation scenes and camera settings, and are compared against simulation ground truth.

\begin{table*}[t]
\centering
\setlength{\tabcolsep}{3.5pt}
\caption{\textbf{RQ2: Representative 4D interface level diagnostics.}
We report key metrics for geometry, local geometry-motion consistency, global trajectory evolution, and novel-time continuity.
For RGB-only baselines, \textbf{OTF} denotes an off-the-shelf diagnostic pipeline that derives depth and motion cues from pretrained estimators, instantiated with Depth Anything V2~\cite{depth_anything_v2} and SEA-RAFT~\cite{wang2024sea}.
Phys4D improves most 4D diagnostics, showing that its gains go beyond RGB-level physics scores and enter the RGB-D-motion world interface.}
\label{tab:rq2_4d_diagnostics}
\resizebox{\textwidth}{!}{
\begin{tabular}{llcccccccc}
\toprule
\textbf{Backbone} 
& \textbf{Method} 
& \textbf{AbsRel} $\downarrow$
& \textbf{Warp L1} $\downarrow$
& \textbf{Flow EPE} $\downarrow$
& \textbf{4D Chamfer} $\downarrow$
& \textbf{Mean Drift} $\downarrow$
& \textbf{Fail Rate} $\downarrow$
& \textbf{Novel Depth} $\downarrow$
& \textbf{Novel Warp} $\downarrow$ \\
\midrule
WAN2.2 
& Base + OTF
& 0.3929 
& 0.7990 
& 1.2516 
& 0.5058 
& 0.5369 
& 12.38\% 
& 0.5841 
& 1.1076 \\
CogVideoX 
& Base + OTF
& 0.3483 
& 0.7054 
& 1.2343 
& 0.4923 
& 0.5239 
& 11.34\% 
& 0.5432 
& 1.1254 \\
Open-Sora V1.2 
& Base + OTF
& 0.4216
& 0.7436 
& 1.2718 
& 0.5286 
& 0.5637 
& 13.54\% 
& 0.6534 
& 1.1573 \\
\midrule
Phys4D + WAN2.2
& RGB+OTF
& 0.3347
& 0.6698
& 1.0824
& 0.4873
& 0.5126
& 10.91\%
& 0.5527
& 1.0954 \\
Phys4D + WAN2.2
& Ours
& \textbf{0.2386} 
& \textbf{0.4558} 
& \textbf{0.4478} 
& \textbf{0.4159} 
& \textbf{0.4312} 
& \textbf{8.36\%} 
& \textbf{0.4746} 
& \textbf{1.0289} \\
\bottomrule
\end{tabular}
}
\end{table*}

\begin{figure*}[htbp]
    \centering
    \includegraphics[width=\textwidth]{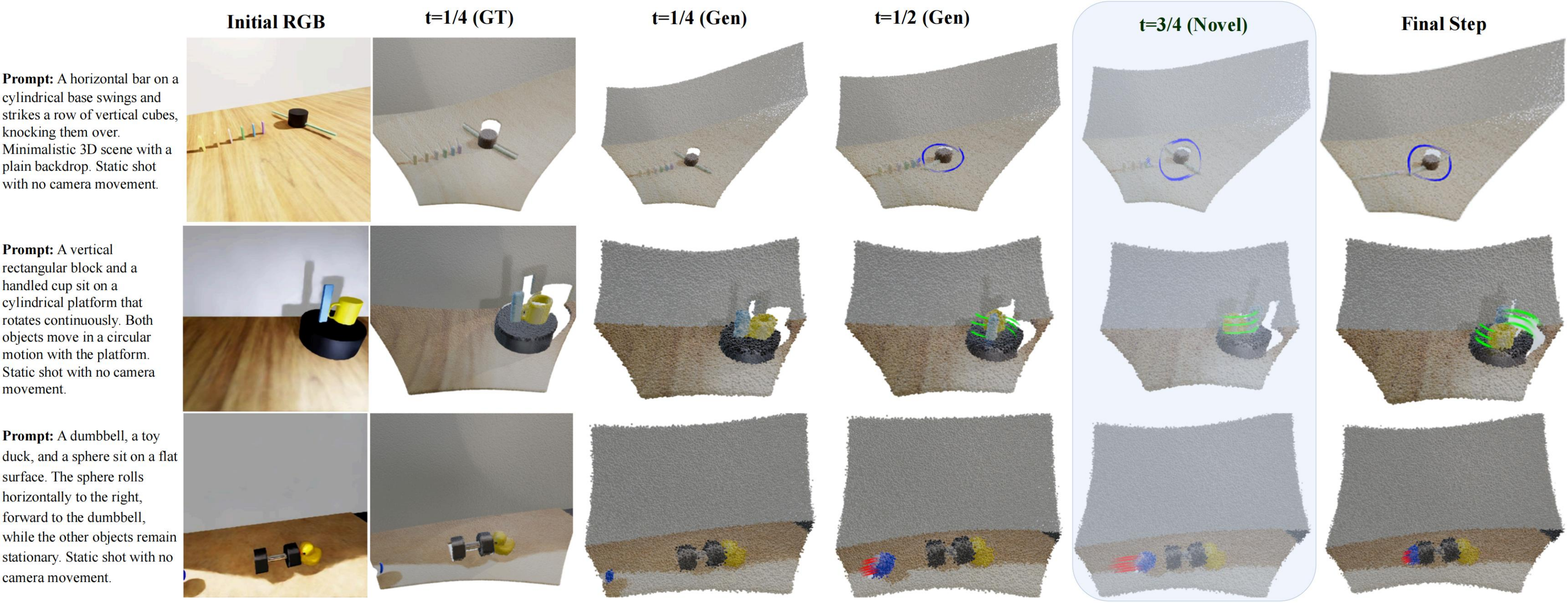}
    \caption{\textbf{Qualitative visualization of 4D interface-level diagnostics.}
    From left to right: ground-truth 4D point cloud; generated 4D point cloud
    at $1/4$ of the sequence; generated 4D point cloud at $1/2$ of the sequence;
    novel-time generated 4D point cloud at $3/4$ of the sequence; and generated
    4D point cloud at the final frame. This visualization illustrates whether
    the lifted RGB-D-motion representation maintains coherent geometry and
    object motion over the generated sequence.}
    \label{fig:app_4d_diagnostics}
\end{figure*}

\textbf{Results.}  
Table~\ref{tab:rq2_4d_diagnostics} shows that Phys4D improves most 4D diagnostics, including geometry, motion consistency, trajectory evolution, and novel-time continuity.
This indicates that Phys4D improves the underlying RGB-D-motion world interface beyond RGB-level plausibility.
Figure~\ref{fig:app_4d_diagnostics} illustrates this qualitatively: across the generated sequence, the lifted RGB-D-motion representation maintains coherent object geometry and motion that stays aligned with the ground-truth 4D point cloud, including at the novel (unobserved) timestamp.
Additional analysis is provided in Appendix~\ref{app:rq2_4d_diagnostics}.

\subsection{RQ3: Phys4D Preserves the Real-Video Visual Prior}
\label{sec:rq3_prior_preservation}
\textbf{Evaluation and implementation.}  
We evaluate whether simulation-derived physics injection harms the pretrained real-video visual prior using VBench~\cite{huang2024vbench}.
For each backbone, we compare the base model and its Phys4D-enhanced counterpart under the same prompts, sampling settings, and evaluation pipeline.
We report \textit{Aesthetic Quality}, \textit{Imaging Quality}, \textit{Color}, \textit{Appearance Style}, and \textit{Human Action}, and define \textbf{Visual Avg.} as the average of the first three metrics.
\begin{table*}[htbp]
\centering
\setlength{\tabcolsep}{3.2pt}
\caption{\textbf{RQ3: VBench visual prior preservation.}
We evaluate whether Phys4D preserves the pretrained real-video visual prior using VBench 1.0.
\textbf{Visual Avg.} averages Aesthetic Quality, Imaging Quality, and Color.
Phys4D keeps visual quality close to the base model while maintaining or improving Appearance Style and Human Action.}
\label{tab:rq3_vbench_prior_preservation}
\resizebox{\textwidth}{!}{
\begin{tabular}{ll l ccc c cc}
\toprule
\textbf{Backbone} 
& \textbf{Params} 
& \textbf{Method}
& \makecell{\textbf{Aesthetic}\\\textbf{Quality} $\uparrow$}
& \makecell{\textbf{Imaging}\\\textbf{Quality} $\uparrow$}
& \textbf{Color} $\uparrow$
& \makecell{\textbf{Visual}\\\textbf{Avg.} $\uparrow$}
& \makecell{\textbf{Appearance}\\\textbf{Style} $\uparrow$}
& \makecell{\textbf{Human}\\\textbf{Action} $\uparrow$} \\
\midrule
CogVideoX      
& 5B   
& Base     
& 0.6124
& 0.6931
& 0.9018
& 0.7358
& 0.2107
& 0.7452 \\
CogVideoX      
& 5B   
& + Phys4D 
& 0.6097
& 0.6886
& 0.8995
& 0.7326
& \textbf{0.2184}
& \textbf{0.7728} \\
\midrule
Open-Sora V1.2 
& 1.1B 
& Base     
& 0.5948
& 0.6715
& 0.8842
& 0.7168
& 0.1972
& 0.6872 \\
Open-Sora V1.2 
& 1.1B 
& + Phys4D 
& 0.5917
& 0.6782
& 0.8821
& 0.7173
& \textbf{0.2011}
& \textbf{0.7023} \\
\midrule
WAN2.2         
& 5B   
& Base     
& 0.6286
& 0.7164
& 0.9112
& 0.7521
& 0.2139
& 0.7433 \\
WAN2.2         
& 5B   
& + Phys4D 
& \textbf{0.6351}
& 0.7119
& 0.9087
& 0.7519
& \textbf{0.2162}
& \textbf{0.7712} \\
\bottomrule
\end{tabular}
}
\end{table*}

\textbf{Results.}  
Table~\ref{tab:rq3_vbench_prior_preservation} shows that Phys4D keeps Visual Avg. nearly unchanged across all three backbones, while consistently maintaining or improving Appearance Style and Human Action.
This indicates that Phys4D preserves open-domain visual quality and transfers simulation mainly as physical structure rather than simulator appearance. For more details, please refer to \ref{app:experiment_prior_preservation}

\subsection{RQ4: Structure-Only Local and Global Injection Makes Phys4D Work}
\label{sec:rq4_ablation}
\textbf{Setup.}  
We ablate Phys4D on WAN2.2-5B using Physics-IQ under the same switch-frame I2V protocol as RQ1.
Stage~I is not treated as a separate RGB-generation ablation because it freezes the video generator and only learns the RGB-D-motion interface.
Thus, RQ4 focuses on whether Stage~II/III inject physical structure through local geometry-motion supervision and global trajectory-level alignment. Full ablations and implementation details are provided in Appendix~\ref{app:rq4_ablation}.

\textbf{Stage-wise ablation: local injection enables global alignment.}
We isolate the contribution of each training stage (Table~\ref{tab:rq4_stage_ablation}).
Stage~III alone gives only a modest gain over the Original-protocol base (16.8$\rightarrow$18.4), whereas Stage~II provides the main local structure injection (21.33$\rightarrow$32.20) and Stage~III built on top of it corrects residual global trajectory violations (32.20$\rightarrow$35.80).
This supports the local-to-global design: local geometry-motion injection initializes the interface, and trajectory-level alignment refines global consistency.

\begin{table}[htbp]
\centering
\small
\setlength{\tabcolsep}{6pt}
\caption{\textbf{Stage-wise ablation on WAN2.2-5B.}
Direct Stage~III-only optimization provides only a small gain over the base model, while Stage~II provides a much larger improvement and enables Stage~III to further improve trajectory-level physical consistency.}
\label{tab:rq4_stage_ablation}
\begin{tabular}{lcc}
\toprule
\textbf{Configuration} & \textbf{Physics-IQ} $\uparrow$ & \textbf{$\Delta$ vs. Base} \\
\midrule
Base WAN2.2-5B & 21.33 & -- \\
+ Stage~III only w/o Stage~II & 18.4 & +1.6 \\
+ Stage~II & 32.20 & +10.87 \\
+ Stage~II + Stage~III & \textbf{35.80} & \textbf{+14.47} \\
\bottomrule
\end{tabular}
\end{table}

\textbf{Simulator RGB negative control: transfer structure, not appearance.}
We test whether Phys4D improves physics simply by fitting simulator RGB (Table~\ref{tab:rq4_sim_rgb_negative_control}).
Full simulator RGB LoRA fine-tuning improves physics (16.8$\rightarrow$24.2 Physics-IQ) but degrades the prior (0.586$\rightarrow$0.528) and FVD (140.0$\rightarrow$168.5), indicating appearance contamination.
Restricting RGB matching to high-noise timesteps better preserves the prior, and full Phys4D attains the best physics scores while keeping the prior close to the base model and improving FVD to 121.5.
This supports transferring simulation as geometry, motion, and trajectory structure rather than simulator appearance.

\begin{table*}[htbp]
\centering
\small
\setlength{\tabcolsep}{4pt}
\caption{\textbf{Simulator RGB fine-tuning negative control.}
Naive simulator RGB fine-tuning improves physical scores but degrades prior-related VBench-2.0 dimensions and visual quality.
Restricting simulator supervision to high-noise structure-forming timesteps improves prior preservation, and adding structural losses gives the best overall trade-off.}
\label{tab:rq4_sim_rgb_negative_control}
\resizebox{\textwidth}{!}{
\begin{tabular}{lccccc p{0.23\textwidth}}
\toprule
\textbf{Variant}
& \textbf{Physics-IQ} $\uparrow$
& \textbf{VB2-Physics} $\uparrow$
& \textbf{PhyGenBench} $\uparrow$
& \makecell{\textbf{VB2 Prior}\\\textbf{Avg.} $\uparrow$}
& \textbf{FVD} $\downarrow$
& \textbf{Interpretation} \\
\midrule
Base WAN2.2-5B
& 16.8
& 0.4928
& 0.45
& \textbf{0.586}
& 140.0
& pretrained real-video prior \\

Full simulator RGB LoRA fine-tuning
& 24.2
& 0.5481
& 0.50
& 0.528
& 168.5
& physics improves, prior degrades \\

Low-noise simulator RGB FM
& 22.9
& 0.5336
& 0.49
& 0.542
& 160.2
& appearance-stage updates damage the prior \\

Full-interval simulator RGB FM
& 26.4
& 0.5624
& 0.51
& 0.551
& 153.7
& stronger physics but visible appearance drift \\

High-noise $\mathcal{L}_{\mathrm{FM}}^{\mathrm{RGB}}$ only
& 27.2
& 0.5798
& 0.52
& 0.574
& 143.8
& structure-stage update better preserves the prior \\

Full Phys4D
& \textbf{30.9}
& \textbf{0.6247}
& \textbf{0.57}
& 0.581
& \textbf{121.5}
& high-noise RGB matching plus structural losses \\
\bottomrule
\end{tabular}
}
\end{table*}

\textbf{Component ablations.}
Three further ablations (full results in Appendix~\ref{app:rq4_ablation}) support the remaining design choices.
The gains are not explained by auxiliary heads alone, since removing depth and flow still improves over the base model (16.8$\rightarrow$21.4), and depth and flow are complementary (27.8 jointly).
High-noise adaptation is consistently better than full-interval or low-noise updates (27.8 vs. 24.6 vs. 22.9), supporting structure-stage physics injection.
Finally, removing geometry, motion, or warp supervision degrades performance (30.9$\rightarrow$23.8/24.6/25.9), confirming that Phys4D relies on coupled geometry-motion learning rather than a single loss term.

\section{Conclusion}
We presented Phys4D, a three-stage framework that augments pretrained video diffusion models
with a physics-aware RGB-D-motion interface.
Our results highlight the importance of explicit 4D interface training and evaluation
for advancing generative world models.
Additional discussion of limitations, broader impact, and responsible release is provided in Appendix~\ref{app:limitations_impact}.

\newpage
\bibliographystyle{plainnat}
\bibliography{main}

\newpage
\appendix
\begin{center}
{\LARGE\bfseries Appendix}\par
\vspace{1em}
\end{center}
\addcontentsline{toc}{section}{Appendix}
\suppressfloats[t] 
\section*{Appendix Overview}

This appendix provides supplementary materials that support the main paper and improve clarity, completeness, and reproducibility. Specifically, it expands on data construction, training details, extended experiments, qualitative analyses, theoretical details, discussion of limitations and impact, and physics-based simulation settings that are not fully described in the main text due to space constraints. To further support reproducibility and future research, we will release the code, datasets, introduction slides, and video demos through the project page: https://sensational-brioche-7657e7.netlify.app/
\paragraph{Reproducibility, assets, and release.} The supplementary material documents our data construction pipeline, benchmark protocol, model components, and stage-wise training procedure. For anonymous submission, we provide an anonymized project page for paper resources and plan to release code, benchmark assets, and representative data artifacts under a CC BY 4.0 license upon acceptance, together with documentation for reproduction and responsible use.

\paragraph{Training summary.} Detailed stage-wise training protocol, compute cost, and inference overhead are collected in Appendix~D.
\begin{figure*}[htbp]
    \centering
    \includegraphics[width=\textwidth]{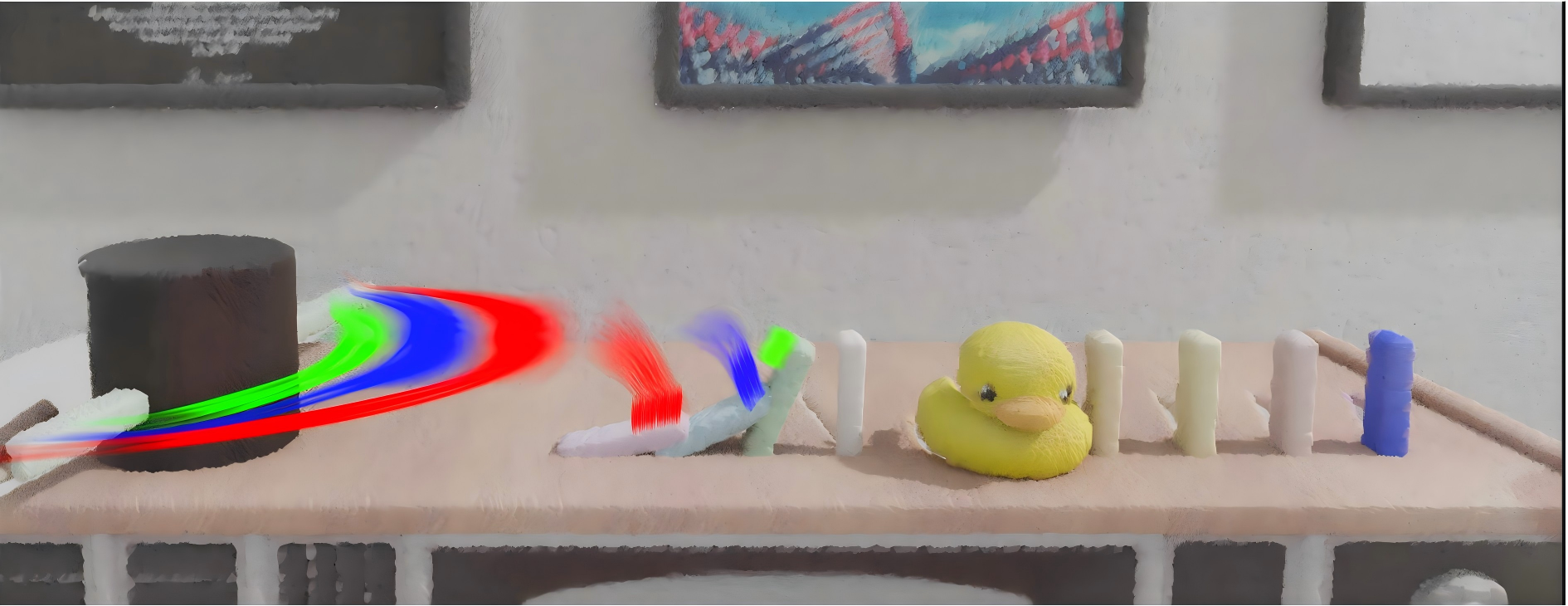}
    \caption{\textbf{Qualitative result of a Phys4D-generated scene.}}
    \label{app_fig:quality_demo}
\end{figure*}

\begingroup
\setlength{\parskip}{0em}
\paragraph{Appendix A: Data.} 
We describe the data sources used throughout training, including internet videos, model-generated samples, and simulation-generated data. This section details data preprocessing, filtering strategies, and annotation procedures for depth and motion supervision, clarifying how diverse supervision signals are constructed across different stages of training.
\paragraph{Appendix B: Detailed Problem Formulation.}
This section expands the RGB-D-motion interface used throughout the paper, clarifying how depth, optical flow, lifted 3D motion, and trajectories define a video-compatible representation for physics-consistent world modeling.

\paragraph{Appendix C: Implementation}
This appendix provides implementation details for the proposed \textbf{Phys4D} framework,
covering data preparation, model training, and reinforcement learning fine-tuning.
Our implementation follows the three-stage training paradigm described in the main paper
and is designed to progressively augment video diffusion models from appearance-level generation
with a physics-aware \textbf{RGB-D-motion interface}.

\paragraph{Appendix D: Training Details.}
This section summarizes the stage-wise training protocol, hardware footprint, and the measured inference overhead introduced by Phys4D, consolidating reproducibility details that were previously scattered across the supplementary material.

\paragraph{Appendix E: Qualitative Results.}
We present additional qualitative results to complement the quantitative evaluations in the main paper. This section includes visual demonstrations of generated videos, depth maps, and motion fields, highlighting improvements in fine-grained physical consistency, temporal coherence, and sequence-level trajectory coherence.

\paragraph{Appendix F: Experiments.}
We report extended experimental results and ablation studies that are omitted from the main paper for brevity. These experiments further analyze the effects of different training stages, supervision signals, and design choices, providing deeper insight into the behavior of the proposed method.

\paragraph{Appendix G: Additional Ablation Studies.}
This section collects supplementary ablations that further isolate the effect of individual design choices, supervision signals, and training stages beyond the main experimental section.

\paragraph{Appendix H: Additional 4D World-Level Diagnostics.}
This section presents extended RQ2 diagnostics on the RGB-D-motion world interface, including per-frame geometry, local temporal geometry-motion consistency, and sequence-level 4D evolution. It is framed as a diagnostic evaluation on held-out simulated scenes rather than as a new benchmark contribution.

\paragraph{Appendix I: Preliminaries and Theory.}
We include mathematical preliminaries and theoretical details underlying our approach, including flow matching, deterministic ODE sampling, and the stochastic formulations used for reinforcement learning fine-tuning. This section contains derivations and clarifications that support the methodology presented in the main paper.

\paragraph{Appendix J: Additional Related Work.}
We discuss related work that could not be fully covered in the main paper, providing further context on physics-aware video generation, 4D modeling, and learning-based world representations.

\paragraph{Appendix K: Physics Simulation.}
We describe the physics simulation settings used for supervision and reward construction, including scene configurations, object properties, and motion extraction procedures. This section explains how simulation is leveraged as a source of high-fidelity geometric and motion information for fine-grained physical alignment.

\paragraph{Appendix L: Limitations and Impact Statement.}
This section consolidates the paper's limitations, broader impact, and responsible release stance into a dedicated discussion chapter in the appendix.

\endgroup

\section{Data}
\subsection{Simulation Data}
\label{app:sim_data}

To enable large-scale 4D data collection at high throughput and low cost,
we build our simulation pipeline on an \emph{asynchronous parallel execution} framework.
Our system collects data from a large number of simulation sub-environments concurrently,
significantly improving GPU utilization and simulation throughput compared to sequential
or synchronously parallel simulators.
This design allows us to generate large-scale, diverse simulation data
at substantially higher speed and lower computational cost.

\paragraph{Tiled Rendering for Efficient Visual Simulation.}
To further improve rendering efficiency, we adopt \emph{tiled camera rendering},
where multiple camera views from different simulation sub-environments
are rendered within a single large framebuffer.
This approach amortizes rendering overhead across environments
and avoids redundant GPU context switching.
We follow the tiled rendering strategy provided by Isaac Lab~\cite{isaaclab_tiled},
which enables scalable visual data generation with minimal performance degradation
as the number of parallel environments increases.

\paragraph{Asynchronous Execution.}
While parallelism is critical for simulator efficiency,
naïve synchronous execution leads to significant idle time when sub-environments
exhibit heterogeneous dynamics.
In vision-based control and data collection, a policy typically performs inference once
to generate a target pose or waypoint, after which the robot executes a trajectory
of variable length.
Synchronizing such trajectories across environments results in unnecessary blocking
and under-utilization of computational resources.

To address this challenge, our simulator supports \emph{fully asynchronous execution}.
Each policy-driven motion is encapsulated as an atomic action with a well-defined interface,
allowing high-level policies to invoke and monitor actions independently across environments.
This design eliminates global synchronization barriers and maximizes hardware utilization,
enabling efficient closed-loop data collection under heterogeneous execution.

\paragraph{Data Management and Asynchronous Control.}
Supporting asynchronous execution is non-trivial, as the underlying simulator
operates in a SIMD (Single Instruction, Multiple Data) paradigm,
whereas asynchronous execution requires MIMD (Multiple Instruction, Multiple Data) semantics.
To bridge this gap, we introduce a motion-planning-based abstraction layer that
translates high-level asynchronous commands into low-level, time-continuous trajectories.
This abstraction allows each sub-environment to progress independently
while preserving simulator stability.

\paragraph{Scene Management and Heterogeneous Environments.}
To support heterogeneous parallel simulation, different sub-environments
may require distinct scene configurations.
We address this by introducing a \emph{Layout Manager} that programmatically
controls asset placement across environments.
The Layout Manager imports user-specified objects with configurable randomization,
enabling diverse, controllable, and reproducible scene layouts.
This design allows us to efficiently scale data collection across
a wide range of scene configurations within a single simulator instance.

\paragraph{Multi-Modal Camera Strategies.}
We employ two complementary camera strategies to balance static scene coverage and dynamic viewpoint variation. \textbf{Fixed multi-view cameras} provide dense spatial coverage by placing multiple static cameras around each scene, capturing aligned RGB images together with calibrated intrinsics and extrinsics and dense geometric annotations such as depth and optical flow. In parallel, \textbf{dynamic trajectory cameras} simulate realistic cinematography by sampling continuous camera motions (e.g., orbit, dolly, crane, tracking, and first-person) along parameterized trajectories, producing frame-level pose annotations for training trajectory-aware video generation models.

\paragraph{Simulation Throughput and Data Scale.}
Our data collection pipeline is optimized for high-throughput simulation.
On a single NVIDIA RTX 4090 GPU with 24\,GB memory, we run up to 1{,}024 simulation
environments in parallel, enabling efficient large-scale data generation at relatively
low computational cost.
Each environment is simulated for 10 seconds of effective interaction time.

We construct a base set of 200 simulation scenes and expand it to approximately
250{,}000 distinct environments through domain randomization over initial object
configurations, physical parameters, and surface textures.
For each environment, we collect five camera trajectories, including three fixed-view
cameras and two dynamic camera motions, providing diverse spatial and temporal
observations of the same underlying physical scene.
This yields approximately 1.25M annotated 10-second videos, corresponding to
about 3.5K hours of simulation video, together with over 15\,TB of multi-modal
annotations.

\subsection{Internet Video Pretraining Data}
\label{sec:app_internet_data}

\paragraph{Motivation.}
To improve the robustness and generalization of the depth and motion heads,
we leverage large-scale real-world videos for pseudo-supervised pretraining.
Compared to simulation data, Internet videos exhibit substantially greater
diversity in scene layout, appearance, and motion patterns, which is critical
for learning domain-agnostic geometry and motion representations.
This pretraining stage equips the model with strong visual and geometric priors
before introducing physics-grounded supervision.

\paragraph{Data Sources.}
We assemble the pretraining corpus using a hybrid collection strategy that
combines automated web crawling with selective manual curation.
Videos are sourced from multiple public platforms, including YouTube, Pexels,
Pixabay, Mixkit, and Bilibili, ensuring broad coverage of real-world environments
and camera motions.
We retain only videos with permissive licenses when license metadata is available,
exclude private or personal content, and provide a removal mechanism for released metadata.

\paragraph{Automated Quality Filtering.}
To maintain data quality at scale, we apply a multi-stage automated filtering
pipeline.
We first use the LAION aesthetic scorer~\cite{schuhmann2022laion} to remove visually
low-quality or noisy clips.
Next, we employ an OCR-based text detector based on PaddleOCR
to identify and discard videos containing prominent overlaid text, watermarks,
or subtitles that may interfere with depth and motion estimation.
Finally, videos that are extremely short, corrupted, or visually inconsistent
are automatically removed.
This filtering process yields a large-scale, diverse, and high-quality video
dataset suitable for robust geometry and motion pretraining.

\paragraph{Final Dataset Curation.}
Starting from an initial pool of approximately 100{,}000 candidate video clips,
our automated filtering pipeline retains about 5{,}000 high-quality videos
that satisfy our fine-grained requirements.
These selected clips exhibit sufficient motion diversity, clear object boundaries,
and stable visual quality, making them suitable for robust depth and motion
pretraining.
This curated subset forms the final Internet video dataset used in our
pseudo-supervised pretraining stage.

\label{app:pretrain_data}

\section{Detailed Problem Formulation}
\label{app:problem_fomulation}

In this section, we provide additional details for the RGB-D-motion formulation used in
Sec.~\ref{sec:problem_definition}. The goal is not to define a complete dynamic 3D
reconstruction problem, but to specify a video-compatible interface through which
physical structure can be exposed, supervised, and evaluated.

\subsection{RGB-D-motion Interface}

Given the input image $I_0$ and an optional text prompt $p$, the model generates RGB
frames $\{I_t\}_{t=0}^{T}$ together with geometry and motion variables:
\[
\{D_t\}_{t=0}^{T}, \qquad
\{F_{t\rightarrow t+1}\}_{t=0}^{T-1}.
\]
Here $D_t \in \mathbb{R}^{H\times W}$ denotes a depth map in the camera coordinate
system, and $F_{t\rightarrow t+1}\in \mathbb{R}^{H\times W\times 2}$ denotes optical
flow from frame $t$ to frame $t+1$. These variables form the minimal interface used by
Phys4D: RGB frames provide visual observations, depth exposes per-frame geometry, and
flow exposes inter-frame motion.

Throughout the paper, we use ``4D'' to refer to geometry and motion evolving over time.
This is weaker than full dynamic volumetric reconstruction, but stronger than purely
2D video generation. It provides a tractable representation that is compatible with
video diffusion models and can be supervised using both pseudo-labels from real videos
and structural annotations from simulation.

\subsection{Lifting Depth and Flow to 3D Motion}

The RGB-D-motion interface can be lifted into camera-space 3D motion when camera
intrinsics are available or estimated. Let $u=(x,y)$ be a pixel coordinate and
$\tilde{u}=(x,y,1)^\top$ its homogeneous coordinate. Given camera intrinsics $K$, the
corresponding 3D point at time $t$ is
\[
X_t(u) = D_t(u) K^{-1}\tilde{u}.
\]
The optical flow predicts a correspondence
\[
u' = u + F_{t\rightarrow t+1}(u).
\]
Using the depth at the target location, the corresponding point at time $t+1$ is
\[
X_{t+1}(u') = D_{t+1}(u')K^{-1}\tilde{u}'.
\]
The induced scene flow is therefore
\[
S_{t\rightarrow t+1}(u)
=
X_{t+1}(u') - X_t(u).
\]
This relation couples depth and 2D motion into a single 3D displacement field. In
practice, this lifting is only evaluated on valid correspondences, excluding occluded
pixels, out-of-bound flow endpoints, and regions with unreliable depth or motion.

\subsection{Trajectory Representation}

Beyond adjacent-frame motion, physical consistency often depends on longer-range
evolution. We therefore define lifted trajectories by recursively composing the
predicted flow and depth over time. Starting from a pixel $u_0$ in frame $0$, its image
trajectory is
\[
u_{t+1} = u_t + F_{t\rightarrow t+1}(u_t),
\]
and its lifted 3D trajectory is
\[
\tau(u_0) =
\left\{
X_t(u_t)
\right\}_{t=0}^{T},
\qquad
X_t(u_t)=D_t(u_t)K^{-1}\tilde{u}_t .
\]
The collection of valid lifted trajectories is denoted as
\[
\mathcal{T}=\{\tau(u_0)\}_{u_0\in\Omega_{\mathrm{valid}}},
\]
where $\Omega_{\mathrm{valid}}$ contains pixels whose trajectories remain visible and
geometrically valid for the considered time horizon. These trajectories summarize
non-local scene evolution and are used to reason about physical behavior that cannot be
fully captured by per-frame depth or adjacent-frame flow alone.

\subsection{Simulation-derived Structural Supervision}

The coupled-physics simulator provides temporally aligned structural annotations
\[
\mathcal{Y}^{\mathrm{sim}}_{\mathrm{structure}}
=
\{D, F, S, M, V, C, \mathcal{T}\}^{\mathrm{sim}},
\]
where $D$ is depth, $F$ is optical flow, $S$ is scene flow, $M$ is a dynamic or object
mask, $V$ is visibility, $C$ is contact information, and $\mathcal{T}$ denotes lifted
trajectories. These annotations are exported from simulation because they are
structural and physically meaningful. In contrast, rendered RGB frames from simulation
are treated as domain-specific observations and are not used as the primary appearance
target for training the video generator.

This distinction is important. The simulator provides executable physical knowledge:
geometry, motion, contact, deformation, coupling, and visibility. The pretrained video
diffusion model provides the real-video visual prior. Phys4D transfers the former into
the latter through structural variables under appearance-constrained adaptation, rather
than by fully steering the generator toward simulator-specific appearance statistics.

\subsection{Geometry-motion Consistency}

The predicted depth and motion should be mutually consistent over time. Given optical
flow $F_{t\rightarrow t+1}$, depth from frame $t$ can be forward-warped into frame
$t+1$. Let $\mathcal{W}(\cdot)$ denote flow-based warping. For visible and valid pixels,
a basic consistency condition is
\[
\mathcal{W}(D_t, F_{t\rightarrow t+1}) \approx D_{t+1}.
\]
Equivalently, after lifting into 3D, the scene flow induced by $(D_t,D_{t+1},F)$ should
agree with the predicted or simulated 3D displacement:
\[
S_{t\rightarrow t+1}(u)
\approx
X_{t+1}(u+F_{t\rightarrow t+1}(u)) - X_t(u).
\]
These constraints do not require photometric matching to simulated RGB frames. They
only require the exposed geometry and motion to describe a coherent evolving scene.

\subsection{Scope of the Formulation}

The formulation makes three design choices. First, it uses depth, optical flow, scene
flow, and trajectories as a tractable proxy for 4D scene evolution, rather than
requiring full dynamic volumetric reconstruction. Second, it separates visual realism
from physical structure: visual realism is inherited from the pretrained video model,
while physical structure is injected through geometry, motion, visibility, contact, and
trajectory supervision. Third, it supports both local and global physical constraints:
local consistency is expressed through depth-flow-scene-flow relations, while global
consistency is expressed through lifted trajectories over the generated sequence.
\section{Implementation}
\subsection{Stage-I Auxiliary Head Architecture}
\label{app:impl_model}

Stage I augments the frozen pretrained DiT video backbone with two lightweight
auxiliary heads for geometry and motion prediction. The backbone parameters are
kept fixed throughout this stage; only the auxiliary heads are optimized.

\paragraph{Motion head.}
We attach a motion prediction head to diffusion features captured from the DiT
output layer. The head consists of a LayerNorm followed by a modulated linear
projection, producing patchified motion tokens that are unpatchified into a
multi-channel motion field:
\[
\hat{M}_{t\rightarrow t+1}
=
(\hat{d}x,\hat{d}y,\hat{c}_2,\hat{c}_3)
\in \mathbb{R}^{H\times W\times 4}.
\]
The first two channels $(\hat{d}x,\hat{d}y)$ represent the optical flow
$\hat F_{t\rightarrow t+1}$ between adjacent frames, while the remaining channels
encode auxiliary motion information used by the motion head. Unless otherwise
specified, the supervised motion loss is applied to the optical-flow channels.

\paragraph{Depth head.}
We attach a depth prediction head following the spatiotemporal design of Video
Depth Anything~\cite{Chen2025VideoDA}. The head follows a DPT-style decoder and
incorporates temporal attention and multi-scale feature fusion to encourage
temporally coherent depth prediction. Specifically, we uniformly sample four
feature maps from the DiT backbone and process them with Reassemble and Fusion
layers to form a multi-scale feature pyramid. To enable cross-frame reasoning,
four temporal attention layers are inserted at low-resolution stages, where each
layer applies multi-head self-attention along the temporal dimension with
absolute positional embeddings. The decoder outputs per-frame depth maps
\[
\hat D_t \in \mathbb{R}^{H\times W}.
\]

Both heads share the same frozen diffusion feature interface. This design exposes
depth and motion variables required by later physics supervision, while leaving
the pretrained RGB video generator unchanged.

\subsection{Stage-I Pseudo-Supervised Losses}
\label{app:stage1_loss}

Stage I trains the auxiliary heads using pseudo labels from curated internet
videos and videos self-generated by the pretrained generator. We annotate both
sources with off-the-shelf monocular depth~\cite{Chen2025VideoDA} and optical
flow~\cite{Teed2020RAFTRA} estimators, obtaining pseudo depth
$\tilde D_t$ and pseudo flow $\tilde F_{t\rightarrow t+1}$. Estimator confidence,
validity, or visibility maps are used as supervision weights when available.

\paragraph{Depth loss.}
Since monocular depth is scale ambiguous, we normalize both the predicted and
pseudo depth within each clip before supervision. Let $\mathcal{N}(\cdot)$ denote
per-clip depth normalization. The pseudo-depth loss is
\[
\mathcal{L}_{D}^{pseudo}
=
\sum_{t,u}
w_t^D(u)\,
\rho
\left(
\mathcal{N}(\hat D_t(u))
-
\mathcal{N}(\tilde D_t(u))
\right),
\]
where $u$ indexes image locations, $w_t^D(u)$ is the depth confidence or validity
weight, and $\rho(\cdot)$ is a robust regression loss such as smooth-$\ell_1$.

\paragraph{Motion loss.}
The motion head is supervised using pseudo optical flow between adjacent frames:
\[
\mathcal{L}_{F}^{pseudo}
=
\sum_{t,u}
w_t^F(u)\,
\rho
\left(
\hat F_{t\rightarrow t+1}(u)
-
\tilde F_{t\rightarrow t+1}(u)
\right),
\]
where $w_t^F(u)$ masks invalid, low-confidence, or occluded correspondences.

\paragraph{Stage-I objective.}
The total Stage-I loss is
\[
\mathcal{L}_{S1}
=
\lambda_D \mathcal{L}_{D}^{pseudo}
+
\lambda_F \mathcal{L}_{F}^{pseudo}.
\]
Gradients from $\mathcal{L}_{S1}$ are applied only to the depth and motion heads.
The pretrained DiT backbone and the base video generation objective remain
unchanged. Thus, Stage I provides a prior-preserving RGB-D-motion interface
without injecting simulation physics or altering the original RGB generator.

\subsection{Stage-II Masked Geometry-Motion Losses}
\label{app:stage2_losses}

Stage II uses simulation annotations to inject local physical structure while
keeping the pretrained video backbone frozen. Let $\theta$ denote the frozen
backbone parameters, $\phi$ the trainable physics adapters, $\psi$ the learned
dynamic gate, and $\eta_D,\eta_F$ the depth and motion heads. Gradients from all
Stage-II losses are blocked from $\theta$ and update only
$\{\phi,\psi,\eta_D,\eta_F\}$.

Table~\ref{tab:app_stage2_loss_summary} summarizes the supervision source,
updated modules, and role of each Stage-II term. Importantly, the RGB-level term
is a self-consistency regularizer on generated frames and does not use
simulator-rendered RGB as a photometric target, whereas
$\mathcal{L}_{FM}^{high}$ does use latents encoded from simulator-rendered RGB
videos under a restricted high-noise adapter-only pathway.

\begin{table}[htbp]
\centering
\small
\setlength{\tabcolsep}{3.5pt}
\caption{\textbf{Stage-II loss summary.}
Stage II combines restricted simulator-latent adaptation with structure-level supervision. $\mathcal{L}_{FM}^{high}$ uses latents encoded from simulator-rendered RGB videos, but only at high-noise denoising steps and only through LoRA adapters. In contrast, $\mathcal{L}_{rgb\_warp}^{self}$ compares generated RGB frames under motion and does not match generated frames to simulator-rendered RGB.}
\label{tab:app_stage2_loss_summary}
\resizebox{\textwidth}{!}{%
\begin{tabular}{*{6}{p{0.16\textwidth}}}
\toprule
\textbf{Loss} &
\textbf{Uses simulator RGB latent?} &
\textbf{Photometric sim-RGB reconstruction target?} &
\textbf{Supervision source} &
\textbf{Updated modules} &
\textbf{Purpose} \\
\midrule
$\mathcal{L}_{FM}^{high}$ &
Yes, simulator-rendered RGB latent &
No, not a low-noise photometric reconstruction loss &
high-noise noised training latents encoded from simulator-rendered RGB videos &
adapters only &
keep adapted high-noise denoising on the video manifold \\
$\mathcal{L}_{depth}^{mask}$ &
No &
No &
sim depth + visibility + dynamic masks &
depth head + adapters &
metric local geometry supervision \\
$\mathcal{L}_{motion}^{mask}$ &
No &
No &
sim flow / projected motion + visibility + dynamic masks &
motion head + adapters &
local physical motion supervision \\
$\mathcal{L}_{geom\_warp}^{mask}$ &
No &
No &
sim flow / projected motion + visibility + dynamic masks &
depth head + adapters &
geometry-motion consistency under physical correspondences \\
$\mathcal{L}_{rgb\_warp}^{self}$ &
No &
No &
generated RGB frames + predicted motion &
adapters + motion head &
temporal RGB self-consistency without simulator RGB matching \\
$\mathcal{L}_{gate}$ &
No &
No &
sim dynamic / object masks &
gate + adapters &
learn physics-relevant regions \\
\bottomrule
\end{tabular}%
}
\end{table}

\paragraph{Mask weighting.}
For each frame $t$, the simulator provides a dynamic or object mask
$M_t^{dyn}$ and visibility masks $M_t^{vis}$ or
$M_{t\rightarrow t+1}^{vis}$. We use dynamic masks as soft weights rather than
hard exclusions:
\begin{equation}
w_t(u)
=
1+\beta M_t^{dyn}(u),
\label{eq:app_stage2_weight}
\end{equation}
where $u$ indexes image locations and $\beta$ controls the emphasis on
physics-relevant regions. This preserves supervision on global scene geometry
while increasing the contribution of moving objects, deformations, and
interaction regions.

\paragraph{Masked depth supervision.}
Given simulator depth $D_t^{sim}$ and predicted depth $\hat D_t$, we define
\begin{equation}
\mathcal{L}_{depth}^{mask}
=
\frac{
\sum_{t,u}
M_t^{vis}(u)\,
w_t(u)\,
\rho_D
\left(
\hat D_t(u)-D_t^{sim}(u)
\right)
}{
\sum_{t,u} M_t^{vis}(u)\,w_t(u)+\epsilon
},
\label{eq:app_stage2_depth}
\end{equation}
where $\rho_D$ is a robust regression loss, e.g., smooth-$\ell_1$, and
$\epsilon$ avoids division by zero. Since simulation provides metric depth, we
supervise depth in the simulator scale; when necessary, we apply per-sequence
normalization for numerical stability.

\paragraph{Masked motion supervision.}
The simulator provides optical flow or projected motion fields
$F_{t\rightarrow t+1}^{sim}$. We supervise the motion head by
\begin{equation}
\mathcal{L}_{motion}^{mask}
=
\frac{
\sum_{t,u}
M_{t\rightarrow t+1}^{vis}(u)\,
w_t(u)\,
\rho_F
\left(
\hat F_{t\rightarrow t+1}(u)
-
F_{t\rightarrow t+1}^{sim}(u)
\right)
}{
\sum_{t,u} M_{t\rightarrow t+1}^{vis}(u)\,w_t(u)+\epsilon
},
\label{eq:app_stage2_motion}
\end{equation}
where $M_{t\rightarrow t+1}^{vis}$ removes invalid or occluded correspondences.
This term trains the motion head to predict physically grounded local motion
without using simulator RGB appearance as a target.

\paragraph{Gate supervision.}
The dynamic gate $\hat G_t$ is supervised by simulator object or dynamic masks:
\begin{equation}
\mathcal{L}_{gate}
=
\mathrm{BCE}(\hat G_t,M_t^{dyn})
+
\mathrm{Dice}(\hat G_t,M_t^{dyn}).
\label{eq:app_stage2_gate}
\end{equation}
Simulator masks are used only during training. At inference time, the gate is
predicted from model features and no simulator annotation is required.

\paragraph{Simulation-flow guided geometry warp.}
Depth and motion supervision alone do not guarantee temporal coherence. We
therefore use simulator flow or projected motion as a physically valid
correspondence field to couple predicted geometry across adjacent frames:
\begin{equation}
\mathcal{L}_{geom\_warp}^{mask}
=
\frac{
\sum_{t,u}
M_{t\rightarrow t+1}^{vis}(u)\,
w_t(u)\,
\left\|
\mathcal{W}
\left(
\hat D_t,
F_{t\rightarrow t+1}^{sim}
\right)(u)
-
\hat D_{t+1}(u)
\right\|_1
}{
\sum_{t,u} M_{t\rightarrow t+1}^{vis}(u)\,w_t(u)+\epsilon
}.
\label{eq:app_stage2_geom_warp}
\end{equation}
Here $\mathcal{W}(\cdot)$ warps predicted depth at time $t$ according to the
simulator-provided motion. This teacher-forced warp avoids the degenerate case
where predicted depth and predicted flow co-adapt to a self-consistent but
physically incorrect correspondence. The predicted flow is instead learned
directly through $\mathcal{L}_{motion}^{mask}$.

\paragraph{Generated-frame RGB warp self-consistency.}
We include a weak RGB warp term to stabilize temporal appearance, but it is not a
photometric loss against simulator-rendered RGB. Let $\hat I_t$ denote generated
RGB frames and $\hat F_{t\rightarrow t+1}$ the predicted flow. We define
\begin{equation}
\mathcal{L}_{rgb\_warp}^{self}
=
\frac{
\sum_{t,u}
M_{t\rightarrow t+1}^{vis}(u)\,
w_t(u)\,
\rho_I
\left(
\mathcal{W}
\left(
\hat I_t,
\hat F_{t\rightarrow t+1}
\right)(u)
-
\hat I_{t+1}(u)
\right)
}{
\sum_{t,u} M_{t\rightarrow t+1}^{vis}(u)\,w_t(u)+\epsilon
}.
\label{eq:app_stage2_rgb_self}
\end{equation}
This term only encourages adjacent generated frames to be consistent under the
model-predicted motion. It does not compare $\hat I_t$ or $\hat I_{t+1}$ to any
simulator RGB frame, and therefore does not introduce direct simulator appearance
fitting.

\subsection{High-Noise Flow Matching for Stage-II Adapter Training}
\label{app:stage2_fm}

Stage II keeps the pretrained video backbone frozen and trains only lightweight
physics adapters. The flow-matching term is used to keep the adapted high-noise
denoising trajectory compatible with the video generation manifold. It uses
latents encoded from simulator-rendered RGB videos, but it is not a full-backbone
simulation-domain fine-tuning objective and it is not a low-noise photometric
matching loss to simulator-rendered RGB.

\paragraph{Flow matching preliminaries.}
Let $\mathbf{z}_0\sim \mathcal{N}(0,I)$ denote noise and
$\mathbf{z}_1\sim q(\mathbf{z}_1\mid c)$ denote a data latent conditioned on
prompt or image condition $c$. For interpolation time $s\in[0,1]$, we use the
linear probability path
\begin{equation}
\mathbf{z}_s
=
(1-s)\mathbf{z}_0+s\mathbf{z}_1 .
\label{eq:app_linear_path}
\end{equation}
The corresponding conditional vector field is
\begin{equation}
u_s(\mathbf{z}_s\mid \mathbf{z}_1)
=
\frac{d\mathbf{z}_s}{ds}
=
\mathbf{z}_1-\mathbf{z}_0 .
\label{eq:app_target_velocity}
\end{equation}
A flow model predicts a velocity field $v(\mathbf{z}_s,s,c)$ and is trained by
\begin{equation}
\mathcal{L}_{FM}
=
\mathbb{E}_{s,\mathbf{z}_0,\mathbf{z}_1,c}
\left[
\left\|
v(\mathbf{z}_s,s,c)
-
u_s(\mathbf{z}_s\mid \mathbf{z}_1)
\right\|_2^2
\right].
\label{eq:app_fm}
\end{equation}

\paragraph{Adapter-parameterized velocity.}
In Stage II, the velocity field is induced by the frozen pretrained model and the
trainable physics adapters:
\begin{equation}
v_{\theta,\phi}
=
v_{\theta}
+
\Delta v_{\phi},
\label{eq:app_adapter_velocity}
\end{equation}
where $\theta$ is frozen and $\phi$ contains only the lightweight adapter
parameters. The adapter update is implemented through the mask-gated residual
pathway:
\begin{equation}
h_\ell'
=
h_\ell
+
\alpha(\sigma_s)\,
\hat G_\ell
\odot
A_\ell(h_\ell).
\label{eq:app_gated_adapter}
\end{equation}
Here $\sigma_s$ is the noise level associated with interpolation time $s$,
$\alpha(\sigma_s)$ is the high-noise gate, and $\hat G_\ell$ is the learned
dynamic-region gate.

\paragraph{High-noise restriction.}
We restrict flow-matching updates to structure-forming high-noise steps. Let
$\gamma(s)$ denote a high-noise weighting function, e.g.,
\begin{equation}
\gamma(s)
=
\mathbb{I}[\sigma_s\geq \sigma_h],
\label{eq:app_hard_noise_gate}
\end{equation}
or a smooth version
\begin{equation}
\gamma(s)
=
\mathrm{sigmoid}
\left(
\frac{\sigma_s-\sigma_h}{\tau}
\right).
\label{eq:app_soft_noise_gate}
\end{equation}
The Stage-II high-noise flow-matching term is
\begin{equation}
\mathcal{L}_{FM}^{high}
=
\mathbb{E}_{s,\mathbf{z}_0,\mathbf{z}_1,c}
\left[
\gamma(s)
\left\|
v_{\theta,\phi}(\mathbf{z}_s,s,c)
-
(\mathbf{z}_1-\mathbf{z}_0)
\right\|_2^2
\right].
\label{eq:app_high_noise_fm}
\end{equation}
Because $\theta$ is frozen, gradients from
$\mathcal{L}_{FM}^{high}$ update only $\phi$. Moreover, the adapter itself is
modulated by $\alpha(\sigma_s)$, so low-noise appearance-refinement steps receive
no or negligible simulation-induced updates. This keeps simulation influence
concentrated on coarse structure and motion rather than texture, lighting, or
renderer-specific appearance statistics.
Equivalently, $\mathcal{L}_{FM}^{high}$ uses latents encoded from
simulator-rendered RGB videos, but only at high-noise denoising steps and only
through LoRA adapters. It is therefore a restricted simulator-latent adaptation
term, not full simulator RGB fine-tuning or low-noise photometric appearance
fitting.

\paragraph{Gradient paths.}
The frozen backbone provides features and velocity predictions but is not
updated. The trainable paths in Stage II are
\[
\mathcal{L}_{FM}^{high}
\rightarrow
\phi,
\qquad
\mathcal{L}_{depth}^{mask},
\mathcal{L}_{motion}^{mask},
\mathcal{L}_{geom\_warp}^{mask},
\mathcal{L}_{rgb\_warp}^{self}
\rightarrow
\{\phi,\eta_D,\eta_F\},
\qquad
\mathcal{L}_{gate}
\rightarrow
\{\phi,\psi\}.
\]
Thus, simulation-derived supervision affects generation only through the
mask-gated adapter and the auxiliary geometry-motion heads, while the pretrained
backbone remains unchanged.

\subsection{Stage-II Overall Objective}
\label{app:stage2_objective}

The full Stage-II objective is
\begin{equation}
\mathcal{L}_{S2}
=
\mathcal{L}_{FM}^{high}
+
\lambda_R \mathcal{L}_{rgb\_warp}^{self}
+
\lambda_D\mathcal{L}_{depth}^{mask}
+
\lambda_F\mathcal{L}_{motion}^{mask}
+
\lambda_W\mathcal{L}_{geom\_warp}^{mask}
+
\lambda_G\mathcal{L}_{gate}.
\label{eq:app_stage2_full_objective}
\end{equation}
The flow-matching term keeps the high-noise adapter pathway compatible with the
video generator. The depth, motion, and geometry-warp terms inject local
structure from simulation. The RGB warp term is only a generated-frame temporal
self-consistency regularizer and does not use simulator RGB as a photometric
target. The gate term learns where physical changes occur. All terms are
optimized with the pretrained backbone frozen, so Stage II performs controlled
local physics injection rather than full simulation-domain fine-tuning.

\subsection{Stage-III Reward Implementation: Uniform 4D Chamfer}
\label{app:impl_rl}

Stage III optimizes a global 4D trajectory-level reward over completed video
samples. Unlike Stage II, which uses mask-weighted local supervision, Stage III
evaluates the generated sequence as a complete spatiotemporal object. Visibility
and dynamic-region filtering are used only to remove invalid or inactive points;
after this filtering, the reward uses a uniform 4D Chamfer form. It does not
introduce category-specific physical terms such as contact, momentum, or
fluid-specific rules.

\paragraph{Generated 4D point construction.}
Given a generated RGB-D-motion sequence
$\{I_t,\hat D_t,\hat F_{t\rightarrow t+1}\}_{t=0}^{T-1}$, we construct a
spatiotemporal point set from the predicted depth. For each valid pixel
$u=(u_x,u_y)$ at frame $t$, we back-project the depth into 3D:
\begin{equation}
\mathbf{X}^{gen}_{t}(u)
=
\hat D_t(u)\,
K_t^{-1}
\begin{bmatrix}
u_x\\
u_y\\
1
\end{bmatrix},
\label{eq:app_gen_unproject}
\end{equation}
where $K_t$ is the camera intrinsic matrix. When camera extrinsics are available,
we transform all points into a common world or reference coordinate frame:
\begin{equation}
\mathbf{X}^{gen,w}_{t}(u)
=
R_t^{-1}
\left(
\mathbf{X}^{gen}_{t}(u)-\mathbf{t}_t
\right).
\label{eq:app_gen_world}
\end{equation}
For fixed-camera environments, the camera coordinate system is used as the common
reference frame. We then append a discrete timestamp $\tau_t$ to obtain a 4D
point:
\begin{equation}
p^{gen}_{t,u}
=
\left(
\mathbf{X}^{gen,w}_{t}(u),
\tau_t
\right).
\label{eq:app_gen_4d_point}
\end{equation}
The motion prediction $\hat F_{t\rightarrow t+1}$ is used to estimate validity,
and dynamic-region masks are used to retain active regions. We then sample a
fixed number of valid points per frame to avoid any single frame dominating the
reward, and collect them over time to form
\begin{equation}
\mathcal{P}^{gen}
=
\{p^{gen}_{t,u}\}_{t,u}.
\label{eq:app_pgen}
\end{equation}

\paragraph{Simulation 4D trajectory construction.}
The simulator provides geometry and motion in a world coordinate
system. For each simulation point $\mathbf{X}^{sim,w}_{t,j}$ at frame $t$, we
construct the corresponding 4D point
\begin{equation}
p^{sim}_{t,j}
=
\left(
\mathbf{X}^{sim,w}_{t,j},
\tau_t
\right).
\label{eq:app_sim_4d_point}
\end{equation}
If the simulator provides rendered depth rather than explicit world points, we
apply the same unprojection and camera-to-world transformation as in
Eq.~\eqref{eq:app_gen_unproject}--\eqref{eq:app_gen_world}. We remove invalid,
invisible, out-of-view, or inactive points using simulator visibility and
dynamic-region annotations, and sample the same number of points per frame as
for the generated sequence. This yields the simulator trajectory point set
\begin{equation}
\mathcal{P}^{sim}
=
\{p^{sim}_{t,j}\}_{t,j}.
\label{eq:app_psim}
\end{equation}

\paragraph{4D point distance.}
For two 4D points
$p=(\mathbf{x}_p,\tau_p)$ and $q=(\mathbf{x}_q,\tau_q)$, where
$\mathbf{x}\in\mathbb{R}^3$, we define the spatiotemporal distance
\begin{equation}
d_{4D}(p,q)
=
\|\mathbf{x}_p-\mathbf{x}_q\|_2^2
+
\lambda_t
|\tau_p-\tau_q|^2.
\label{eq:app_4d_point_distance}
\end{equation}
The coefficient $\lambda_t$ controls the relative weight of temporal mismatch.
In our implementation, $\lambda_t$ is chosen according to the typical per-frame
displacement in the simulator and is kept fixed across physical categories.

\paragraph{4D Chamfer distance.}
Given the generated and simulated 4D point sets, we compute the symmetric 4D
Chamfer Distance:
\begin{equation}
\mathrm{CD}_{4D}
\left(
\mathcal{P}^{gen},
\mathcal{P}^{sim}
\right)
=
\frac{1}{|\mathcal{P}^{gen}|}
\sum_{p\in\mathcal{P}^{gen}}
\min_{q\in\mathcal{P}^{sim}}
d_{4D}(p,q)
+
\frac{1}{|\mathcal{P}^{sim}|}
\sum_{q\in\mathcal{P}^{sim}}
\min_{p\in\mathcal{P}^{gen}}
d_{4D}(p,q).
\label{eq:app_4d_chamfer}
\end{equation}
Unlike same-time 3D alignment, this metric compares generated and simulated
trajectories as spatiotemporal objects. It therefore penalizes both spatial
deviation and temporal misalignment.

\paragraph{Stage-III reward.}
The Stage-III reward is the negative 4D Chamfer Distance:
\begin{equation}
R_{4D}(V)
=
-
\mathrm{CD}_{4D}
\left(
\mathcal{P}^{gen},
\mathcal{P}^{sim}
\right).
\label{eq:app_stage3_reward}
\end{equation}
This is the only trajectory-alignment reward used in Stage III. We deliberately avoid
category-specific rewards such as contact penalties, momentum constraints,
penetration rules, fluid-specific objectives, or material-state classifiers.
Contact annotations may be available from simulation, but they are not used as
explicit category-specific rewards. Rigid objects, fluids, cloth, deformables,
and granular materials are all supervised through the same observable 4D
trajectory alignment.

\section{Training Details}
\label{app:training_details}

This section consolidates the stage-wise training protocol, compute footprint, and inference overhead of Phys4D for reproducibility.
Unless otherwise noted, the reported training details correspond to WAN2.2-5B, which is also the backbone used for the main ablation studies.

\subsection{Stage-wise Training Protocol}
\label{app:training_protocol}

Phys4D follows the same three-stage training pipeline described in the main paper.
Stage~I trains only the auxiliary depth and motion heads while keeping the pretrained DiT video backbone frozen, thereby constructing the RGB-D-motion interface without altering RGB generation.
Stage~II then fine-tunes lightweight high-noise LoRA adapters together with the auxiliary heads using simulation-derived geometry, motion, and warp supervision.
Stage~III keeps the backbone frozen and further optimizes the same adapter space using trajectory-level policy optimization for global physical consistency.

\subsection{Training Cost and Inference Overhead}
\label{app:rq4_cost}

\paragraph{Training cost.}
All main ablations are conducted on WAN2.2-5B.
Stage~I trains only the auxiliary depth and motion heads with the frozen DiT backbone and takes approximately 1 day on 8$\times$H100 GPUs.
Stage~II fine-tunes high-noise LoRA adapters and takes approximately 2 days.
Stage~III optimizes the same LoRA parameters via trajectory-level policy optimization and takes approximately 1.5 days.
The total wall-clock cost is approximately 4.5 days on 8$\times$H100 GPUs.

\paragraph{Inference overhead.}
Phys4D adds lightweight LoRA adapters and auxiliary depth/motion heads.
The auxiliary heads are not part of the iterative denoising loop, so inference remains close to the original backbone.
We measure per-video inference time for the base model and the Phys4D-enhanced model.
The measured runtimes are summarized in Table~\ref{tab:app_runtime}.

\begin{table}[htbp]
\centering
\small
\setlength{\tabcolsep}{5pt}
\caption{\textbf{Inference overhead.}
Phys4D introduces only slight inference slowdown because it uses lightweight LoRA adapters and small auxiliary heads outside the iterative denoising loop.}
\label{tab:app_runtime}
\begin{tabular}{lccc}
\toprule
\textbf{Model}
& \textbf{Baseline} $\downarrow$
& \textbf{+ Phys4D} $\downarrow$
& \textbf{Slowdown} \\
\midrule
Open-Sora V1.2 & 40s/video & 42s/video & +2s \\
WAN2.2-5B      & 120s/video & 125s/video & +5s \\
CogVideoX-5B   & 75s/video & 78s/video & +3s \\
\bottomrule
\end{tabular}
\end{table}

\paragraph{Analysis.}
Phys4D introduces only small inference overhead: +2s for Open-Sora V1.2, +5s for WAN2.2-5B, and +3s for CogVideoX-5B.
This supports the design goal of injecting physical structure without substantially increasing inference cost.

\subsection{Stage~III RL Training Details and Curves}
\label{app:rq4_rl_training}

\paragraph{Motivation.}
Stage~II provides dense local supervision through depth, motion, and warp consistency, but these losses are local: they mainly constrain adjacent frames and cannot directly optimize sequence-level physical outcomes such as trajectory drift, penetration, or delayed object-interaction failures. 
Stage~III therefore uses a trajectory-level policy optimization objective to correct residual global physical violations after Stage~II has established a locally coherent RGB-D-motion interface.

\paragraph{Denoising MDP.}
Following diffusion policy optimization, we formulate the reverse denoising process as a finite-horizon Markov decision process. 
At denoising step $t$, the state is
\begin{equation}
s_t = (c, t, z_t),
\end{equation}
where $c$ denotes the text and image conditioning signal and $z_t$ is the noisy latent.
The action is the next latent in the reverse process,
\begin{equation}
a_t = z_{t-1},
\end{equation}
and the policy is induced by the video diffusion model:
\begin{equation}
\pi_\theta(a_t \mid s_t) = p_\theta(z_{t-1}\mid z_t, c).
\end{equation}
A complete denoising trajectory $\tau=(z_T,z_{T-1},\ldots,z_0)$ produces a final video sample after decoding $z_0$.

\paragraph{Trainable parameters.}
During Stage~III, the pretrained DiT backbone remains frozen. 
We optimize only the lightweight adapter parameters initialized from Stage~II, while using the RGB-D-motion interface to lift generated videos into 4D trajectories for reward computation. 
This confines simulation-driven updates to a low-rank adapter space and reduces the risk of overfitting to simulator-specific appearance statistics.

\paragraph{Reward.}
After decoding the final video, we predict its depth and motion, lift it into a 4D trajectory representation, and compare it with simulator ground truth using the 4D Chamfer distance as a spatiotemporal trajectory agreement metric. 
Each point is represented as $(x,y,z,\sqrt{\lambda_t}\, t)$, where $\lambda_t$ controls the relative weight of temporal mismatch. 
The terminal reward is defined as the negative 4D trajectory discrepancy:
\begin{equation}
R(\tau) = - \mathrm{CD}_{4D}\!\left(\widehat{\mathcal{P}}, \mathcal{P}^{\mathrm{gt}}\right),
\end{equation}
where $\widehat{\mathcal{P}}$ and $\mathcal{P}^{\mathrm{gt}}$ denote predicted and ground-truth spatio-temporal point sets.
This reward encourages spatially accurate and temporally aligned world evolution, rather than appearance matching.
We normalize the reward to $[-1,1]$, and the temporal term in the 4D Chamfer reward helps keep the signal smooth and optimization well-behaved.

\paragraph{Optimization.}
We optimize the denoising policy with PPO. 
Let $\pi_{\theta_{\mathrm{old}}}$ be the frozen rollout policy used to collect samples and
\begin{equation}
\rho_t(\theta)=
\frac{\pi_\theta(a_t\mid s_t)}
{\pi_{\theta_{\mathrm{old}}}(a_t\mid s_t)}
\end{equation}
be the likelihood ratio.
The clipped PPO objective is
\begin{equation}
\mathcal{L}_{\mathrm{PPO}}
=
-\mathbb{E}_t
\left[
\min\left(
\rho_t(\theta) A_t,\,
\mathrm{clip}(\rho_t(\theta),1-\epsilon,1+\epsilon) A_t
\right)
\right]
+
\beta\,\mathrm{KL}\!\left(\pi_\theta\,\|\,\pi_{\mathrm{ref}}\right),
\end{equation}
where $A_t$ is the normalized advantage derived from the sequence-level 4D reward, $\pi_{\mathrm{ref}}$ is the Stage~II reference policy, and the KL term prevents the adapter update from drifting too far from the pretrained video prior.
Since the reward is computed on the completed generated sequence, we assign the normalized terminal reward to the denoising trajectory and regularize each denoising step with the KL penalty.

\begin{figure*}[htbp]
    \centering
    \includegraphics[width=0.95\textwidth]{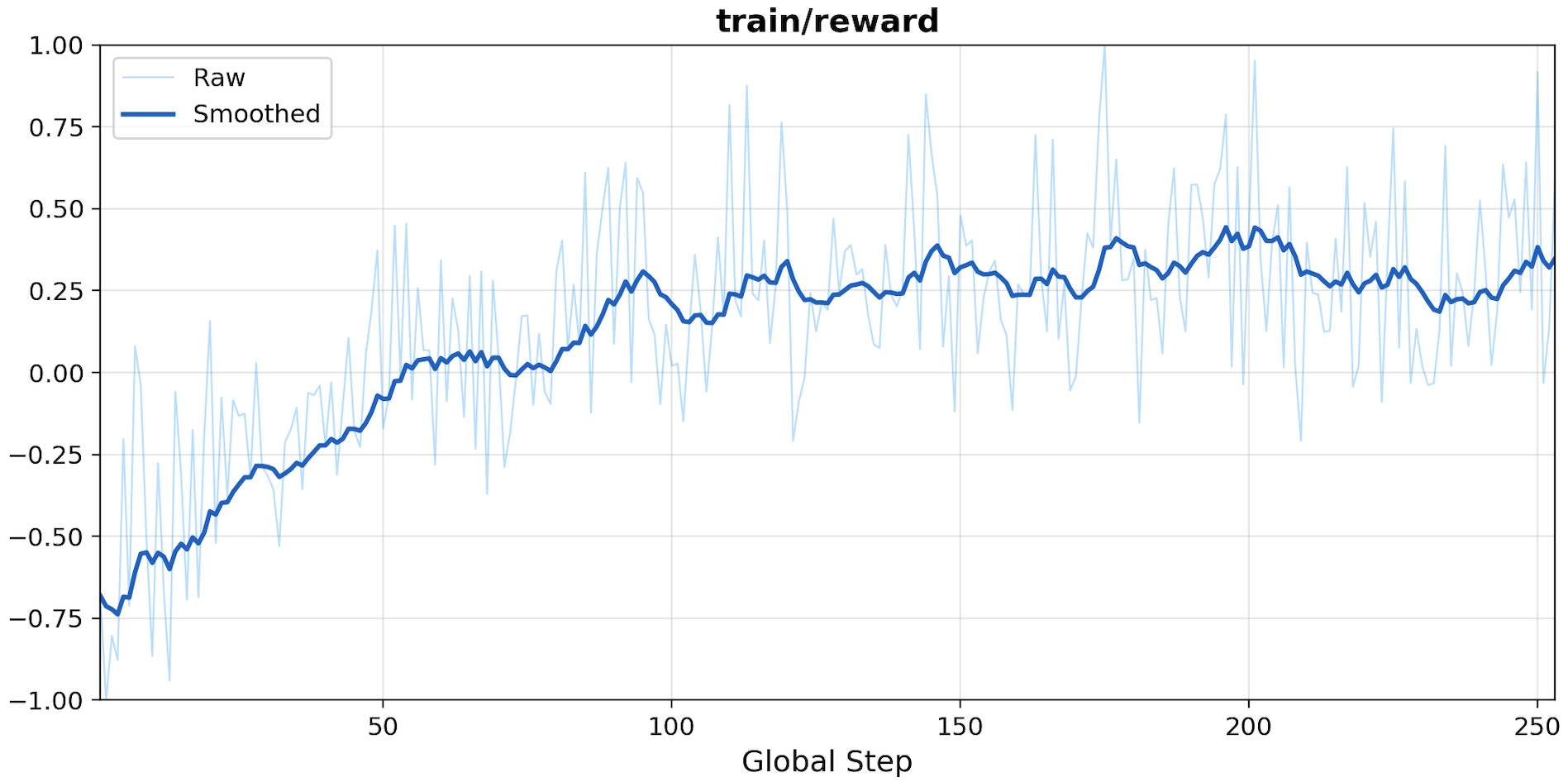}
    \caption{\textbf{Stage~III RL training curves.}
    Stage~III starts from the Stage~II checkpoint and optimizes only lightweight adapter parameters while keeping the pretrained DiT backbone frozen.
    We plot the moving average of the trajectory-level 4D reward, the corresponding 4D Chamfer distance, and the KL divergence to the Stage~II reference policy.
    The reward improves during training while the KL remains bounded, indicating stable trajectory-level optimization without large deviation from the pretrained generative prior.}
    \label{fig:app_rl_training_curve}
\end{figure*}

\paragraph{Training cost.}
Stage~III takes approximately 1.5 days on 8$\times$H100 GPUs when initialized from Stage~II.
Together with Stage~I and Stage~II, the full training pipeline takes approximately 4.5 days on 8$\times$H100 GPUs for WAN2.2-5B.
Stage~III does not introduce additional inference-time iterative modules; after training, inference only uses the lightweight adapters and the same denoising process as the base video generator.

\paragraph{Analysis.}
The RL training curve in Fig.~\ref{fig:app_rl_training_curve} complements the stage-wise ablation in Table~\ref{tab:rq4_stage_ablation}.
Stage~II improves Physics-IQ from 21.33 to 32.20 by injecting local physical structure, and Stage~III further improves it to 35.80 by optimizing global trajectory-level alignment.
The training curve shows that this improvement is not caused by an unstable reward spike: the trajectory-level reward increases smoothly, 4D Chamfer decreases, and the KL to the Stage~II reference policy remains controlled.
This supports the role of Stage~III as a stable global physical alignment step built on top of the local geometry-motion initialization from Stage~II.

\section{Qualitative Result}
\begin{figure*}[htbp]
    \centering
    \includegraphics[width=\textwidth]{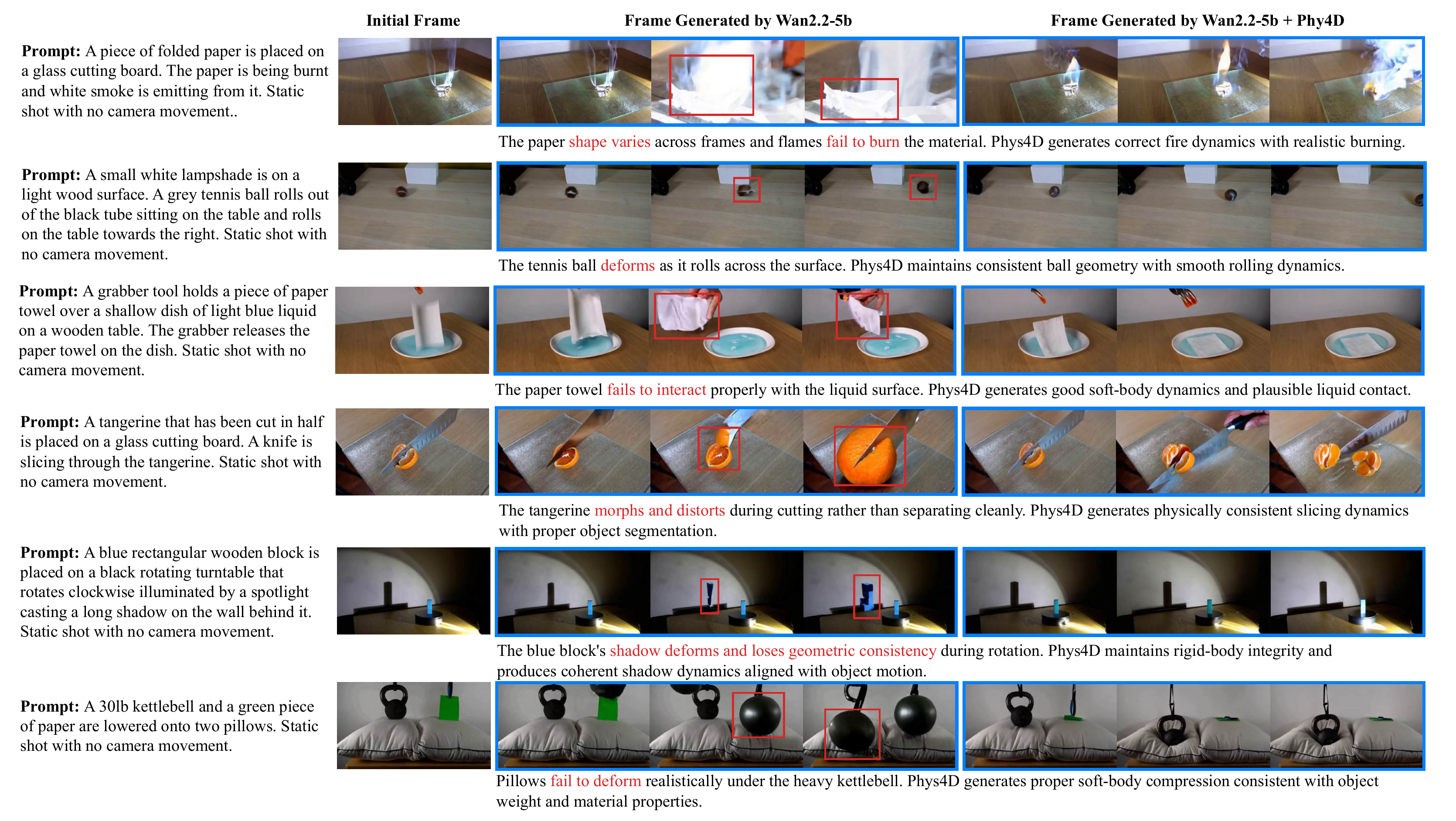}
    \caption{\textbf{Qualitative Result of a Phys4D-Generated Scene.}}
    \label{fig:supp_quality_demo}
\end{figure*}

Figure~\ref{fig:supp_quality_demo} and Figure~\ref{fig:quality_demo} qualitatively compare Phys4D with the base video diffusion model across a diverse set of physical interaction scenarios under static camera settings, including rigid-body manipulation, multi-object dynamics, soft-body deformation, fluid flow, and contact-rich interactions.
While the baseline model often generates visually coherent frames, it frequently violates object-level physical constraints, leading to shape distortion, object duplication, incorrect interaction outcomes, or implausible motion patterns over time.

In manipulation scenarios, such as object placement and grasping, baseline models tend to introduce non-physical deformations or geometry drift, even for nominally rigid objects. Phys4D, in contrast, preserves object integrity throughout the interaction, maintaining consistent geometry and contact behavior as objects are manipulated or released.
For multi-object scenes, baseline generations commonly suffer from object count inconsistencies and unstable trajectories, whereas Phys4D maintains correct object identities and produces smooth, coherent motion consistent with the underlying scene layout.

We further observe substantial improvements in fluid–object interactions. In pouring scenarios, baseline models often violate basic physical principles, such as gravity alignment or container geometry, resulting in fluid emerging from incorrect locations. Phys4D generates fluid motion that is directionally consistent with gravity and constrained by the container structure, yielding more realistic flow behavior over time.

Across all scenarios, these qualitative results highlight that Phys4D goes beyond frame-wise visual plausibility, enforcing object-level continuity, interaction consistency, and physically grounded dynamics. This supports the claim that Phys4D captures a coherent observable geometry-motion interface, rather than synthesizing isolated, appearance-consistent frames.
\section{Experiments}

\subsection{Detailed analysis of three physics evaluation benchmarks}
\label{app:experiment_physics}

This section provides additional experimental details and full quantitative results. 
We evaluate whether Phys4D transfers simulation-derived physical structure to generated RGB videos using three external video-level physics evaluations: Physics-IQ~\cite{motamed2025physicsiq}, VBench-2.0 Physics~\cite{vbench2025vbench2}, and PhyGenBench~\cite{phygenbench2025}. 
For all benchmarks, the base model and its Phys4D-enhanced counterpart are evaluated under matched prompts, conditioning inputs, sampling settings, video length, and official evaluation pipelines.

\subsection{Physics-IQ}
\label{app:physics_iq}

\paragraph{Benchmark and protocol.}
Physics-IQ~\cite{motamed2025physicsiq} evaluates whether video generation models can predict physically plausible future videos from real-world captured physical interactions. 
Each test case provides a switch frame and a text description, and the model generates a 5-second continuation. 
Following the official protocol, generated videos are compared with ground-truth future videos using four complementary metrics: MSE, Spatiotemporal IoU, Spatial IoU, and Weighted Spatial IoU. 
MSE measures pixel-level future-frame reconstruction error, while the IoU metrics evaluate whether the generated motion occurs at the correct location, time, and magnitude. 
The final Physics-IQ score aggregates these metrics into a normalized physical prediction score.

For evaluation, we generate 5-second videos conditioned on the switch frame following the Physics-IQ protocol. 
We use the same stochastic sampling setup for the base and Phys4D-enhanced models, with temperature $1.0$ and top-$p=0.95$ where applicable. 
We report matched-protocol comparisons only. 
Public leaderboard numbers obtained under different model checkpoints, input modalities, preprocessing, or generation settings are not mixed with our same-backbone comparisons.
The full matched-protocol results are presented in Table~\ref{tab:app_physics_iq_full}.

\begin{table*}[htbp]
\centering
\small
\setlength{\tabcolsep}{5pt}
\caption{\textbf{Full Physics-IQ results under the matched switch-frame I2V protocol.}
MSE measures future-frame reconstruction error. 
ST-IoU, S-IoU, and WS-IoU measure spatiotemporal, spatial, and weighted spatial motion-mask overlap, respectively. 
Phys4D consistently improves all Physics-IQ metrics across multiple backbones.}
\label{tab:app_physics_iq_full}
\resizebox{\textwidth}{!}{
\begin{tabular}{ll l cccc c}
\toprule
\textbf{Backbone} & \textbf{Params} & \textbf{Method} 
& \textbf{MSE} $\downarrow$
& \textbf{ST-IoU} $\uparrow$
& \textbf{S-IoU} $\uparrow$
& \textbf{WS-IoU} $\uparrow$
& \textbf{Verified} $\uparrow$ \\
\midrule
CogVideoX      & 5B   & Base     & 0.013 & 0.116 & 0.222 & 0.142 & 25.38 \\
CogVideoX      & 5B   & + Phys4D & \textbf{0.008} & \textbf{0.182} & \textbf{0.270} & \textbf{0.169} & \textbf{37.20} \\
\midrule
WAN2.2         & 5B   & Base     & 0.016 & 0.088 & 0.150 & 0.105 & 21.33 \\
WAN2.2         & 5B   & + Phys4D & \textbf{0.013} & \textbf{0.158} & \textbf{0.238} & \textbf{0.149} & \textbf{35.80} \\
\midrule
Open-Sora V1.2 & 1.1B & Base     & 0.021 & 0.072 & 0.135 & 0.092 & 14.5 \\
Open-Sora V1.2 & 1.1B & + Phys4D & \textbf{0.014} & \textbf{0.119} & \textbf{0.211} & \textbf{0.128} & \textbf{24.7} \\
\bottomrule
\end{tabular}
}
\end{table*}

\paragraph{Analysis.}
Phys4D achieves substantial improvements over base video diffusion models across multiple backbones. 
When applied to CogVideoX-5B, Phys4D improves Spatiotemporal IoU from 0.116 to 0.182, a $56.9\%$ relative gain, and Spatial IoU from 0.222 to 0.270, a $21.6\%$ relative gain, while reducing MSE from 0.013 to 0.008, a $38.5\%$ reduction. 
Similarly, Phys4D improves WAN2.2-5B's Spatiotemporal IoU from 0.088 to 0.158, a $79.5\%$ relative gain, and Open-Sora V1.2 from 0.072 to 0.119, a $65.3\%$ relative gain. 
The overall Physics-IQ Verified score also improves from 25.38 to 37.20 on CogVideoX-5B, from 21.33 to 35.80 on WAN2.2-5B, and from 14.5 to 24.7 on Open-Sora V1.2.
These consistent improvements across architectures indicate that the physics-grounded training paradigm is architecture-agnostic and effectively transfers to different video diffusion backbones.

Notably, Phys4D on CogVideoX-5B achieves the best Physics-IQ score among the reported same-backbone comparisons, with an MSE of 0.008, Spatiotemporal IoU of 0.182, Spatial IoU of 0.270, Weighted Spatial IoU of 0.169, and overall Verified score of 37.20.
This suggests that simulation-grounded structure injection can substantially improve the physical prediction ability of open-source video diffusion models.

\paragraph{Statistical significance and consistency.}
We report benchmark averages over the full held-out evaluation sets rather than isolated examples. 
While we do not include confidence intervals, the observed gains are consistent across multiple backbones, metrics, and ablations, providing convergent evidence that Phys4D improves observable physical plausibility.

\subsection{VBench-2.0 Physics}
\label{app:vbench2_physics}

\paragraph{Benchmark and protocol.}
VBench-2.0~\cite{vbench2025vbench2} extends video generation evaluation from superficial faithfulness, such as visual quality and temporal smoothness, to intrinsic faithfulness. 
Its Physics dimension evaluates whether generated videos follow real-world physical principles across four sub-dimensions: Mechanics, Thermotics, Material, and Multi-View Consistency. 
These sub-dimensions assess physical state changes, thermal transformations, material behavior, and geometry consistency under viewpoint or camera changes.

We follow the official VBench-2.0 Physics prompt suite and evaluation pipeline. 
For each backbone, the base model and the Phys4D-enhanced model generate videos using the same prompts and sampling settings. 
We report each physics sub-dimension and the overall Physics Score, computed as the average over the four sub-dimensions. 
Higher scores indicate stronger intrinsic physical faithfulness.
Table~\ref{tab:app_vbench2_physics_full} provides the complete per-backbone breakdown.

\begin{table*}[htbp]
\centering
\small
\setlength{\tabcolsep}{5pt}
\caption{\textbf{Full VBench-2.0 Physics results.}
VBench-2.0 Physics evaluates intrinsic physical faithfulness across Mechanics, Thermotics, Material, and Multi-View Consistency. 
Phys4D improves the overall Physics Score across all three backbones and improves all individual sub-dimensions.}
\label{tab:app_vbench2_physics_full}
\resizebox{\textwidth}{!}{
\begin{tabular}{ll l ccccc}
\toprule
\textbf{Backbone} & \textbf{Params} & \textbf{Method}
& \textbf{Mechanics} $\uparrow$
& \textbf{Thermotics} $\uparrow$
& \textbf{Material} $\uparrow$
& \makecell{\textbf{Multi-View}\\\textbf{Consistency}} $\uparrow$
& \textbf{Physics Score} $\uparrow$ \\
\midrule
CogVideoX      & 5B   & Base     & 0.5982 & 0.5618 & 0.4572 & 0.3224 & 0.4849 \\
CogVideoX      & 5B   & + Phys4D & \textbf{0.7284} & \textbf{0.6706} & \textbf{0.5002} & \textbf{0.4824} & \textbf{0.5954} \\
\midrule
WAN2.2         & 5B   & Base     & 0.6220 & 0.5786 & 0.4255 & 0.3450 & 0.4928 \\
WAN2.2         & 5B   & + Phys4D & \textbf{0.7604} & \textbf{0.7002} & \textbf{0.5418} & \textbf{0.4964} & \textbf{0.6247} \\
\midrule
Open-Sora V1.2 & 1.1B & Base     & 0.5821 & 0.5508 & 0.4083 & 0.3079 & 0.4623 \\
Open-Sora V1.2 & 1.1B & + Phys4D & \textbf{0.7147} & \textbf{0.6603} & \textbf{0.5068} & \textbf{0.4778} & \textbf{0.5899} \\
\bottomrule
\end{tabular}
}
\end{table*}

\paragraph{Analysis.}
Phys4D consistently improves the overall VBench-2.0 Physics Score across all three backbones. 
CogVideoX-5B improves from 0.4849 to 0.5954, a $22.8\%$ relative gain; WAN2.2-5B improves from 0.4928 to 0.6247, a $26.8\%$ relative gain; and Open-Sora V1.2 improves from 0.4623 to 0.5899, a $27.6\%$ relative gain. 
The improvements are especially clear in Mechanics, Thermotics, and Multi-View Consistency, indicating that Phys4D improves both physical state evolution and geometry consistency. 
For example, WAN2.2-5B improves from 0.6220 to 0.7604 in Mechanics and from 0.3450 to 0.4964 in Multi-View Consistency. 
In this matched evaluation, Phys4D also improves the Material sub-dimension across all three backbones.
\subsection{PhyGenBench}
\label{app:phygenbench}

\paragraph{Benchmark and protocol.}
PhyGenBench~\cite{phygenbench2025} evaluates physical commonsense in text-to-video generation. 
It contains prompts covering four physical domains: Mechanics, Optics, Thermal phenomena, and Material properties. 
The benchmark is designed around explicit physical laws and observable physical phenomena, such as gravity, buoyancy, reflection, refraction, melting, boiling, hardness, and solubility. 
Its evaluator, PhyGenEval, uses a hierarchical protocol to assess whether the generated video contains the expected physical phenomenon, follows the correct physical order, and remains natural over the full video.

We use the official PhyGenBench prompt suite and PhyGenEval pipeline. 
For each prompt, we generate videos with the base backbone and the corresponding Phys4D-enhanced model under the same sampling settings. 
We report scores for Mechanics, Optics, Thermal, and Material categories, as well as their average. 
Higher scores indicate stronger physical commonsense correctness.
The full per-category results appear in Table~\ref{tab:app_phygenbench_full}.

\begin{table*}[htbp]
\centering
\small
\setlength{\tabcolsep}{5pt}
\caption{\textbf{Full PhyGenBench results.}
PhyGenBench evaluates physical commonsense generation across Mechanics, Optics, Thermal phenomena, and Material properties. 
Phys4D consistently improves the average score across all three backbones and improves all per-category scores.}
\label{tab:app_phygenbench_full}
\resizebox{\textwidth}{!}{
\begin{tabular}{ll l ccccc}
\toprule
\textbf{Backbone} & \textbf{Params} & \textbf{Method}
& \textbf{Mechanics} $\uparrow$
& \textbf{Optics} $\uparrow$
& \textbf{Thermal} $\uparrow$
& \textbf{Material} $\uparrow$
& \textbf{Average} $\uparrow$ \\
\midrule
CogVideoX      & 5B   & Base     & 0.40 & 0.56 & 0.41 & 0.43 & 0.45 \\
CogVideoX      & 5B   & + Phys4D & \textbf{0.53} & \textbf{0.66} & \textbf{0.57} & \textbf{0.56} & \textbf{0.58} \\
\midrule
WAN2.2         & 5B   & Base     & 0.42 & 0.50 & 0.42 & 0.46 & 0.45 \\
WAN2.2         & 5B   & + Phys4D & \textbf{0.52} & \textbf{0.63} & \textbf{0.56} & \textbf{0.57} & \textbf{0.57} \\
\midrule
Open-Sora V1.2 & 1.1B & Base     & 0.43 & 0.50 & 0.44 & 0.39 & 0.44 \\
Open-Sora V1.2 & 1.1B & + Phys4D & \textbf{0.53} & \textbf{0.62} & \textbf{0.57} & \textbf{0.56} & \textbf{0.57} \\
\bottomrule
\end{tabular}
}
\end{table*}

\paragraph{Analysis.}
Phys4D improves PhyGenBench scores across all three backbones. 
CogVideoX-5B improves from 0.45 to 0.58, a $28.9\%$ relative gain; WAN2.2-5B improves from 0.45 to 0.57, a $26.7\%$ relative gain; and Open-Sora V1.2 improves from 0.44 to 0.57, a $29.5\%$ relative gain. 
The improvements are consistent across Mechanics, Optics, Thermal, and Material categories. 
For instance, CogVideoX-5B improves from 0.40 to 0.53 in Mechanics and from 0.41 to 0.57 in Thermal, while WAN2.2-5B improves from 0.50 to 0.63 in Optics and from 0.46 to 0.57 in Material. 
These results suggest that Phys4D improves not only low-level motion prediction but also semantic physical commonsense as judged by an independent VLM-based evaluation pipeline.

\subsection{Additional Experiments}
\label{app:additional_experiments}

This section collects additional experiments and qualitative results that support the main RQ-driven evaluation but are not included in the main paper due to space constraints.
These results are not used as the primary evidence for the main claims.
Instead, they provide broader reference comparisons, qualitative examples, and longer-video diagnostics.

\subsubsection{Additional Physics-IQ Reference Comparison}
\label{app:additional_physics_iq_reference}

\paragraph{Setting.}
In the main paper and Sec.~\ref{app:physics_iq}, we focus on matched same-backbone comparisons, where each Phys4D-enhanced model is compared against its own base model under the same switch-frame I2V protocol.
For completeness, we also include a broader reference comparison with representative open-source, commercial, and physics-aware methods.
Because different models may use different checkpoints, input interfaces, preprocessing, and generation protocols, this table should be interpreted as a reference comparison rather than the primary matched-protocol evidence.

\begin{table}[htbp]
\centering
\small
\setlength{\tabcolsep}{6pt}
\caption{\textbf{Additional Physics-IQ reference comparison.}
This table provides broader context against representative video generation models and related physics-aware methods.
Because protocols and model interfaces may differ, the matched same-backbone results in Table~\ref{tab:app_physics_iq_full} remain the primary evidence.}
\label{tab:app_physics_iq_reference}
\begin{tabular}{lcc}
\toprule
\textbf{Model} & \textbf{Model Size} & \textbf{Physics-IQ Score} $\uparrow$ \\
\midrule
WAN2.2-5B + Phys4D        & 5B   & \textbf{35.80} \\
Veo3                      & N/A  & 23.2 \\
Runway Gen 3              & N/A  & 22.8 \\
WAN2.2-5B                 & 5B   & 21.33 \\
WAN2.2-14B                & 14B  & 20.8 \\
WAN2.2-5B + PhysGen3D     & 5B   & 20.4 \\
VideoPoet                 & 8B   & 20.3 \\
Lumiere                   & 4B   & 19.0 \\
Stable Video Diffusion    & 1.1B & 14.8 \\
Pika 1.0                  & N/A  & 13.0 \\
Sora                      & N/A  & 10.0 \\
\bottomrule
\end{tabular}
\end{table}

\paragraph{Analysis.}
The reference comparison shows that Phys4D improves WAN2.2-5B substantially over its base model and over WAN2.2-5B combined with PhysGen3D.
In this broader reference setting, WAN2.2-5B + Phys4D obtains a Physics-IQ score of 35.80, compared with 21.33 for WAN2.2-5B and 20.4 for WAN2.2-5B + PhysGen3D.
It also remains competitive with stronger commercial or larger-scale video models in this reference setting.
However, because these models may not share identical protocols, we do not use this table to make the primary quantitative claim.
The main claim is instead based on same-backbone matched comparisons in Table~\ref{tab:app_physics_iq_full}.
A complementary re-evaluation under the newer Physics-IQ Verified protocol is reported in Sec.~\ref{app:physics_iq_verified}.

\subsubsection{Physics-IQ Verified Re-evaluation}
\label{app:physics_iq_verified}

\paragraph{Motivation.}
The results in the main paper and in Table~\ref{tab:app_physics_iq_full} are obtained under the
original Physics-IQ workflow~\cite{motamed2025physicsiq}, using a frozen January~2026 snapshot of
its switch frames, prompts, ground-truth videos and masks, and scoring code.
After our submission, Physics-IQ Verified~\cite{radsch2026physicsiqverified} revised ambiguous
prompts, corrected artifacts and spurious activations in the ground-truth masks, and changed the
score aggregation from dataset-level to equally weighted sample-level aggregation.
These revisions materially affect absolute scores and model rankings, so Original and Verified
numbers should not be directly compared.
To remove benchmark-version ambiguity and to provide a direct comparison with current larger
image-to-video models, we additionally re-evaluated our checkpoints using the official
Physics-IQ Verified data, masks, best-practice prompts, generation protocol, and scoring pipeline.
All results use the switch-frame I2V setting, and every baseline is regenerated by us under this
single frozen protocol.

\paragraph{Protocol-shift control.}
Switching the \emph{same} WAN2.2-5B checkpoint from Original to Verified changes the base score by
only $+4.5$ ($16.8\rightarrow21.33$), confirming that the discrepancy between our reported numbers
and public leaderboard entries is driven primarily by checkpoint scale and inference settings
rather than by generation length, resolution, or the switch-frame setting.
We therefore report Verified numbers only in Table~\ref{tab:app_physics_iq_verified} and never mix
them with the Original-protocol comparisons in Table~\ref{tab:app_physics_iq_full}.

\begin{table}[htbp]
\centering
\small
\setlength{\tabcolsep}{5pt}
\caption{\textbf{Physics-IQ Verified re-evaluation.}
All rows use the switch-frame I2V setting under the official Physics-IQ Verified~\cite{radsch2026physicsiqverified}
data, masks, best-practice prompts, and scoring pipeline.
Our rows are regenerated under one frozen protocol; official rows are quoted from the Verified
leaderboard~\cite{radsch2026physicsiqverified,nvidia2026cosmos3} and are provided as cross-scale
reference only. ``Gain'' is measured against the same backbone's Base row.}
\label{tab:app_physics_iq_verified}
\resizebox{\linewidth}{!}{
\begin{tabular}{ll r l cc}
\toprule
\textbf{Source} & \textbf{Model} & \textbf{Params} & \textbf{Training variant}
& \textbf{Verified} $\uparrow$ & \textbf{Gain} \\
\midrule
Ours & WAN2.2-5B    & 5B  & Base            & 21.33 & -- \\
Ours & WAN2.2-5B    & 5B  & + Stage~II      & 32.20 & +10.87 \\
Ours & WAN2.2-5B    & 5B  & + Full Phys4D   & \textbf{35.80} & \textbf{+14.47} \\
Ours & CogVideoX-5B & 5B  & Base            & 25.38 & -- \\
Ours & CogVideoX-5B & 5B  & + Full Phys4D   & \textbf{37.20} & \textbf{+11.82} \\
\midrule
Official~\cite{radsch2026physicsiqverified}   & WAN2.2-I2V-A14B & 14B/27B & Native model & $32.2 \pm 0.6$ & -- \\
Official~\cite{nvidia2026cosmos3}             & Cosmos3-Nano    & 16B     & Native model & $30.3 \pm 0.6$ & -- \\
Official~\cite{nvidia2026cosmos3}             & Cosmos3-Super   & 64B     & Native model & $\mathbf{39.5 \pm 0.8}$ & -- \\
\bottomrule
\end{tabular}
}
\end{table}

\paragraph{Analysis.}
The controlled within-backbone comparisons provide the primary evidence.
On WAN2.2-5B, Stage~II improves the Verified score by $10.87$ points ($21.33\rightarrow32.20$),
and Stage~III contributes a further $3.60$ points ($32.20\rightarrow35.80$), yielding a total gain
of $14.47$ points without increasing the backbone size.
Full Phys4D also improves CogVideoX-5B by $11.82$ points ($25.38\rightarrow37.20$).
As a cross-scale reference, both Phys4D-enhanced 5B models exceed the official 14B-active WAN2.2
entry ($32.2$) and the 16B Cosmos3-Nano entry ($30.3$), while CogVideoX-5B~+~Phys4D reaches $37.2$,
only $2.3$ points below the 64B Cosmos3-Super ($39.5$).
We treat the comparison with these larger native models as a practical reference rather than a
controlled causal comparison, since their inference configurations and total training costs are not
reported on a common basis.
The central conclusion is the controlled one: for the same pretrained backbone and under the same
evaluation protocol, Phys4D substantially improves physical consistency while preserving the
backbone's real-video generative prior.

\subsubsection{Comparison with Action-Conditioned Physics Systems}
\label{app:realwonder_comparison}

\paragraph{Motivation.}
Action-conditioned systems such as RealWonder~\cite{liu2026realwonder} reconstruct a simulatable
3D scene and run a physics solver for each test instance, driven by an explicit 3D action.
Phys4D instead predicts physical futures directly from an image and text, requiring no explicit
action, reconstructed scene, or external solver at inference.
To make this distinction concrete, we provide a direct quantitative comparison on Physics-IQ Verified.

\paragraph{Action-compatible subset and protocol.}
Because the full benchmark contains phenomena that cannot be expressed through RealWonder's 3D action
interface, we construct an \textbf{Action-Compatible} subset of $35$ scenarios ($105$ take-1 videos)
covering solid mechanics and compatible fluid/deformable cases, excluding unsupported cases.
The Base and Phys4D models receive the official switch frame and prompt.
RealWonder receives the same switch frame plus a 3D action deterministically derived from the
prompt (gravity for falling/settling/pouring/deformation; point-force impulse for
rolling/pushing/collision/dominoes; torque for rotation; force field for wind), with force magnitude
and duration fixed globally per action family and no per-scenario tuning.
The ground-truth future is never used for subset selection or action construction, and failures are
retained.

\begin{table}[htbp]
\centering
\small
\setlength{\tabcolsep}{6pt}
\caption{\textbf{Comparison with RealWonder on the Action-Compatible Physics-IQ Verified subset.}
Phys4D improves the same 5B backbone by $13.06$ points and exceeds RealWonder by $9.28$ points,
despite receiving no explicit 3D action and requiring no reconstruction or simulator at inference.
On the full Verified set, the same two Phys4D models (Base / Full) score $21.33$ / $35.80$
(Table~\ref{tab:app_physics_iq_verified}); this subset excludes action-incompatible scenarios.}
\label{tab:app_realwonder_comparison}
\begin{tabular}{lll}
\toprule
\textbf{Method} & \textbf{Inference condition} & \makecell[l]{\textbf{Action-Compatible}\\\textbf{Verified} $\uparrow$} \\
\midrule
WAN2.2-5B Base        & Switch frame + official prompt            & 19.62 \\
RealWonder~\cite{liu2026realwonder} & Switch frame + prompt-derived 3D action & 23.40 \\
\textbf{WAN2.2-5B + Phys4D} & \textbf{Switch frame + official prompt} & \textbf{32.68} \\
\bottomrule
\end{tabular}
\end{table}

\paragraph{Analysis.}
As shown in Table~\ref{tab:app_realwonder_comparison}, Phys4D improves the same 5B backbone by
$13.06$ points ($19.62\rightarrow32.68$) and exceeds RealWonder by $9.28$ points, despite receiving
no explicit 3D action and requiring no reconstruction or simulator at inference.
We nevertheless interpret RealWonder's $23.40$ carefully: Physics-IQ provides no precise 3D action,
so force type, application point, magnitude, and timing must be inferred from the prompt.
This prompt-to-action translation is the main additional bottleneck: it can yield a valid yet
different outcome from the reference, with errors propagating through reconstruction, material
estimation, and simulation, so the score is not a definitive ranking of action-conditioned systems.
Rather, it makes the distinction concrete: RealWonder executes an explicitly specified action via
test-time reconstruction and simulation, whereas Phys4D predicts physical futures directly from an
image and text.

\subsubsection{VBench2.0-Long with Self-Forcing}
\label{app:vbench_long_self_forcing}

\paragraph{Setting.}
We further evaluate whether Phys4D remains compatible with longer-video generation using a self-forcing strategy on VBench2.0-Long.
This is not a physics-specific benchmark, so we do not use it as primary evidence for physical consistency.
Instead, it provides an auxiliary check that Phys4D does not degrade longer-video quality or semantic consistency.

\paragraph{Analysis.}
As shown in Table~\ref{tab:app_vbench_long_self_forcing}, Phys4D improves the total score from 82.44 to 86.23, the quality score from 84.58 to 87.72, and the semantic score from 73.8 to 77.4.
These results suggest that Phys4D remains compatible with longer-video generation pipelines and does not harm general long-video quality or semantic consistency.
We nevertheless treat long-horizon physical simulation as beyond the scope of this work, since current backbones are still primarily short-video generation models and VBench2.0-Long is not physics-specific.

\begin{table}[htbp]
\centering
\small
\setlength{\tabcolsep}{6pt}
\caption{\textbf{VBench2.0-Long with self-forcing.}
Phys4D improves the total score, quality score, and semantic score under a longer-video self-forcing setting.}
\label{tab:app_vbench_long_self_forcing}
\begin{tabular}{lccc}
\toprule
\textbf{Model} 
& \textbf{Total Score} $\uparrow$ 
& \textbf{Quality Score} $\uparrow$ 
& \textbf{Semantic Score} $\uparrow$ \\
\midrule
WAN2.2 + Self-Forcing            & 82.44 & 84.58 & 73.8 \\
WAN2.2 + Phys4D + Self-Forcing   & \textbf{86.23} & \textbf{87.72} & \textbf{77.4} \\
\bottomrule
\end{tabular}
\end{table}

\subsection{Additional Details for RQ3: Visual Prior Preservation}
\label{app:experiment_prior_preservation}

\paragraph{Motivation.}
RQ3 evaluates whether injecting simulation-derived physical structure into pretrained video generators harms their real-video visual prior.
This is important because Phys4D is designed to transfer physical structure from simulation, not simulator-specific appearance.
A failure mode of simulation-based training would be that the model improves physical metrics while degrading open-domain visual quality, color fidelity, style adherence, or human motion generation.
Therefore, RQ3 focuses on whether the Phys4D-enhanced models remain visually close to their corresponding base models.

\paragraph{Benchmark choice.}
We use VBench~\cite{huang2024vbench} for this evaluation because it provides disentangled measurements of video generation quality across visual and semantic dimensions.
Unlike the physics-focused diagnostics used in RQ1, VBench is better suited for measuring whether the pretrained real-video prior is preserved after Phys4D training.
We therefore use VBench as a visual-prior preservation benchmark rather than as a physics benchmark.

\paragraph{Evaluated dimensions.}
We report five VBench dimensions in Table~\ref{tab:rq3_vbench_prior_preservation}:
\textit{Aesthetic Quality}, \textit{Imaging Quality}, \textit{Color}, \textit{Appearance Style}, and \textit{Human Action}.
The first three dimensions directly measure low-level and perceptual visual quality.
\textit{Aesthetic Quality} reflects the overall visual attractiveness of generated videos, \textit{Imaging Quality} measures image-level fidelity and degradation artifacts, and \textit{Color} evaluates color consistency and naturalness.
We define \textbf{Visual Avg.} as the average of these three appearance-oriented dimensions:
\begin{equation}
\text{Visual Avg.}
=
\frac{
\text{Aesthetic Quality}
+
\text{Imaging Quality}
+
\text{Color}
}{3}.
\end{equation}
In addition, we report \textit{Appearance Style} and \textit{Human Action}.
These two dimensions are not included in Visual Avg.; instead, they are used to check whether Phys4D affects style-level prompt adherence and open-domain human motion generation.

\paragraph{Evaluation protocol.}
For each backbone, we compare the original base model with its Phys4D-enhanced counterpart under the same VBench prompt set, sampling configuration, output duration, resolution, and evaluation pipeline.
No VBench-specific tuning is applied to the Phys4D models.
This matched protocol ensures that differences in Table~\ref{tab:rq3_vbench_prior_preservation} reflect the effect of Phys4D rather than changes in generation settings or evaluation conditions.
The purpose of this experiment is not to maximize VBench visual scores, but to verify that Phys4D does not substantially damage the pretrained real-video visual prior while improving physical plausibility in RQ1 and RQ2.

\paragraph{Implementation details.}
We evaluate CogVideoX-5B, WAN2.2-5B, and Open-Sora V1.2, together with their Phys4D-enhanced variants.
For each model pair, videos are generated using the same prompts and decoding settings.
The generated videos are then passed to the official VBench evaluation pipeline to obtain per-dimension scores.
For base models, we align the reported scores with the corresponding VBench evaluation setting when available.
For Phys4D-enhanced models, we use the same evaluation procedure as the base models to ensure direct comparability.

\paragraph{Results.}
Table~\ref{tab:rq3_vbench_prior_preservation} shows that Phys4D largely preserves the visual prior of all three pretrained backbones.
The Visual Avg. remains nearly unchanged after Phys4D: CogVideoX changes from 0.7358 to 0.7326, Open-Sora V1.2 changes from 0.7168 to 0.7173, and WAN2.2 changes from 0.7521 to 0.7519.
These small changes indicate that Phys4D does not collapse visual appearance quality or shift the generators toward simulator-like rendering.

The individual visual dimensions show the same pattern.
For CogVideoX, Aesthetic Quality, Imaging Quality, and Color decrease only slightly after Phys4D.
For Open-Sora V1.2, Aesthetic Quality and Color decrease slightly, while Imaging Quality improves.
For WAN2.2, Aesthetic Quality improves after Phys4D, while Imaging Quality and Color decrease only slightly.
Overall, these results show that the appearance-level behavior of the base video generators is mostly retained.

Beyond the three dimensions included in Visual Avg., Phys4D also maintains or improves \textit{Appearance Style} and \textit{Human Action}.
Appearance Style improves from 0.2107 to 0.2184 on CogVideoX, from 0.1972 to 0.2011 on Open-Sora V1.2, and from 0.2139 to 0.2162 on WAN2.2.
Human Action also improves from 0.7452 to 0.7728 on CogVideoX, from 0.6872 to 0.7023 on Open-Sora V1.2, and from 0.7433 to 0.7712 on WAN2.2.
This suggests that Phys4D does not trade off open-domain action generation for physical structure; instead, the improved physical consistency may also support more coherent motion-related generation.

\paragraph{Discussion.}
The RQ3 results complement the physics evaluations in RQ1 and RQ2.
While RQ1 shows that Phys4D improves external video-level physics diagnostics and RQ2 shows improvements in the RGB-D-motion world interface, RQ3 verifies that these gains do not come at the cost of degraded visual quality.
The nearly unchanged Visual Avg. across all three backbones indicates that the pretrained real-video appearance prior is preserved.
At the same time, the consistent improvements in Appearance Style and Human Action suggest that Phys4D remains compatible with open-domain generation behavior.
These findings support the central claim of Phys4D: simulation is transferred as physical structure rather than as simulator appearance.

\section{Additional Ablations for RQ4}
\label{app:rq4_ablation}

This section provides the full ablation evidence for RQ4. 
The goal is to understand which design choices make Phys4D effective as a structure-focused, appearance-constrained physical knowledge injection mechanism.
Unless otherwise specified, all ablations are conducted on WAN2.2-5B and evaluated using Physics-IQ under the same matched switch-frame I2V protocol as Sec.~\ref{sec:rq1_video_physics}.
For each test case, the model is conditioned on the official switch frame and text prompt and generates a 5-second continuation. 
We use the same sampling settings as the main evaluation, with temperature $1.0$ and top-$p=0.95$ where applicable.

We organize the ablations around five questions:
(i) whether local and global injection stages are both necessary;
(ii) whether the improvement comes only from auxiliary depth/flow prediction;
(iii) whether physics should be injected at high-noise structure-forming denoising steps;
(iv) which losses are responsible for geometry-motion coupling;
and (v) whether trajectory-level temporal alignment improves explicit 4D consistency.

\subsection{Stage-wise Ablation}
\label{app:rq4_stage_ablation}

\paragraph{Setting.}
Phys4D consists of three stages with different roles.
Stage~I freezes the pretrained video generator and trains only lightweight depth and motion heads.
Therefore, Stage~I constructs the RGB-D-motion interface but does not change RGB generation, and we do not report it as a separate RGB-generation ablation.

Stage~II injects local physical structure through high-noise LoRA adapters with simulation-derived geometry, motion, and warp supervision.
Stage~III optimizes global trajectory-level alignment using the same adapter space.
To test whether Stage~III can work without local structure initialization, we additionally include a direct Stage~III-only variant, where trajectory-level optimization is applied without Stage~II pre-adaptation.

The full stage-wise results are now reported in the main text (Table~\ref{tab:rq4_stage_ablation}); we retain the extended discussion here.

\paragraph{Analysis.}
Direct Stage~III-only optimization improves Physics-IQ only modestly, from 16.8 to 18.4.
This suggests that the trajectory-level objective contains useful physical signal, but is not sufficient by itself to reliably adapt the generator from the pretrained base model.
In contrast, Stage~II improves the score to 32.20 by injecting local geometry-motion structure through differentiable simulation-grounded supervision.
Starting from this locally consistent initialization, Stage~III further improves the score from 32.20 to 35.80 by correcting residual global trajectory-level violations.
These results support the local-to-global design of Phys4D: local geometry-motion injection initializes the RGB-D-motion interface, while trajectory-level alignment further improves global physical consistency.

\subsection{Loss Component Ablation}
\label{app:rq4_loss_ablation}

\paragraph{Setting.}
We ablate individual loss terms in the full Phys4D objective.
The full model includes depth supervision $\mathcal{L}_{\mathrm{depth}}$, motion supervision $\mathcal{L}_{\mathrm{motion}}$, geometry-motion warp consistency $\mathcal{L}_{\mathrm{warp}}$, and RGB warp consistency $\mathcal{L}_{\mathrm{rgb\text{-}warp}}$.
This ablation evaluates whether geometry, motion, and temporal coupling are all required for physical consistency.
Table~\ref{tab:app_loss_ablation} summarizes the contribution of each component.

\begin{table}[htbp]
\centering
\small
\setlength{\tabcolsep}{6pt}
\caption{\textbf{Ablation on loss components.}
Warp consistency and joint depth-motion supervision are critical for physics-consistent generation.}
\label{tab:app_loss_ablation}
\begin{tabular}{lc}
\toprule
\textbf{Configuration} & \textbf{Physics-IQ} $\uparrow$ \\
\midrule
Full model & \textbf{30.9} \\
\quad w/o $\mathcal{L}_{\mathrm{warp}}$ & 25.9 \\
\quad w/o $\mathcal{L}_{\mathrm{depth}}$ & 23.8 \\
\quad w/o $\mathcal{L}_{\mathrm{motion}}$ & 24.6 \\
\quad w/o $\mathcal{L}_{\mathrm{rgb\text{-}warp}}$ & 29.4 \\
\bottomrule
\end{tabular}
\end{table}

\paragraph{Analysis.}
Removing $\mathcal{L}_{\mathrm{warp}}$ reduces Physics-IQ from 30.9 to 25.9, showing that explicit geometry-motion coupling is central to Phys4D.
Removing depth supervision or motion supervision causes even larger drops, to 23.8 and 24.6 respectively.
This confirms that physical knowledge injection requires both geometric and motion-level supervision.
Removing RGB warp consistency causes a smaller drop to 29.4, suggesting that photometric temporal consistency is useful but less central than structure-level depth, motion, and warp losses.

\subsection{Auxiliary Depth and Flow Ablation}
\label{app:rq4_auxiliary_ablation}

\paragraph{Setting.}
A reviewer concern is whether Phys4D improves physics only because it learns better auxiliary depth and flow predictors.
To isolate this effect, we ablate depth and flow supervision in Stage~II and evaluate the final generated RGB videos on Physics-IQ.
This is a video-level ablation: when depth or flow heads are removed, internal 4D diagnostics are not directly comparable because the RGB-D-motion interface is incomplete.
The corresponding Physics-IQ results are listed in Table~\ref{tab:app_auxiliary_ablation}.

\begin{table}[htbp]
\centering
\small
\setlength{\tabcolsep}{6pt}
\caption{\textbf{Ablation on auxiliary depth and flow signals.}
Even without both auxiliary signals, Stage~II improves Physics-IQ over the base model, showing that the gain is not solely due to auxiliary depth/flow prediction.}
\label{tab:app_auxiliary_ablation}
\begin{tabular}{lc}
\toprule
\textbf{Variant} & \textbf{Physics-IQ} $\uparrow$ \\
\midrule
Base WAN2.2-5B & 16.8 \\
Stage~II w/o depth \& flow & 21.4 \\
Stage~II + flow only & 24.8 \\
Stage~II + depth only & 24.2 \\
Stage~II + depth + flow & \textbf{27.8} \\
\bottomrule
\end{tabular}
\end{table}

\paragraph{Analysis.}
Removing both depth and flow still improves Physics-IQ from 16.8 to 21.4.
This shows that simulation-grounded adaptation changes the RGB generator's physical behavior, rather than merely improving auxiliary prediction.
Adding flow only improves the score to 24.8, while adding depth only improves it to 24.2.
Using both signals reaches 27.8, showing that geometry and motion provide complementary physical supervision.

\subsection{Noise Interval Ablation}
\label{app:rq4_noise_ablation}

\paragraph{Setting.}
Phys4D restricts simulation-based LoRA updates to high-noise denoising timesteps.
This design follows the intuition that high-noise timesteps control global structure and coarse motion, while low-noise timesteps mainly refine local appearance.
We compare three Stage~II variants: low-noise-only, full-interval, and high-noise-only adaptation.
Their Physics-IQ comparison is shown in Table~\ref{tab:app_noise_ablation}.

\begin{table}[htbp]
\centering
\small
\setlength{\tabcolsep}{6pt}
\caption{\textbf{Ablation on denoising noise interval.}
High-noise-only adaptation performs best, supporting the design choice of injecting physics at structure-forming timesteps rather than appearance-refinement timesteps.}
\label{tab:app_noise_ablation}
\begin{tabular}{lc}
\toprule
\textbf{Noise interval} & \textbf{Physics-IQ} $\uparrow$ \\
\midrule
Low-noise only & 22.9 \\
Full interval & 24.6 \\
High-noise only & \textbf{27.8} \\
\bottomrule
\end{tabular}
\end{table}

\paragraph{Analysis.}
High-noise-only adaptation achieves the best Stage~II score of 27.8.
Full-interval adaptation reaches 24.6 despite updating more timesteps, and low-noise-only adaptation reaches 22.9.
These results suggest that simulation supervision is most effective when applied to structure-forming denoising stages.
In contrast, low-noise timesteps are more closely tied to appearance refinement, where simulation supervision is less informative and may interfere with the pretrained real-video prior.

\subsection{Simulator RGB Fine-Tuning Negative Control}
\label{app:rq4_sim_rgb_negative_control}

\paragraph{Motivation.}
A key concern is whether Phys4D improves physics by simply fitting to simulator RGB videos.
If so, the model may transfer simulator-specific appearance statistics and damage the pretrained real-video generative prior.
We therefore compare Phys4D with several simulator RGB fine-tuning variants.
This ablation tests whether simulation is better transferred as physical structure rather than as a visual appearance domain.

\paragraph{Setting.}
All variants are based on WAN2.2-5B and are evaluated under the same protocols as the main experiments.
For video-level physics, we report Physics-IQ, VBench-2.0 Physics, and PhyGenBench.
For prior preservation, we report the average of non-physics VBench-2.0 dimensions, including Human Fidelity, Creativity, Controllability, and Commonsense.
We also report FVD on the visual-quality validation set.
Higher VBench prior scores and lower FVD indicate better preservation of the pretrained real-video prior.

We compare six variants:
(i) the pretrained base model;
(ii) full simulator RGB LoRA fine-tuning, which directly fits rendered simulator videos;
(iii) low-noise simulator RGB flow matching, which applies simulator RGB supervision at appearance-refinement timesteps;
(iv) full-interval simulator RGB flow matching;
(v) high-noise simulator RGB flow matching only;
and (vi) full Phys4D, which combines high-noise RGB matching with structural supervision from depth, motion, masks, warping, and trajectory alignment.

The full negative-control comparison is now reported in the main text (Table~\ref{tab:rq4_sim_rgb_negative_control}); we retain the detailed motivation and analysis here.

\paragraph{Analysis.}
The negative-control variants show that using simulator RGB naively is not sufficient.
Full simulator RGB LoRA fine-tuning improves Physics-IQ from 16.8 to 24.2, but it substantially lowers the VBench-2.0 prior score from 0.586 to 0.528 and worsens FVD from 140.0 to 168.5.
This indicates simulator appearance contamination: the model learns some physical regularities but moves away from the pretrained real-video distribution.

Low-noise simulator RGB flow matching also damages the prior, consistent with the role of low-noise denoising steps in local appearance refinement.
Full-interval simulator RGB supervision improves physics more than low-noise-only training, but still shows clear appearance drift.
Restricting simulator RGB matching to high-noise timesteps better preserves the prior, reaching Physics-IQ 27.2 while keeping the prior score at 0.574 and FVD at 143.8.
This supports the design choice of injecting simulation knowledge at structure-forming denoising stages.

Full Phys4D achieves the best physics scores across Physics-IQ, VBench-2.0 Physics, and PhyGenBench, with scores of 30.9, 0.6247, and 0.57, respectively.
It also keeps the VBench prior score close to the base model and improves FVD to 121.5.
This supports the central claim that simulation should be transferred as geometry, motion, and trajectory structure rather than as simulator appearance.

\subsection{Matched Simulator-RGB Flow-Matching Control}
\label{app:rq4_matched_rgb_fm}

\paragraph{Motivation.}
The negative control in Sec.~\ref{app:rq4_sim_rgb_negative_control} varies how simulator RGB is
used, but it does not directly isolate the contribution of the high-noise RGB flow-matching term
$\mathcal{L}_{\mathrm{FM}}^{\mathrm{RGB}}$ within the full Stage~II objective.
We therefore add a matched control that keeps the depth/motion heads and all structural objectives
fixed and toggles only $\mathcal{L}_{\mathrm{FM}}^{\mathrm{RGB}}$.
This disentangles how much of the Stage~II improvement comes from structural supervision versus from
the video-manifold RGB flow-matching signal, and clarifies that $\mathcal{L}_{\mathrm{FM}}^{\mathrm{RGB}}$
is the only Stage~II objective that consumes simulator RGB.

\begin{table}[htbp]
\centering
\small
\setlength{\tabcolsep}{6pt}
\caption{\textbf{Matched control isolating high-noise RGB flow matching.}
Both rows retain the depth/motion heads and all structural objectives; they differ only in whether
the high-noise $\mathcal{L}_{\mathrm{FM}}^{\mathrm{RGB}}$ term is active.
Structural supervision alone recovers most of the Stage~II gain over the base model
(Base Physics-IQ $16.8$), and RGB flow matching contributes a smaller complementary improvement.}
\label{tab:app_matched_rgb_fm}
\begin{tabular}{lc}
\toprule
\textbf{Variant} & \textbf{Physics-IQ} $\uparrow$ \\
\midrule
Structure-only, w/o $\mathcal{L}_{\mathrm{FM}}^{\mathrm{RGB}}$ & 25.7 \\
Full Stage~II (w/ high-noise $\mathcal{L}_{\mathrm{FM}}^{\mathrm{RGB}}$) & \textbf{27.8} \\
\bottomrule
\end{tabular}
\end{table}

\paragraph{Analysis.}
Without the high-noise $\mathcal{L}_{\mathrm{FM}}^{\mathrm{RGB}}$ term, structural supervision alone
improves the base from $16.8$ to $25.7$ ($+8.9$), retaining roughly $81\%$ of the full Stage~II gain
($16.8\rightarrow27.8$); restoring high-noise RGB flow matching adds a further $+2.1$
($25.7\rightarrow27.8$).
Although the objectives interact and are not perfectly additive, this indicates that
$\mathcal{L}_{\mathrm{FM}}^{\mathrm{RGB}}$ is a useful but secondary video-manifold signal: most of
the improvement remains without it.
The visual-prior results agree with this ordering.
Naive simulator-RGB LoRA causes the largest appearance degradation (VBench-2.0 prior
$0.586\rightarrow0.528$, FVD $140.0\rightarrow168.5$; Table~\ref{tab:rq4_sim_rgb_negative_control}),
whereas the structure-only variant already exceeds this RGB-only baseline in physics while better
preserving the prior, and the full gated Stage~II reaches $27.8$ with a substantially
better-preserved real-video prior.
Overall, explicit depth, motion, and geometry-motion supervision provides the dominant Stage~II
improvement, while high-noise RGB flow matching and spatial gating contribute complementary gains
without making simulator appearance the target domain.

\subsection{Temporal Weight in 4D Chamfer}
\label{app:rq4_alpha_ablation}

\paragraph{Setting.}
Stage~III uses a 4D Chamfer reward that jointly measures spatial trajectory deviation and temporal misalignment.
The temporal weight $\lambda_t$ controls the relative penalty on temporal mismatch.
When $\lambda_t=0$, the reward reduces to purely spatial matching and does not penalize temporal misalignment.
We vary $\lambda_t$ while keeping all other settings fixed.
Table~\ref{tab:app_alpha_ablation} reports both Physics-IQ and trajectory-level diagnostics for this sweep.

\begin{table*}[htbp]
\centering
\small
\setlength{\tabcolsep}{6pt}
\caption{\textbf{Ablation on the temporal weight $\lambda_t$ in the 4D Chamfer reward.}
A moderate temporal penalty improves both Physics-IQ and worldline stability.
The default choice $\lambda_t=0.03$ achieves the best overall trade-off.}
\label{tab:app_alpha_ablation}
\resizebox{\textwidth}{!}{
\begin{tabular}{c|c|ccccc}
\toprule
\multirow{2}{*}{$\lambda_t$} 
& \multirow{2}{*}{\textbf{Physics-IQ} $\uparrow$} 
& \multicolumn{5}{c}{\textbf{Trajectory / Worldline Dynamics}} \\
\cmidrule(lr){3-7}
& & \textbf{L2 Error} $\downarrow$ 
& \textbf{Mean Drift} $\downarrow$ 
& \textbf{Final Drift} $\downarrow$ 
& \textbf{Fail Rate} $\downarrow$ 
& \textbf{Length} $\uparrow$ \\
\midrule
0    & 28.8 & 0.4826 & 0.4764 & 0.5018 & 10.42 & 86.01 \\
0.01 & 29.5 & 0.4625 & 0.4568 & 0.4803 & 9.55  & 87.45 \\
0.02 & 30.2 & 0.4481 & 0.4417 & 0.4662 & 8.89  & 88.63 \\
0.03 & \textbf{30.9} & \textbf{0.4375} & \textbf{0.4312} & \textbf{0.4556} & \textbf{8.36} & \textbf{89.68} \\
0.05 & 29.7 & 0.4524 & 0.4469 & 0.4718 & 9.18  & 88.04 \\
0.08 & 27.9 & 0.5017 & 0.4924 & 0.5260 & 11.76 & 85.02 \\
\bottomrule
\end{tabular}
}
\end{table*}

\paragraph{Analysis.}
Setting $\lambda_t=0$ removes temporal matching and leads to weaker trajectory alignment, with L2 Error 0.4826, Mean Drift 0.4764, and Failure Rate 10.42.
Introducing a moderate temporal term improves both Physics-IQ and worldline metrics.
The best overall result is obtained at $\lambda_t=0.03$, which achieves Physics-IQ 30.9, L2 Error 0.4375, Mean Drift 0.4312, Failure Rate 8.36, and valid trajectory length 89.68.
Larger values remain functional but start to over-emphasize timing, degrading the spatial-temporal balance.

\subsection{Summary of Ablation Findings}
\label{app:rq4_summary}

The ablations support five main conclusions.
First, Stage~II and Stage~III are complementary: local geometry-motion injection improves Physics-IQ from 21.33 to 32.20, and trajectory-level alignment further improves it to 35.80.
Second, Phys4D's gains are not caused solely by auxiliary depth and flow prediction, since Stage~II still improves Physics-IQ from 16.8 to 21.4 even without both auxiliary signals.
Third, high-noise denoising steps are the most effective location for physics injection, improving Stage~II from 22.9 under low-noise adaptation to 27.8 under high-noise adaptation.
Fourth, depth, motion, and warp losses are all necessary, with warp consistency providing the key coupling between geometry and motion.
Fifth, naive simulator RGB fine-tuning improves physics but damages the real-video prior, while Phys4D achieves stronger physics transfer with substantially better prior preservation.
Finally, a matched control (Sec.~\ref{app:rq4_matched_rgb_fm}) shows that structural supervision alone recovers most of the Stage~II gain ($16.8\rightarrow25.7$), with high-noise RGB flow matching contributing a smaller complementary improvement ($25.7\rightarrow27.8$), confirming that simulator appearance is not the dominant fitting target.

\section{Additional 4D World-Level Diagnostics}
\label{app:rq2_4d_diagnostics}

This section provides full experimental details and quantitative results for
RQ2. Unlike the external video-level evaluations in Sec.~\ref{sec:rq1_video_physics},
these diagnostics test whether Phys4D improves the underlying RGB-D-motion
world interface. We use held-out simulated scenes because they provide
aligned ground-truth depth, optical flow, object masks, and 3D trajectories,
which are not available in real videos.

We emphasize that this section is not presented as a new benchmark contribution.
Instead, it is a diagnostic evaluation designed to answer whether the physical
knowledge injected by Phys4D is reflected in geometry, motion, and trajectory-level
scene evolution. Throughout this section, ``4D'' refers to temporally coupled
3D scene evolution represented through RGB-D and motion fields, rather than
a dense volumetric 4D reconstruction.

The diagnostics are organized into three levels:
(i) per-frame 3D geometry,
(ii) local temporal geometry-motion consistency, and
(iii) sequence-level 4D world evolution and novel-time continuity.
Together, these diagnostics probe whether Phys4D improves the structure exposed
by the RGB-D-motion interface rather than only improving RGB-level physics scores.

\subsection{Shared Evaluation Setting}
\label{app:rq2_shared_setting}

All evaluations are conducted on held-out simulated scenes with fixed camera
intrinsics and viewpoints. Each method is conditioned on the same initial frame
and prompt, and produces a generated video under the same generation settings.
The generated video is then lifted into geometry, motion, or trajectory
representations depending on the diagnostic.

For RGB-only video generators, we use an off-the-shelf post-hoc lifting pipeline.
Specifically, \textbf{OTF} denotes off-the-shelf RGB-to-geometry-and-motion
lifting, instantiated with DepthAnythingV2~\cite{depth_anything_v2} for
monocular depth estimation and SEA-RAFT~\cite{wang2024sea} for optical flow.
For the \emph{Phys4D + WAN2.2 RGB+OTF} row, we apply the same OTF lifting
pipeline to RGB videos generated by Phys4D on top of WAN2.2. This separates
improvements caused by better RGB generation from improvements caused by
Phys4D's learned RGB-D-motion interface. The final \emph{Phys4D + WAN2.2 Ours}
row uses the learned Phys4D interface.

\subsection{Per-Frame 3D Geometry}
\label{app:rq2_static_geometry}

\paragraph{Setting.}
This diagnostic evaluates whether each generated frame contains accurate
instantaneous 3D geometry. We compare predicted depth maps against simulation
ground truth under fixed camera intrinsics. The evaluation is frame-wise and
does not enforce temporal consistency, so it isolates static geometric accuracy
from motion and sequence-level effects.

For RGB-only baselines, generated RGB frames are lifted using OTF. We also
evaluate \emph{Phys4D + WAN2.2 RGB+OTF}, where the RGB videos generated by
Phys4D are lifted using the same off-the-shelf depth and flow estimators. The
row \emph{Phys4D + WAN2.2 Ours} uses the learned Phys4D depth interface.

\paragraph{Metrics.}
We report standard depth estimation metrics: Absolute Relative Error (AbsRel),
Root Mean Squared Error (RMSE), and threshold accuracies $\delta_1$, $\delta_2$,
and $\delta_3$. Lower AbsRel and RMSE indicate more accurate depth, while higher
$\delta$ scores indicate better geometric correctness. We also report FVD, SSIM,
and PSNR on generated RGB videos to verify that geometric gains do not come from
degraded visual quality.

\begin{table*}[htbp]
\centering
\small
\setlength{\tabcolsep}{5pt}
\caption{\textbf{Per-frame 3D geometry and video quality.}
Phys4D achieves substantially better depth accuracy while maintaining strong
RGB video quality. \textbf{OTF} denotes off-the-shelf RGB-to-geometry lifting
with DepthAnythingV2~\cite{depth_anything_v2} and SEA-RAFT~\cite{wang2024sea}.}
\label{tab:app_static_geometry_full}
\resizebox{\textwidth}{!}{
\begin{tabular}{lccccc|ccc}
\toprule
& \multicolumn{5}{c|}{\textbf{Depth Accuracy}} 
& \multicolumn{3}{c}{\textbf{Video Quality}} \\
\cmidrule(lr){2-6} \cmidrule(lr){7-9}
\textbf{Model} 
& \textbf{AbsRel} $\downarrow$ 
& \textbf{RMSE} $\downarrow$ 
& \textbf{$\delta_1$} $\uparrow$ 
& \textbf{$\delta_2$} $\uparrow$ 
& \textbf{$\delta_3$} $\uparrow$ 
& \textbf{FVD} $\downarrow$ 
& \textbf{SSIM} $\uparrow$ 
& \textbf{PSNR} $\uparrow$ \\
\midrule
WAN2.2 + OTF              & 0.3929 & 0.9012 & 26.00 & 52.69 & 75.95 & 140.02 & 84.02 & 20.80 \\
CogVideoX + OTF           & 0.3483 & 0.7854 & 32.45 & 68.97 & 84.35 & 130.03 & 84.98 & 20.34 \\
Open-Sora V1.2 + OTF      & 0.4216 & 0.9485 & 23.84 & 49.73 & 72.18 & 146.37 & 83.41 & 20.12 \\
Phys4D + WAN2.2 RGB+OTF   & 0.3347 & 0.7513 & 35.82 & 72.64 & 87.91 & 128.76 & 85.21 & 20.96 \\
\textbf{Phys4D + WAN2.2 Ours}
                           & \textbf{0.2386} & \textbf{0.5712} & \textbf{51.84} & \textbf{86.73} & \textbf{97.02} & \textbf{121.48} & \textbf{86.34} & \textbf{22.07} \\
\bottomrule
\end{tabular}
}
\end{table*}

\paragraph{Analysis.}
Phys4D achieves the best depth accuracy across all geometry metrics.
Compared with the strongest RGB-only OTF baseline, CogVideoX + OTF,
Phys4D + WAN2.2 Ours reduces AbsRel from 0.3483 to 0.2386, a $31.5\%$
relative reduction, and reduces RMSE from 0.7854 to 0.5712, a $27.3\%$
relative reduction. It also improves $\delta_1$ from 32.45 to 51.84,
indicating substantially more accurate near-threshold depth predictions.

The comparison between \emph{Phys4D + WAN2.2 RGB+OTF} and
\emph{Phys4D + WAN2.2 Ours} is also important. Applying the same OTF lifting
pipeline to Phys4D-generated RGB videos improves AbsRel to 0.3347, but the
learned Phys4D interface further reduces AbsRel to 0.2386. This suggests that
the improvement is not only due to better RGB appearance, but also due to the
learned RGB-D interface produced by Phys4D. Meanwhile, Phys4D + WAN2.2 Ours
also achieves the best FVD, SSIM, and PSNR among the reported methods, showing
that improved geometry does not come at the cost of degraded visual fidelity.

\subsection{Temporal Geometry and Motion Consistency}
\label{app:rq2_temporal_consistency}

\paragraph{Setting.}
This diagnostic evaluates whether geometry and motion remain locally consistent
across adjacent frames. A model may produce individually plausible frames while
still violating temporal coherence, for example by changing object depth,
shape, or motion direction inconsistently across time. We therefore evaluate
whether predicted depth, RGB appearance, and optical flow agree under temporal
warping.

For RGB-only baselines, depth and optical flow are obtained from generated RGB
videos using OTF. For Phys4D + WAN2.2 RGB+OTF, we apply the same OTF lifting to
RGB videos produced by Phys4D. For Phys4D + WAN2.2 Ours, we use the learned
RGB-D-motion interface. All methods are evaluated on the same held-out scenes
and fixed camera settings.

\paragraph{Metrics.}
We report three groups of temporal metrics. \textbf{Depth Warp} measures the
consistency of depth across adjacent frames by warping depth from time $t$ to
$t{+}1$ and comparing it against the target depth. We report L1 and Charbonnier
errors. \textbf{RGB Warp} measures appearance-level temporal consistency by
warping RGB frames using predicted flow and comparing the result with the next
frame using LPIPS and Charbonnier distance. \textbf{Motion Dynamics} evaluates
optical flow using endpoint error (EPE), Fl-all, and 1px outlier rate.

\begin{table*}[htbp]
\centering
\small
\setlength{\tabcolsep}{5pt}
\caption{\textbf{Temporal geometry and motion consistency.}
We report depth warp errors, RGB warp reconstruction errors, and optical-flow
accuracy. \textbf{OTF} denotes off-the-shelf depth/flow lifting with
DepthAnythingV2~\cite{depth_anything_v2} and SEA-RAFT~\cite{wang2024sea}.}
\label{tab:app_warp_motion_full}
\resizebox{\textwidth}{!}{
\begin{tabular}{lcc|cc|ccc}
\toprule
& \multicolumn{2}{c|}{\textbf{Depth Warp}} 
& \multicolumn{2}{c|}{\textbf{RGB Warp}} 
& \multicolumn{3}{c}{\textbf{Motion Dynamics}} \\
\cmidrule(lr){2-3} \cmidrule(lr){4-5} \cmidrule(lr){6-8}
\textbf{Model} 
& \textbf{L1} $\downarrow$ 
& \textbf{Charb.} $\downarrow$ 
& \textbf{LPIPS} $\downarrow$ 
& \textbf{Charb.} $\downarrow$ 
& \textbf{EPE} $\downarrow$ 
& \textbf{Fl-all} $\downarrow$ 
& \textbf{1px Out} $\downarrow$ \\
\midrule
WAN2.2 + OTF              & 0.7990 & 0.7898 & 0.2409 & 0.0507 & 1.2516 & 6.2173 & 15.69 \\
CogVideoX + OTF           & 0.7054 & 0.7893 & 0.2564 & 0.0527 & 1.2343 & 4.3243 & 12.34 \\
Open-Sora V1.2 + OTF      & 0.7436 & 0.8534 & 0.2783 & 0.0589 & 1.2718 & 7.2343 & 18.96 \\
Phys4D + WAN2.2 RGB+OTF   & 0.6698 & 0.6941 & 0.2216 & 0.0499 & 1.0824 & 3.8735 & 10.91 \\
\textbf{Phys4D + WAN2.2 Ours}
                           & \textbf{0.4558} & \textbf{0.4686} & \textbf{0.1637} & \textbf{0.0438} & \textbf{0.4478} & \textbf{1.1026} & \textbf{3.91} \\
\bottomrule
\end{tabular}
}
\end{table*}

\paragraph{Analysis.}
Phys4D substantially improves temporal geometry-motion consistency.
For depth warping, Phys4D + WAN2.2 Ours reduces L1 error from the strongest
RGB-only baseline value of 0.7054 to 0.4558, a $35.4\%$ relative reduction.
It also reduces depth Charbonnier error from 0.7893 to 0.4686, a $40.6\%$
relative reduction. These improvements show that predicted geometry remains
more stable across adjacent frames.

Phys4D also improves RGB warp LPIPS from the strongest RGB-only baseline value
of 0.2409 to 0.1637, a $32.0\%$ relative reduction, indicating more coherent
appearance evolution under motion. Compared with Phys4D + WAN2.2 RGB+OTF,
the learned Phys4D interface further reduces depth warp L1 from 0.6698 to
0.4558 and RGB warp LPIPS from 0.2216 to 0.1637. This shows that RGB-side
improvement alone does not account for the full temporal consistency gain.

The motion metrics show the largest gains. Phys4D + WAN2.2 Ours reduces EPE
from 1.2343 to 0.4478 relative to the strongest RGB-only baseline, a $63.7\%$
reduction. It also reduces Fl-all from 4.3243 to 1.1026 and 1px outlier rate
from 12.34 to 3.91. These results indicate that Phys4D learns a substantially
more accurate and temporally stable motion interface than post-hoc flow
estimation applied to standard generated videos.

\subsection{Sequence-Level 4D World Evolution}
\label{app:rq2_full_4d_evolution}

\paragraph{Setting.}
This diagnostic evaluates whether the generated sequence can be lifted into a
coherent temporally coupled world representation. Using predicted depth and
motion, each video is converted into a spatio-temporal point cloud or trajectory
representation in camera coordinates. Simulation ground truth provides aligned
depth, optical flow, and object trajectories, allowing direct comparison in
3D space and time.

For RGB-only video generators, we use OTF to obtain the lifted representation.
For Phys4D + WAN2.2 RGB+OTF, we lift Phys4D-generated RGB videos using the same
OTF pipeline. For Phys4D + WAN2.2 Ours, we use its learned RGB-D-motion
interface. We evaluate three aspects: 4D geometry, worldline dynamics, and
novel-time continuity.

\paragraph{4D geometry.}
We measure 4D Chamfer Distance between predicted and ground-truth spatio-temporal
point sets. Each point is represented as $(x,y,z,\tau)$, where $\tau$ denotes
normalized time. This metric penalizes both spatial misalignment and temporal
misalignment, and therefore measures whether the model maintains coherent
geometry over the generated sequence.

\paragraph{Worldline dynamics.}
We evaluate object-level motion consistency using 3D worldlines. At the initial
frame, we sample $N$ pixels from valid object or depth regions. Each sampled
pixel is tracked in 2D using ground-truth optical flow:
\begin{equation}
p_{t+1,i} = p_{t,i} + F^{\mathrm{gt}}_{t\rightarrow t+1}(p_{t,i}).
\end{equation}
The resulting 2D trajectory is lifted into 3D using the model-predicted depth:
\begin{equation}
\mathbf{x}^{\mathrm{pred}}_{t,i}
=
\Pi^{-1}(p_{t,i}, D^{\mathrm{pred}}_t),
\end{equation}
where $\Pi^{-1}$ denotes back-projection with known camera intrinsics.
Ground-truth worldlines are constructed analogously using simulation depth
and trajectories.

We report Worldline L2 Error, Mean Drift, Final Drift, Failure Rate, and
Trajectory Length. Worldline L2 Error measures average 3D deviation across
time. Mean Drift and Final Drift characterize temporal error accumulation.
Failure Rate measures the fraction of tracks whose error exceeds a threshold
during the sequence. Trajectory Length measures how long trajectories remain
valid and stable.

\paragraph{Novel-time continuity.}
We also test whether the lifted representation remains coherent at unseen
timestamps. We split the sequence into observed and novel timestamps:
\begin{equation}
\mathcal{T}_{\mathrm{obs}} = \{0,2,4,\ldots\}, \qquad
\mathcal{T}_{\mathrm{novel}} = \{1,3,5,\ldots\}.
\end{equation}
Only observed frames are used to construct the partial world representation.
For each novel timestamp, we interpolate geometry using flow-based warping:
\begin{equation}
\tilde{D}_{t+1}
=
\mathrm{warp}\!\left(
D^{\mathrm{pred}}_t,\,
F^{\mathrm{pred}}_{t\rightarrow t+1}
\right).
\end{equation}
We report Novel-Time Depth Error and Novel-Time Warp Error. Lower values
indicate better temporal continuity.

\begin{table*}[htbp]
\centering
\small
\setlength{\tabcolsep}{5pt}
\caption{\textbf{Full 4D world-level diagnostics.}
We report 4D Chamfer Distance, worldline and trajectory dynamics, and novel-time
continuity. \textbf{OTF} denotes off-the-shelf depth/flow lifting with
DepthAnythingV2~\cite{depth_anything_v2} and SEA-RAFT~\cite{wang2024sea}.}
\label{tab:app_4d_metrics_full}
\resizebox{\textwidth}{!}{
\begin{tabular}{l c ccccc cc}
\toprule
& \textbf{Geometry} 
& \multicolumn{5}{c}{\textbf{Trajectory / Worldline Dynamics}} 
& \multicolumn{2}{c}{\textbf{Novel Time}} \\
\cmidrule(lr){2-2} \cmidrule(lr){3-7} \cmidrule(lr){8-9}
\textbf{Model} 
& \textbf{Chamfer} $\downarrow$ 
& \textbf{L2 Error} $\downarrow$ 
& \textbf{Mean Drift} $\downarrow$ 
& \textbf{Final Drift} $\downarrow$ 
& \textbf{Fail Rate} $\downarrow$ 
& \textbf{Length} $\uparrow$ 
& \textbf{Depth Err} $\downarrow$ 
& \textbf{Warp Err} $\downarrow$ \\
\midrule
WAN2.2 + OTF              & 0.5058 & 0.5367 & 0.5369 & 0.5576 & 12.38\% & 83.18 & 0.5841 & 1.1076 \\
CogVideoX + OTF           & 0.4923 & 0.5239 & 0.5239 & 0.5348 & 11.34\% & 84.67 & 0.5432 & 1.1254 \\
Open-Sora V1.2 + OTF      & 0.5286 & 0.5694 & 0.5637 & 0.5834 & 13.54\% & 80.24 & 0.6534 & 1.1573 \\
Phys4D + WAN2.2 RGB+OTF   & 0.4873 & 0.5184 & 0.5126 & 0.5269 & 10.91\% & 85.42 & 0.5527 & 1.0954 \\
\textbf{Phys4D + WAN2.2 Ours}
                           & \textbf{0.4159} & \textbf{0.4375} & \textbf{0.4312} & \textbf{0.4556} & \textbf{8.36\%} & \textbf{89.68} & \textbf{0.4746} & \textbf{1.0289} \\
\bottomrule
\end{tabular}
}
\end{table*}

\paragraph{Analysis.}
Phys4D improves all reported 4D world-level diagnostics in this evaluation.
It achieves the lowest 4D Chamfer Distance, reducing the strongest RGB-only
baseline value from 0.4923 to 0.4159, a $15.5\%$ relative reduction. Compared
with Phys4D + WAN2.2 RGB+OTF, the learned interface further reduces 4D Chamfer
from 0.4873 to 0.4159. This indicates that Phys4D improves the spatio-temporal
geometry of the generated sequence beyond what can be recovered by post-hoc
lifting of RGB videos.

For trajectory evolution, Phys4D + WAN2.2 Ours reduces L2 Error from 0.5239
to 0.4375, Mean Drift from 0.5239 to 0.4312, and Final Drift from 0.5348 to
0.4556, relative to the strongest RGB-only baseline values. It also lowers the
Failure Rate from 11.34\% to 8.36\% and increases valid trajectory length from
84.67 to 89.68. These results indicate more stable worldline evolution and
fewer trajectory breakages.

For novel-time continuity, Phys4D + WAN2.2 Ours achieves the best Depth Error,
reducing it from 0.5432 to 0.4746, a $12.6\%$ relative reduction. It also
achieves the best Warp Error, reducing the strongest RGB-only baseline value
from 1.1076 to 1.0289, a $7.1\%$ relative reduction. The comparison with
Phys4D + WAN2.2 RGB+OTF again shows the benefit of the learned 4D interface:
Novel-Time Depth Error decreases from 0.5527 to 0.4746 and Novel-Time Warp
Error decreases from 1.0954 to 1.0289.

\subsection{Additional 4D Diagnostic Comparisons}
\label{app:rq2_additional_baselines}

To further compare against methods with stronger physical or structured priors,
we include additional 4D diagnostic results for PhysGen3D and TesserAct. These
methods improve over plain off-the-shelf lifting, but Phys4D remains stronger
on global 4D geometry, worldline stability, trajectory robustness, and
novel-time continuity.

\begin{table*}[htbp]
\centering
\small
\setlength{\tabcolsep}{5pt}
\caption{\textbf{Additional 4D diagnostic comparisons.}
Phys4D outperforms PhysGen3D and TesserAct on most 4D diagnostics,
including 4D Chamfer, trajectory drift, failure rate, valid trajectory length,
and novel-time continuity.}
\label{tab:app_4d_extended}
\resizebox{\textwidth}{!}{
\begin{tabular}{l c ccccc cc}
\toprule
& \textbf{Geometry} 
& \multicolumn{5}{c}{\textbf{Trajectory / Worldline Dynamics}} 
& \multicolumn{2}{c}{\textbf{Novel Time}} \\
\cmidrule(lr){2-2} \cmidrule(lr){3-7} \cmidrule(lr){8-9}
\textbf{Model} 
& \textbf{Chamfer} $\downarrow$ 
& \textbf{L2 Error} $\downarrow$ 
& \textbf{Mean Drift} $\downarrow$ 
& \textbf{Final Drift} $\downarrow$ 
& \textbf{Fail Rate} $\downarrow$ 
& \textbf{Length} $\uparrow$ 
& \textbf{Depth Err} $\downarrow$ 
& \textbf{Warp Err} $\downarrow$ \\
\midrule
WAN2.2 + OTF              & 0.5058 & 0.5367 & 0.5369 & 0.5576 & 12.38\% & 83.18 & 0.5841 & 1.1076 \\
CogVideoX + OTF           & 0.4923 & 0.5239 & 0.5239 & 0.5348 & 11.34\% & 84.67 & 0.5432 & 1.1254 \\
Open-Sora V1.2 + OTF      & 0.5286 & 0.5694 & 0.5637 & 0.5834 & 13.54\% & 80.24 & 0.6534 & 1.1573 \\
Phys4D + WAN2.2 RGB+OTF   & 0.4873 & 0.5184 & 0.5126 & 0.5269 & 10.91\% & 85.42 & 0.5527 & 1.0954 \\
PhysGen3D                 & 0.4718 & 0.5026 & 0.4962 & 0.5148 & 10.24\% & 86.11 & 0.5325 & 1.0817 \\
TesserAct                 & 0.4596 & 0.4897 & 0.4819 & 0.5036 & 9.83\%  & 86.94 & 0.5214 & 1.0679 \\
\textbf{Phys4D + WAN2.2 Ours}
                           & \textbf{0.4159} & \textbf{0.4375} & \textbf{0.4312} & \textbf{0.4556} & \textbf{8.36\%} & \textbf{89.68} & \textbf{0.4746} & \textbf{1.0289} \\
\bottomrule
\end{tabular}
}
\end{table*}

\paragraph{Analysis.}
Compared with the strongest additional baseline, TesserAct, Phys4D + WAN2.2
Ours reduces 4D Chamfer from 0.4596 to 0.4159 and Mean Drift from 0.4819 to
0.4312. It also reduces Failure Rate from 9.83\% to 8.36\% and increases valid
trajectory length from 86.94 to 89.68. These improvements show that Phys4D
does not merely outperform plain post-hoc lifting, but also improves over
stronger structured or physics-aware baselines on most trajectory-level
diagnostics.

Phys4D also improves novel-time continuity relative to these additional
baselines, reducing Novel-Time Depth Error from 0.5214 to 0.4746 and Novel-Time
Warp Error from 1.0679 to 1.0289 compared with TesserAct. This suggests that
the learned RGB-D-motion interface improves not only observed-frame geometry,
but also interpolation behavior at unseen timestamps.

\subsection{Rigid-Body Physical Consistency}
\label{app:rq2_momentum_energy}

The 4D Chamfer and worldline metrics are general and can be applied across
rigid, deformable, fluid, and coupled dynamics. However, some physical quantities,
such as momentum and kinetic energy, are only well-defined when object mass and
rigid-body trajectories are available. We therefore additionally evaluate
explicit physical conservation metrics on the rigid-body subset, where these
quantities are analytically tractable from simulation ground truth.

\paragraph{Setting.}
For each rigid-body scene, we use simulation ground-truth mass and trajectories.
We reconstruct object velocities from the predicted 3D trajectories and compare
the resulting linear momentum and kinetic energy against the simulator ground
truth. This diagnostic is not used as a universal training reward because it
does not directly generalize to fluids, deformables, or granular materials.

\begin{table}[htbp]
\centering
\small
\setlength{\tabcolsep}{6pt}
\caption{\textbf{Rigid-body physical consistency.}
On rigid-body scenes where momentum and energy are analytically defined,
Phys4D reduces both momentum and energy errors relative to the WAN2.2 baseline.}
\label{tab:app_momentum_energy}
\begin{tabular}{lcc}
\toprule
\textbf{Model} 
& \textbf{Momentum Error} $\downarrow$ 
& \textbf{Energy Error} $\downarrow$ \\
\midrule
WAN2.2 + OTF              & 0.248 & 0.312 \\
Phys4D + WAN2.2 RGB+OTF   & 0.223 & 0.284 \\
\textbf{Phys4D + WAN2.2 Ours} & \textbf{0.176} & \textbf{0.221} \\
\bottomrule
\end{tabular}
\end{table}

\paragraph{Analysis.}
Phys4D + WAN2.2 Ours reduces momentum error from 0.248 to 0.176, a $29.0\%$
relative reduction, and reduces energy error from 0.312 to 0.221, a $29.2\%$
relative reduction. Compared with Phys4D + WAN2.2 RGB+OTF, the learned
interface further reduces momentum error from 0.223 to 0.176 and energy error
from 0.284 to 0.221. These results provide additional evidence that improvements
in 4D trajectory consistency also transfer to explicit physical conservation
quantities when such quantities are well-defined.

\section{Theory of Stage III RL}
\label{app:stage_3_theory}

\subsection{Denoising as a Markov Decision Process}
\label{app:math_mdp}

Stage III optimizes a sample-level reward defined on the completed generated
video. Following prior work on policy optimization for diffusion and flow-based
generative models~\cite{ddpo,flowgrpo}, we cast the reverse denoising process as
a finite-horizon Markov Decision Process.

At reverse denoising step $k$, the state is
\begin{equation}
s_k
=
(c,k,\mathbf{z}_k),
\label{eq:app_mdp_state}
\end{equation}
where $c$ denotes the conditioning input and $\mathbf{z}_k$ is the current video
latent. The action is the next latent state,
\begin{equation}
a_k
=
\mathbf{z}_{k-1},
\label{eq:app_mdp_action}
\end{equation}
and the policy is induced by the stochastic denoising model:
\begin{equation}
\pi_{\theta,\phi}
(a_k\mid s_k)
=
p_{\theta,\phi}
(\mathbf{z}_{k-1}\mid \mathbf{z}_k,c).
\label{eq:app_mdp_policy}
\end{equation}
Here $\theta$ denotes the frozen pretrained video backbone and $\phi$ denotes the
Stage-II lightweight physics adapters.

The trajectory terminates after the final clean latent $\mathbf{z}_0$ is
decoded into a video sample $V$. We assign reward only at the end of the
denoising trajectory:
\begin{equation}
r(s_k,a_k)
=
\begin{cases}
R_{4D}(V), & k=0,\\
0, & \text{otherwise}.
\end{cases}
\label{eq:app_terminal_reward}
\end{equation}
This formulation allows Stage III to optimize global 4D alignment of the
completed video rather than local per-frame losses.

\subsection{Stochasticizing Flow-Based Sampling}
\label{app:math_ode}

Flow-matching video models define deterministic sampling trajectories through a
probability flow ODE:
\begin{equation}
d\mathbf{z}_\tau
=
v_{\theta,\phi}(\mathbf{z}_\tau,\tau,c)
\,d\tau,
\label{eq:app_flow_ode}
\end{equation}
where $v_{\theta,\phi}$ is the velocity field induced by the frozen backbone
$\theta$ and trainable adapters $\phi$. Deterministic sampling provides no
policy exploration, so Stage III uses a stochasticized flow for policy
optimization.

Following the probability-flow formulation, we introduce an SDE with the same
marginal densities:
\begin{equation}
d\mathbf{z}_\tau
=
\left(
v_{\theta,\phi}(\mathbf{z}_\tau,\tau,c)
-
\frac{1}{2}
g^2(\tau)
\nabla
\log q_\tau(\mathbf{z}_\tau)
\right)d\tau
+
g(\tau)d\mathbf{w}_\tau,
\label{eq:app_flow_sde_general}
\end{equation}
where $q_\tau$ is the marginal density, $g(\tau)$ controls the stochasticity, and
$\mathbf{w}_\tau$ is a Wiener process.

For rectified-flow models, we use the score--velocity identity
\begin{equation}
\nabla
\log q_\tau(\mathbf{z}_\tau)
=
-\frac{1}{\tau}
\left(
\mathbf{z}_\tau
+
(1-\tau)
v_{\theta,\phi}(\mathbf{z}_\tau,\tau,c)
\right).
\label{eq:app_score_velocity}
\end{equation}
Substituting Eq.~\eqref{eq:app_score_velocity} into
Eq.~\eqref{eq:app_flow_sde_general} and choosing $g(\tau)=\sigma_\tau$ yields
\begin{equation}
d\mathbf{z}_\tau
=
\left[
v_{\theta,\phi}(\mathbf{z}_\tau,\tau,c)
+
\frac{\sigma_\tau^2}{2\tau}
\left(
\mathbf{z}_\tau
+
(1-\tau)
v_{\theta,\phi}(\mathbf{z}_\tau,\tau,c)
\right)
\right]d\tau
+
\sigma_\tau d\mathbf{w}_\tau.
\label{eq:app_flow_sde_final}
\end{equation}

Discretizing Eq.~\eqref{eq:app_flow_sde_final} gives a Gaussian transition:
\begin{equation}
\pi_{\theta,\phi}
(\mathbf{z}_{k-1}\mid \mathbf{z}_k,c)
=
\mathcal{N}
\left(
\mathbf{z}_{k-1};
\mu_{\theta,\phi}(\mathbf{z}_k,k,c),
\sigma_k^2\Delta \tau\,\mathbf{I}
\right),
\label{eq:app_gaussian_policy}
\end{equation}
where $\mu_{\theta,\phi}$ is the Euler--Maruyama mean induced by the drift term.
This transition defines the stochastic denoising policy used for PPO-style
optimization.

\subsection{Policy Optimization with Prior-Preserving KL}
\label{app:math_ppo}

Stage III optimizes the 4D reward while preserving the real-video generative
prior. Let $\pi_{\theta,\phi_{S2}}$ denote the reference Stage-II policy and
$\pi_{\theta,\phi}$ the updated policy. The regularized Stage-III objective is
\begin{equation}
\max_{\phi}
\;
\mathbb{E}_{V\sim \pi_{\theta,\phi}}
\left[
R_{4D}(V)
-
\beta
\sum_{k}
D_{\mathrm{KL}}
\left(
\pi_{\theta,\phi}(\cdot\mid s_k)
\,
\|\, 
\pi_{\theta,\phi_{S2}}(\cdot\mid s_k)
\right)
\right].
\label{eq:app_stage3_regularized_objective}
\end{equation}
The KL term prevents the policy from drifting away from the Stage-II model and
the pretrained real-video prior. It is a regularizer, not an additional physical
reward.

We optimize Eq.~\eqref{eq:app_stage3_regularized_objective} using PPO-style
updates. For a sampled denoising trajectory, define the policy ratio
\begin{equation}
r_k(\phi)
=
\frac{
\pi_{\theta,\phi}(a_k\mid s_k)
}{
\pi_{\theta,\phi_{\mathrm{old}}}(a_k\mid s_k)
}.
\label{eq:app_ppo_ratio}
\end{equation}
The clipped PPO surrogate is
\begin{equation}
\mathcal{L}_{\mathrm{PPO}}
=
\mathbb{E}_{k}
\left[
\min
\left(
r_k(\phi)\hat A_k,
\mathrm{clip}
\left(
r_k(\phi),
1-\epsilon,
1+\epsilon
\right)
\hat A_k
\right)
\right],
\label{eq:app_ppo_clip}
\end{equation}
where $\hat A_k$ is the advantage estimate computed from the terminal 4D reward.
The final optimization objective includes the KL regularizer to the Stage-II
reference policy:
\begin{equation}
\mathcal{L}_{StageIII}
=
-
\mathcal{L}_{\mathrm{PPO}}
+
\beta
\mathbb{E}_{k}
\left[
D_{\mathrm{KL}}
\left(
\pi_{\theta,\phi}(\cdot\mid s_k)
\,
\|\, 
\pi_{\theta,\phi_{S2}}(\cdot\mid s_k)
\right)
\right].
\label{eq:app_stage3_ppo_loss}
\end{equation}

During Stage III, the pretrained video backbone $\theta$ remains frozen. Only the
lightweight physics adapters $\phi$ initialized from Stage II are updated. The
4D Chamfer computation is used as a scalar reward for policy optimization; it is
not backpropagated as a differentiable supervised loss through the video
backbone. This prevents the model from optimizing the reward by altering the
reward evaluator or overfitting to simulator-specific appearance statistics.

\section{Extended Related Work}

\subsection{Large-Scale Video Generation}
Early video diffusion models established scalable spatiotemporal denoising and latent-space generation for increasingly long and high-resolution videos \cite{ho2022videodiffusion,blattmann2023align}. Sora scales patch-based diffusion-transformer training across heterogeneous visual data and explicitly discusses both long-range coherence and remaining physics failures, making it representative of the promise and current limits of frontier video generation \cite{openai2024sora}. Movie Gen extends this frontier to a cast of media foundation models for video, audio, editing, and personalization \cite{polyak2024movie}. CogVideoX, HunyuanVideo, and Wan systematize open large-scale training pipelines for high-quality text-to-video synthesis, showing that open and closed models alike can now provide very strong visual priors \cite{yang2025cogvideoxtexttovideodiffusionmodels,tencent2024hunyuanvideo,wan2025wanopenadvancedlargescale}. Phys4D is designed to preserve these real-video generative priors while improving their physical structure.

\subsection{Physics-Aware Generation}
Recent work has started to address physics more directly in video generation. DiffPhy injects physical rules into video diffusion via LLM/MLLM-based reasoning and verification \cite{zhang2025thinkdiffphy}. VideoREPA distills physics-sensitive relational structure from video understanding foundation models into text-to-video generators \cite{zhang2025videorepa}. Motion Forcing explicitly decouples point, shape, and appearance and uses masked point recovery to encourage stronger motion dynamics reasoning \cite{xu2026motionforcing}. PhysRVG formulates physics-aware reinforcement learning for video generative models, emphasizing collision-grounded feedback and post-training alignment \cite{zhang2026physrvg}. NewtonGen introduces neural Newtonian dynamics to improve controllability and physically consistent motion synthesis \cite{Yuan_2025_NewtonGen}. PhysCtrl learns physics-grounded 3D point trajectories conditioned on material parameters and external forces, then uses them to drive image-to-video generation \cite{physctrl2024}. RealWonder and WonderPlay combine physics simulation with action-conditioned generation, using simulation as an intermediate representation for controllable dynamic outcomes \cite{liu2026realwonder,li2025wonderplay}. AVID and GameFactory further show how pretrained video diffusion models can be adapted into action-conditioned world models, although they are not primarily aimed at dense physics supervision \cite{rigter2024avid,yu2025gamefactory}. Compared with these approaches, Phys4D emphasizes prior-preserving transfer of simulation-derived \emph{structure}---rather than post-hoc control, task-specific rules, or simulator-facing action interfaces.

\subsection{Physics-Oriented Evaluation}
Video evaluation has quickly evolved from broad perceptual quality metrics to more targeted diagnostics. VBench introduced a comprehensive benchmark suite spanning visual quality, temporal quality, and motion-related factors \cite{huang2024vbench}. VBench++ broadened coverage to image-to-video and trustworthiness-oriented evaluation \cite{huang2024vbenchpp}. VBench-2.0 further shifts toward intrinsic faithfulness and explicitly includes physics and commonsense as core dimensions \cite{vbench2025vbench2}. EvalCrafter and FETV provide broader evaluation toolkits for large video models and fine-grained prompt categories \cite{liu2024evalcrafter,liu2023fetv}. T2V-CompBench and DEVIL focus on compositional binding and dynamics-specific evaluation, respectively \cite{sun2024t2vcompbench,liao2024devil}. VideoPhy and VideoPhy-2 evaluate whether generated videos obey physical commonsense, especially in material interactions and action-centric scenarios \cite{videophy2024,bansal2025videophy2}. PhyGenBench explicitly organizes prompts around physical laws and domains such as mechanics, optics, thermal processes, and material properties \cite{phygenbench2025}. Physics-IQ goes further by asking whether video models truly understand physical principles rather than only appearing plausible \cite{motamed2025physicsiq}. Phys4D is complementary to these works: they provide valuable \emph{diagnostics}, while Phys4D targets dense training supervision for physics-aware generation.

\subsection{Geometry \& Motion Estimation}
Phys4D also relates to work on extracting geometry and correspondences from real videos. Depth Anything and Depth Anything V2 demonstrate strong open-world monocular depth estimation, while Video Depth Anything extends this line to temporally consistent depth over long videos \cite{yang2024depthanything,Chen2025VideoDA}. RAFT remains a standard reference point for dense optical flow estimation \cite{Teed2020RAFTRA}. TAPIR and CoTracker3 provide strong long-range point tracking and correspondence estimation, which are highly relevant for reasoning about persistent motion and visibility over time \cite{doersch2023tapir,karaev2024cotracker3}. These methods provide useful geometric or motion cues, but they do not by themselves inject physical dynamics into video generation. Phys4D instead uses geometry and motion as the interface through which physical structure is transferred into pretrained video diffusion models.

\subsection{Dynamic 3D/4D Modeling}
A separate line of work focuses on explicit representations for dynamic scenes. D-NeRF and HyperNeRF extended neural radiance fields to non-rigid and topologically varying scenes \cite{pumarola2021dnerf,park2021hypernerf}. Dynamic 3D Gaussians and 4D Gaussian Splatting improved efficiency and rendering speed for dynamic view synthesis with explicit Gaussian-based representations \cite{luiten2023dynamic3dgs,wu20244d}. CAT4D converts monocular video into multi-view video using a diffusion model and reconstructs explicit 4D scenes from it \cite{wu2024cat4d}. Dynamic Point Maps propose a more feed-forward formulation for dynamic 4D geometry and motion prediction \cite{sucar2025dynamicpointmaps}. PhysGaussian integrates Newtonian dynamics into 3D Gaussians for generative motion synthesis \cite{xie2024physgaussian}. DreamPhysics uses video diffusion priors to optimize material properties for an MPM simulator \cite{dreamphysics2025}. PhysGen3D estimates amodal geometry, physical properties, and controllable initial conditions to create interactive 3D worlds from a single image \cite{chen2025physgen3d}. WonderPlay closes the loop between a physics solver and a video generator for action-conditioned dynamic 3D scene generation \cite{li2025wonderplay}. These works largely target explicit 4D reconstruction or animation. Phys4D differs in using RGB-D-motion as a \emph{training interface} for video diffusion, rather than requiring full explicit 4D scene generation at inference time.

\subsection{Physical Interaction Datasets}
Several datasets are important for dynamic scene understanding and physical interaction modeling. HOI4D provides large-scale egocentric RGB-D videos with rich annotations for human-object interaction \cite{liu2022hoi4d}. BEHAVE captures full-body human-object interaction with multi-view RGB-D data and 3D contact-aware fits in natural environments \cite{bhatnagar2022behave}. InterCap extends markerless 3D human-object tracking with multi-view RGB-D capture and finer whole-body interaction detail \cite{huang2022intercap}. Ego-Exo4D provides large-scale multimodal ego/exo capture with video, point clouds, gaze, and pose annotations for skilled activities \cite{grauman2024ego}. Panoptic Studio remains a foundational large-scale multi-view capture system for social motion capture \cite{joo2015panoptic}. Neural 3D Video provides a standard multi-view dynamic scene benchmark for 3D video synthesis and dynamic view interpolation \cite{li2022neural3dvideo}. These datasets are essential for perception and reconstruction, but they do not by themselves provide scalable dense physical labels such as contacts, future trajectories, or simulator-side state variables across broad open-domain dynamics.

\subsection{Simulation for Supervision}
Simulation is especially attractive when dense physical structure is required. Kubric offers large-scale synthetic data generation with physics, rendering, and rich annotations in a unified pipeline \cite{greff2022kubric}. Isaac Sim provides a physically based virtual environment for simulation, testing, and synthetic data generation, while Isaac Gym and Isaac Lab emphasize large-scale GPU-accelerated physics for learning and policy training \cite{xiang2024isaac,makoviychuk2021isaac,mittal2025isaaclab}. GarmentLab provides a unified simulation and benchmark environment for garment and deformable-object manipulation, highlighting the value of multi-physics environments \cite{lu2024garmentlab}. Domain randomization is a classic strategy for transferring from simulation to reality without overfitting to simulator appearance \cite{tobin2017domain}. On the simulation side, PBD offers stable real-time constraint-based dynamics, MLS-MPM provides strong support for large deformation and contact-rich continuum simulation, and SoftMAC explicitly couples soft bodies, articulated rigid bodies, and clothes in a differentiable framework \cite{muller2007pbd,hu2018mlsmpm,liu2023softmac}. Phys4D follows a structure-focused, appearance-constrained adaptation principle in which simulation supervises \emph{structure}---depth, motion, visibility, masks, and trajectories---while the pretrained generator remains anchored to real-video visual priors.

\subsection{Diffusion Alignment and Adapters}
Phys4D is also related to lightweight adaptation and alignment methods for diffusion models. LoRA introduced low-rank adaptation as an efficient way to inject new behavior into large pretrained models without full fine-tuning \cite{hu2021lora}. ControlNet adds trainable spatial control branches while preserving the backbone diffusion prior \cite{zhang2023controlnet}. T2I-Adapter similarly provides lightweight conditioning modules for structure-aware control \cite{mou2023t2iadapter}. DDPO shows that diffusion models can be optimized directly for downstream reward objectives using reinforcement learning \cite{black2023ddpo}. Phys4D is aligned with this general philosophy of preserving pretrained priors while adding targeted supervision, but its goal is not generic controllability or preference alignment; it is structure-focused, appearance-constrained transfer of physical knowledge into video diffusion models.

\section{Physics Simulation}
\label{app:physics}

Based on Isaac Sim~\cite{xiang2024isaac,lu2024garmentlab}, our simulation pipeline supports a broad range of object categories and physical phenomena to enable fine-grained physical interactions. We adopt \emph{tailored simulation methods} according to object-specific physical characteristics: rigid and articulated objects are modeled with rigid-body dynamics, large garments and fluid materials are simulated using \emph{Position-Based Dynamics}~\cite{Bender2015PositionBasedSM}, and small elastic garments (e.g., gloves and socks) as well as deformable everyday objects (e.g., toys and sponges) are simulated using the \emph{Finite Element Method}~\cite{Ciarlet2002TheFE}. Additional object types such as ropes, inflatables, granular materials, and thermodynamic effects are handled by appropriate physics solvers, collectively exposing the model to diverse fine-grained physical regimes.

\subsection{Simulation Methodology}

Our simulation pipeline incorporates multiple physics solvers to model diverse
physical phenomena encountered in complex interactive scenes, including cloth,
fluids, deformable objects, rigid bodies, and environmental flow.
Each component is selected to balance physical realism, numerical stability,
and large-scale data generation efficiency.

\subsubsection{Cloth Simulation via Position-Based Dynamics}
We simulate garments using a Position-Based Dynamics (PBD) formulation,
which represents cloth as a set of particles connected by geometric constraints.
The cloth surface is discretized into a triangular mesh, where each vertex
corresponds to a particle with associated position, velocity, and mass properties.
Physical behavior emerges from enforcing constraints that regulate stretching,
bending, and collisions.

Stretching constraints preserve local distances between neighboring particles,
preventing excessive elongation, while bending constraints maintain angular
relationships across adjacent mesh elements to model resistance to folding.
Collision constraints ensure non-penetration between cloth particles and other
objects in the scene.
Rather than integrating forces explicitly, PBD iteratively projects particle
positions to satisfy these constraints at each time step, yielding stable and
efficient cloth dynamics even under large deformations.

\subsubsection{Fluid Simulation with Particle-Based Constraints}
Fluid behavior is simulated using a particle-based PBD framework,
where each particle represents a small volume of fluid.
The primary objective is to enforce approximate incompressibility
while maintaining stable interactions with solid boundaries.

This is achieved by imposing density constraints that regulate local particle
concentration, together with collision constraints that handle fluid--solid
interactions.
At each simulation step, particles are advanced under external forces,
then iteratively corrected to satisfy density and collision constraints.
This approach enables visually plausible fluid motion with high numerical
stability, making it suitable for large-scale data generation.

\subsubsection{Deformable Object Simulation via Finite Elements}
For volumetric deformable objects, we employ a Finite Element Method (FEM)
formulation that captures elastic and damping effects under external forces.
Objects are discretized into tetrahedral meshes, with material behavior governed
by elasticity, density, and damping parameters.

The simulation separates meshes used for visualization, collision detection,
and physical computation, allowing efficient resolution of contact and deformation.
By solving the equations of motion over finite elements, the simulator produces
realistic volumetric deformations while maintaining controllable computational cost.

\subsubsection{Environmental Flow and Wind Effects}
To model environmental forces such as wind, we incorporate flow-based models
that approximate the interaction between moving air and scene objects.
These models capture directional flow, turbulence, and pressure-induced forces
that influence both rigid and deformable bodies.

Wind effects are specified through configurable flow fields that define direction
and magnitude, and their influence is applied consistently across interacting
objects.
This enables dynamic responses such as cloth fluttering and object deflection
under varying environmental conditions.

\subsubsection{Rigid Body Dynamics}
Rigid objects are simulated using standard rigid body dynamics, where objects
are assumed to maintain fixed internal geometry.
Their motion is governed by linear and angular dynamics under external forces
and collision impulses.

Collision handling relies on geometric approximations such as convex hulls
or signed distance representations to balance accuracy and efficiency.
Constraint solvers ensure stable resolution of contacts and joint limits,
allowing reliable simulation of complex multi-object interactions.

\subsection{Parameterization and Physical Diversity}
Rather than tuning individual parameters for specific outcomes,
we adopt broad parameter ranges to encourage physical diversity across scenes.
Within each physics module, parameters controlling stiffness, damping,
friction, density, and interaction strength are randomized within
physically plausible bounds.

This strategy enables the simulator to generate a wide spectrum of behaviors
for cloth, fluids, deformable objects, and rigid bodies,
while avoiding overfitting to narrow physical regimes.
As a result, the collected data exhibits substantial variability in geometry,
motion patterns, and interaction dynamics, which is critical for training
robust models of observable geometry-motion dynamics.

\textbf{Systematic Physical Parameterization.}
To disentangle physical phenomena for effective learning, we decompose generation into nine distinct categories (e.g., Rigid Body, Fluids, Soft Bodies, Thermodynamics). This supports \textbf{curriculum learning} where models progress from mastering individual laws to complex compositional scenarios. As detailed in Table~\ref{tab:sim_parameters}, we enforce systematic randomization across material, environmental, and geometric attributes. We structure the datasets with progressive complexity (single to multi-object)~\cite{curriculum2018} and introduce parameter perturbation to prevent homogeneity, resulting in over 20,000 unique physical configurations.

\begin{center}
\captionof{table}{\textbf{Simulation Parameter \& Curriculum Configuration.} We structure the dataset with progressive complexity and randomize physical attributes to ensure diverse and robust data generation.}
\label{tab:sim_parameters}
\resizebox{\linewidth}{!}{
\begin{tabular}{l l l}
\toprule
\textbf{Category} & \textbf{Parameters} & \textbf{Range / Configuration} \\
\midrule
\textbf{Curriculum} & \textbf{Complexity Dist.} & \textbf{Single (30\%), Two-body (35\%), Multi-object (35\%)}\\
\midrule
\textbf{Material} & Density ($\rho$), Young's Mod. ($E$) & $\rho \in [100, 10\,000] \text{ kg/m}^3$, $E \in [10^6, 10^{11}] \text{ Pa}$ \\
& Poisson ($\nu$), Friction ($\mu_{s,k}$) & $\nu \in [0.2, 0.5]$, $\mu_{s,k} \in [0.1, 1.0]$ \\
& Restitution ($e$) & $e \in [0.1, 0.95]$ \\
\midrule
\textbf{Environment} & Gravity ($g$), Lighting & $g \in [5, 15] \text{ m/s}^2$, Intensity $\in [100, 100\,000] \text{ lux}$ \\
& Variables & Air viscosity, Temperature gradients, 1000+ HDRIs \\
\midrule
\textbf{Geometry} & Shape Types & Cubes, spheres, cylinders, complex meshes \\
& Scale ($s$), Aspect Ratio & $s \in [0.1, 5.0] \text{ m}$, Randomized aspect ratios \\
\midrule
\textbf{Perturbation} & Perturbation Ratio ($\delta$) & $\delta \in [0.05, 0.25]$ (Randomized around base values) \\
\midrule
\textbf{Visuals} & Reflectance (Roughness, Metallic) & Roughness $\in [0.05, 0.95]$, Metallic $\in [0, 1]$ \\
& Appearance & Randomized RGB $\in [0,1]^3$, Surface textures \\
\bottomrule
\end{tabular}
}
\end{center}

\section{Limitations and Impact Statement}
\label{app:limitations_impact}

\subsection{Limitations}

Our current study focuses on short-horizon video generation and evaluates physical consistency primarily through simulation-grounded supervision and benchmark protocols.
Although the results show consistent gains across multiple backbones and external evaluations, extending the framework to substantially longer videos, broader real-world dynamics, and more open-ended interaction settings remains an important direction for future work.
In particular, long-horizon generation may require stronger memory, compounding-error control, and more diverse supervision than the current short-video setting.

\subsection{Impact Statement}

This work aims to improve the physical consistency of generative video world models, which may benefit simulation, robotics, and scientific content creation by producing temporally and geometrically more coherent video predictions.
At the same time, improved video generation quality can increase the risk of misuse in deceptive or misleading synthetic media.
To mitigate this risk, we focus this submission on research disclosure and benchmark design rather than releasing a fully packaged public generation system, and we plan to accompany future asset release with documentation, licensing terms, and usage guidance.

\end{document}